\documentclass[table, acmsmall, manuscript]{acmart}

\usepackage[linesnumbered,ruled,vlined]{algorithm2e}
\usepackage{subcaption}    % Include the 'subscription' package for subtables
\usepackage{graphicx}
\usepackage{multirow}
\usepackage{colortbl} % Add this line to include the colorful package
\usepackage{listings}
\usepackage{caption}
\usepackage{pifont}% http://ctan.org/pkg/pifont
\usepackage{booktabs}
\newcommand{\cmark}{\ding{51}}%
\newcommand{\xmark}{\ding{55}}%
\AtBeginDocument{%
  \providecommand\BibTeX{{%
    \normalfont B\kern-0.5em{\scshape i\kern-0.25em b}\kern-0.8em\TeX}}}

\setcopyright{acmcopyright}
\acmDOI{}
\acmISBN{}
\acmSubmissionID{}
\acmJournal{CSUR}
\copyrightyear{2024}

\PassOptionsToPackage{square,sort,comma,numbers}{natbib}

\setcitestyle{nosort}

\begin{document}

%% allowing the author to define a "short title" in page headers.
\title{A Systematic Review of Federated Generative Models}

\author{Ashkan Vedadi Gargary}
\email{aveda002@ucr.edu}
\orcid{0009-0004-2128-5735}
\affiliation{%
  \institution{University of California, Riverside}
  \department{Department of Computer Science and Engineering}
  \city{Riverside}
  \state{California}
  \country{USA}
  \postcode{92521}
}

\author{Emiliano De Cristofaro}
\email{emiliand@ucr.edu}
\orcid{0000-0002-7097-6346}
\affiliation{%
  \institution{University of California, Riverside}
  \department{Department of Computer Science and Engineering}
  \city{Riverside}
  \state{California}
  \country{USA}
  \postcode{92521}
}

\begin{abstract}
    Federated Learning (FL) has emerged as a solution for distributed systems that allow clients to train models on their data and only share models instead of local data. Generative Models are designed to learn the distribution of a dataset and generate new data samples that are similar to the original data. Many prior works have tried proposing Federated Generative Models. Using Federated Learning and Generative Models together can be susceptible to attacks, and designing the optimal architecture remains challenging. 
    
    This survey covers the growing interest in the intersection of FL and Generative Models by comprehensively reviewing research conducted from 2019 to 2024. We systematically compare nearly 100 papers, focusing on their FL and Generative Model methods and privacy considerations. To make this field more accessible to newcomers, we highlight the state-of-the-art advancements and identify unresolved challenges, offering insights for future research in this evolving field.
\end{abstract}



%%


\maketitle

\section{Introduction}
%Why are we doing this: motivation
Deep Generative Models (a class of Machine Learning models) are designed to learn the underlying probability distribution of a dataset and generate new data samples that are similar to the original data~\cite{oussidi2018deep}. 
Many generative models such as GANs~\cite{goodfellow2020generative}, VAEs~\cite{wan2017variational}, and Diffusion Models~\cite{ho2020denoising} have been used during these years.

\begin{figure}[t]
    \centering
    \includegraphics[width=.9\textwidth]{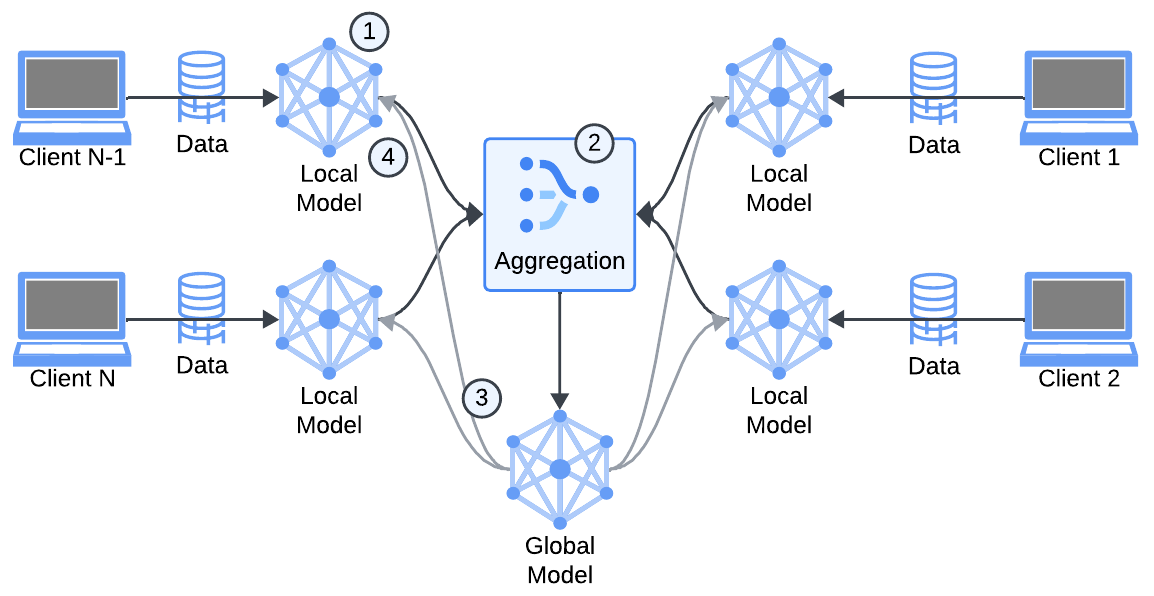}
    \caption{Overview of Federated Learning}
    \label{fig:fl-overview}
\end{figure}

With more people wanting machine learning models that work on many devices without sharing local files, Federated Learning has become popular~\cite{mcmahan2017communication}. This method allows clients to train models on their data and only share models, not the actual data, with a central server, as shown in Figure \ref{fig:fl-overview}. This keeps data private and facilitates the learning process.
% ~\cite{zhao2018federated, zhu2021federated, ghosh2019robust, ye2023heterogeneoussurvey}.

In recent years, many prior works have tried integrating generative models into Federated Learning setups to protect sensitive data and increase model performance by preventing data sharing with central servers. FL and Generative Models can work together in various ways. We can split all the related research into three groups: First, Generative Models can work in a federated manner. This means we can facilitate the generative process in distributed systems without sharing the raw local data with the central servers.
Second, Generative Models can attack FL models or protect them from different attack methods, highlighting security concerns. Third, Generative Models can be used within the FL model to address data heterogeneity and Non-IID.

\paragraph{\textbf{Papers Finding Criteria and Roadmap}}
While extensive research has explored various combinations of Federated Learning (FL) and generative models, this survey focuses explicitly on Federated Generative Models. The main goal of using Federate Generative Models is to keep data in each client to protect them compared to Centralized methods. Finding the best approach according to different applications (e.g., Medical Images as diagnostic tools, Anomaly Detection, Data Augmentation, and Financial Fraud Detection) is challenging for researchers.

We organized all the critical information into large tables. These tables detail the generative model methods, FL methods, data types, privacy measures, evaluation methods, the number of devices involved in the studies, and whether the code is shared (see Section \ref{sec:criteria}). Each paper and methodology is summarized to provide an overview of each approach (see Section \ref{sec:model-review}). Additionally, we include a section dedicated to papers that evaluate integrity and privacy within the context of Federated Generative Models (Section \ref{sec:security}).

In Section \ref{sec:discussion}, we discuss the main takeaways and state-of-the-art Federated Generative Models for further research comparison.

\paragraph{\textbf{Main Takeawyas}} Overall, we find that:
\begin{itemize}
    \item Federated GANs have garnered significant attention from researchers, resulting in numerous clinical applications. Many well-developed papers address privacy and integrity evaluation, satisfying the differential privacy (DP) definition, which enhances the robustness of Federated GANs for various applications such as financial problems and IoT devices. 
    \item Recently, several Diffusion-based Federated Models have been proposed, outperforming well-known GAN-based FL models regarding convergence and communication costs.
    \item Scalability and cross-device FL remain open research challenges that require further investigation, particularly in the context of non-GAN-based FL.
    \item Privacy and integrity consideration in tabular data-based models and non-GAN-based FL remain unsolved and necessitate additional research.
    \item One-shot FL, pre-trained Diffusion Models, and LLM-based generative FL are emerging as hot topics, attracting significant interest from researchers.
\end{itemize}

\section{Background}
\label{sec:background} 
In this section, we elaborate on background knowledge of FL and related concepts (e.g., aggregation methods and different types of FL), DP, and Generative Models.  

\subsection{Federated Learning}
The Federated Learning proposed by Google operates as a decentralized machine learning approach that trains across various clients without the necessity of sharing their local data~\cite{mcmahan2017communication}. There are two concepts in FL: (i) Single Global Server and (ii) Multiple Clients. Instead of sharing (possibly sensitive) training data with a central server, each client (a.k.a participant), such as a smartphone or computer, train local models on their individual datasets and only share model updates. The
central server only sees and aggregates the model updates and propagates the global model to the clients~\cite{mcmahan2017communication}. In general, the training process of a FL system can be summarized in the following steps (Shown in Figure \ref{fig:fl-overview})~\cite{wei2020federated}, 

\begin{enumerate}
    \item \textbf{Local Training}: Each client trains on its local dataset and sends the model updates to a centralized server locally.
    \item \textbf{Aggregation}: The server receives model updates from all participating clients and performs secure aggregation over the uploaded parameters without learning local information.
    \item \textbf{Aggregated Parameters Broadcasting}: The server broadcasts the aggregated parameters of model updates to all clients.
    \item \textbf{Updating Local Models}: Each client updates its local model with the aggregate parameter received from the server, thereby improving its performance.
\end{enumerate}

% Federated Learning's decentralized nature makes it particularly suitable for scenarios in which clients do not want to share their local data. 

\subsubsection{Types of Federated Learning}
\label{sec:type-fl}
Different Federated Learning (FL) types are based on the differences in feature size and data type among clients and the central server. 
\paragraph{Centralized vs. Decentralized}
This category refers to how the training is designed. Centralized FL—the most common approach—uses a central server to manage various model training and aggregation steps across all local data sources. On the other hand, Decentralized FL (also known as peer-to-peer) involves individual clients coordinating among themselves without a central server. In this case, model parameters are passed on from one client to the other in a chain for training.

\begin{figure}[t]
    \centering
    \begin{tabular}{ccc} % The tabular environment with three centered columns
        \includegraphics[width=0.3\textwidth]{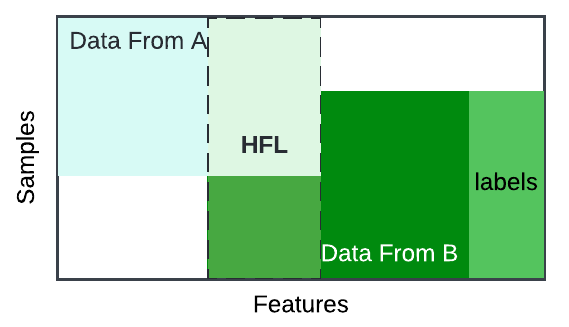} &
        \includegraphics[width=0.3\textwidth]{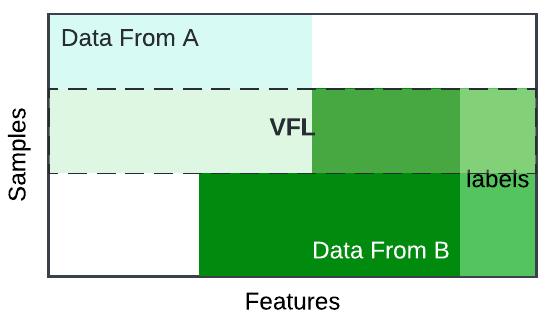} &
        \includegraphics[width=0.3\textwidth]{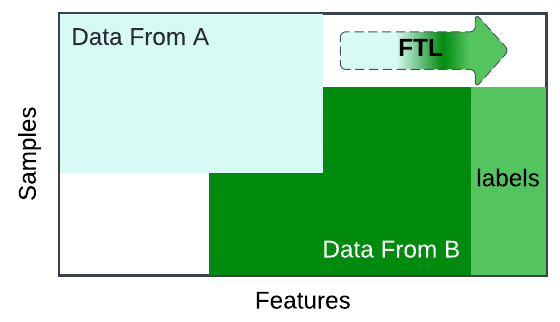} \\
        (a) Horizontal FL & (b) Vertical FL & (c) Federated Transfer Learning \\
    \end{tabular}
    \caption{Different data partition in Federated Learning.} % Caption for the whole figure
    \label{fig:fl-partition}
\end{figure}

\paragraph{Horizontal, Vertical, and Trasfer Learning}
This type is based on how data among different clients are partitioned. In Horizontal FL (a.k.a homogeneous or sample-based), the datasets of different clients have the same features but (little) overlap in the sample space~\cite{zhao2021efficient, yang2019federated}. In contrast, Vertical FL (also known as heterogeneous or feature-based) is used for datasets that contain different feature sets~\cite{wei2022vertical, liu2022vertical}. Additionally, Federated Transfer Learning (FTL) combines concepts of FL and Transfer Learning by transferring knowledge learned from one or more clients (which have plenty of data) to a target domain (which has limited data) under the Federated Learning setting (shown in Figure \ref{fig:fl-partition}).

Assume we have two datasets named $D_i, D_j$. Let $x_i, x_j$ denote feature space and $y_i, y_j$ denote the label space of $D_i, D_j$ respectively. Also, $I_i, I_j$ represents the sample ID space. Equation \ref{equation:HFL} mathematically represents the concept of HFL. Equation \ref{equation:VFL} mathematically demonstrates the concept of VFL and TFL~\cite{guendouzi2023systematic}.

\begin{equation} 
\label{equation:HFL}
    x_i = x_j, y_i = y_j, I_i \neq I_j, \forall D_i,D_j, i\neq j
\end{equation}

\begin{equation} 
\label{equation:VFL}
    x_i \neq x_j, y_i \neq y_j, I_i = I_j, \forall D_i,D_j, i \neq j
\end{equation}

\paragraph{Cross-Silo vs. Cross-Device}
This category depends on the type and size of the devices involved. Cross-silo FL involves models trained on data distributed across different organizations, typically using larger computing devices with a relatively small number of silos or training sets~\cite{huang2022cross, huang2021personalized, tran2024personalized}. On the other hand, cross-device FL occurs at the edge of IoT devices, such as smartphones, involving many millions of devices with lower computing power each~\cite{karimireddy2021breaking}.

% \begin{itemize}
%     \item \textbf{Centralized vs. Decentralized/Peer-to-Peer} - How the training is orchestrated
%     \item \textbf{Horizontal vs. Vertical} - How datasets are partitioned
%     \item \textbf{Cross-Silo vs. Cross-Device} - What types of devices are involved
% \end{itemize}

% \begin{itemize}
%     \item \textbf{Horizontal FL (HFL) ~\cite{zhao2021efficient, yang2019federated}}: The datasets of different clients have the same features but with little intersection of sample space.
%     \item \textbf{Vertical FL (VFL)~\cite{wei2022vertical, liu2022vertical}}:
%     \item \textbf{Federated Transfer Learning (FTL)~\cite{liu2020secure}}:
%     \item \textbf{ Cross-Silo FL~\cite{huang2022cross, huang2021personalized, tran2024personalized}}:
%     \item \textbf{Cross-Device FL~\cite{karimireddy2021breaking}}:
% \end{itemize}

% such as Horizontal FL (HFL)~\cite{zhao2021efficient, yang2019federated}, Vertical FL (VFL)~\cite{wei2022vertical, liu2022vertical}, Federated Transfer Learning (FTL)~\cite{liu2020secure}, Cross-Silo FL~\cite{huang2022cross, huang2021personalized, tran2024personalized} and Cross-Device FL~\cite{karimireddy2021breaking}.

% types of aggregation
\subsubsection{Types of Aggregation}
An aggregation approach is essential to make the process of collaborating the obtained results efficient, whether the messages exchanged are the models themselves, some or all of their parameters, or gradients~\cite{pillutla2022robust}.

Unlike traditional centralized approaches, training data are not pooled at a central server. Each client trains its model on its data locally and then only shares updated parameters with a server.
The central server aggregates models updated parameters from each client without access to the raw data. The various aggregation methods are available in Federated Learning~\cite{qi2023model}. Typically, this happens over multiple rounds; eventually, the model converges, and the parameters are finalized. Based on the different types of Federated Learning (FL) discussed in Section \ref{sec:type-fl} and different data distribution will be discussed in Section \ref{sec:iid}, the appropriate choice of model aggregation approach varies~\cite{sharma2023federated}. The most common methods for parameter aggregation in the reviewed paper include FedAVG, FedSGD~\cite{mcmahan2017communication}, and FedProx~\cite{li2020federated}.

\paragraph{FedAvg}
One of the earliest and most widely used methods in FL is FedAVG (Federated Averaging)~\cite{mcmahan2017communication}. During the aggregation, the parameters of each client are weighted and averaged to produce a global model. The model is training iteratively; let $\theta^{t}_{global}$ denote the latest global model aggregated by the central server at the iteration step $t$, and $m_i$ (for $i = 1,2,3,4,...,K$) denote the devices of all K clients. The 
central server first broadcasts $\theta^{t}_{global}$ to all $m_i$, then, every device (say the $k$-th) initializes $\theta^{t}_{i}$ as $\theta^{t}_{global}$. Then, the $i$-th clients performs $E$ (where $E>0$) local updates: 
\begin{equation}
\theta_i^t(k+1)=\theta_i^t(k)-\eta_k \nabla \ell\left(\theta_i^t(k),\left\{x_{i, j}, y_{i, j}\right\}\right)
\end{equation}
where $k = 1,2,3,...,E$ denotes the local training step and $\ell$ is the classification loss function used by the $i$-th clients, while $x_{i, j}, y_{i, j}$ (where $j = 1,2,3,...,|V_{i}|$) are the training instances hosted locally by the $i$-th client. Finally, $\theta_i^t(k)$ is the local model updated at the $k$-th step of the stochastic gradient descent performed locally; note that $\theta_i^t(k) = \theta_i^t = \theta^{t}_{global}$ when $k = 0$. 

After the clients finish training their local models, they send their parameters to the central server. The global model is derived at the central server and is then aggregated by averaging the local models:
\begin{equation}
    \theta_{global}^{t+1} = \sum_{i=1}^{K}\theta^{t}_{i}(E)
\end{equation}

Note that $\theta^{t}_{i}(E)$ denotes the models trained locally, after $E$ (where $E>0$) rounds of gradients descent~\cite{naseri2022cerberus, mcmahan2017communication}.

\paragraph{FedSGD}
FedSGD (Federated Stochastic Gradient) is also one of the earliest proposed aggregation methods. According to ~\cite{mcmahan2017communication}, FedSGD is a special case of FedAVG. With FedSGD, the clients train their local models using batch Gradient Descent with just one local epoch and upload their gradient $\theta^{t}_{i}$. In other words, we have $E = 1$, and the entire local dataset is treated as a single minibatch for FedSGD.

FedAvg is generally preferred over FedSGD in many practical Federated Learning scenarios due to its reduced communication requirements and better utilization of local computation resources. However, the best choice can depend on specific factors like network bandwidth, clients' computational capabilities, and data distribution.

\paragraph{FedProx}
FedProx ~\cite{li2020federated} extends FedAVG by focusing on handling data and system heterogeneity in Federated Learning environments. In FedProx, clients optimize the loss function by using a regularization term. This term helps mitigate the divergence between the current local and previously obtained global models. 
To address system heterogeneity, FedProx suggests directly solving the previous objective function rather than training local models over multiple epochs. This method can reduce computational overhead.

% \begin{figure}[h]
%     \centering
%     \includegraphics[width=.7\textwidth]{Images/FL-decentralized-centralized.png}
%     \caption{Centralized vs. Decentralized Aggregation}
%     \label{fig:validation-architecture}
% \end{figure}

\subsubsection{Non-IID vs. IID}
\label{sec:iid}
In Federated Learning, IID (Independent and Identically Distributed) and Non-IID data distribution refer to how the data is distributed among the clients.

% \paragraph{IID (Independent and Identically Distributed)}
In an IID setting, the assumption is that the data across different clients are (i) independent of each other and (ii) drawn from the same probability distribution.
In other words, we suppose that we have $K$ clients and let $D_i$ (for $i=1,2,3,..., K$) denote the dataset of each client that is used to train a local model. We have an IID setting if,~\cite{arafeh2022independent}:
\begin{itemize}
    \item For any class $y$ and a feature set $\mathrm{X}, P(X, y)=P(X) . P(y)$ (Independent). 
    \item For any class $y, P\left(y \mid D_1\right)=P\left(y \mid D_2\right) \ldots=P\left(y \mid D_n\right)$ (Identically Distributed).
\end{itemize}

Violating any of the abovementioned rules implies a Non-IID context~\cite{zhao2018federated}.

% Each client's data is representative of the overall data distribution, and there is no correlation between the samples from different clients~\cite{zhao2018federated}.
% \paragraph{Non-IID (Non-Independent and Non-Identically Distributed)}
% In a Non-IID. The setting and the data across clients may exhibit dependencies or variations in distribution. The data on different clients may come from different sources, have different characteristics, or follow different distributions. This is a common scenario in Federated Learning because each client generates data based on the user's unique behavior or environment, making the data distribution specific to that client and different from others.

% IID settings are often assumed in traditional machine learning models, but in real-world FL scenarios, data is usually Non-IID due to the decentralized nature of the data sources~\cite{zhao2018federated}.

% different approaches 
\subsection{Different Generative Models}
Deep generative models (DGMs) are machine learning models designed to learn the underlying probability distribution of a dataset and generate new data samples similar to the original data. These models are powerful tools in both supervised and unsupervised learning tasks, and they have a wide range of applications from image synthesis to natural language processing~\cite{oussidi2018deep}. 

Generative Model functionality can categorized into Data Generation, Learning Data Distributions, Unsupervised Learning, Feature Learning, and Anomaly Detection~\cite{sun2021adversarial}.

The primary function of DGMs is to \textit{generate new data} instances that mimic the original dataset. DGMs can create realistic images, videos, sounds, and text indistinguishable from real-world data~\cite{salakhutdinov2015learning}. DGMs learn to capture the complex probability \textit{distributions} of the input data by understanding the patterns and structures within the data, which enables the models to produce high-quality outputs~\cite{oussidi2018deep, jothiraj2023phoenix}. Besides, By learning the normal distribution of a dataset, generative models can identify outliers or \textit{anomalies} that do not fit the learned distribution. Anomaly detection is useful in fraud detection, monitoring systems, and healthcare. Moreover,  DGMs can learn from unlabeled data as an \textit{unsupervised} learning mechanism, particularly useful in scenarios where labeled data is scarce or expensive to obtain~\cite{harshvardhan2020comprehensive}. In particular, DGMs are capable of \textit{learning useful features} from the data in an unsupervised manner~\cite{zhao2017learning}.

% We often use Deep Generative Models (DGMs) to create data. A deep learning approach in Generative AI that can estimate the likelihood of each data to generate new samples based on underlying distribution~\cite{jothiraj2023phoenix}. The effectiveness of these models depends on how we define the problem and how our models handle different types of data. Generative Model functionality can categorized into synthetic data generation, style transfer, data augmentation, and anomaly detection~\cite{sun2021adversarial}. 
Here, we list the most common Generative Models:
    \paragraph{\textbf{Generative Adversarial Networks (GANs):}} Generative Adversarial Networks (GANs) are a class of machine learning algorithms characterized by two neural networks—the generator and the discriminator—engaged in a min-max game. In this setup, the generator aims to produce data so realistically that the discriminator cannot distinguish them from actual data. On the other hand, the discriminator's goal is to accurately classify real data from the generator's fakes~\cite{goodfellow2020generative}. This min-max game refers to the optimization problem where the generator tries to maximize the probability of the discriminator making a mistake (minimizing its loss). In contrast, the discriminator tries to reduce this probability (maximizing its accuracy). This dynamic competition drives both networks to improve their performance, leading to the generation of realistic synthetic data~\cite{creswell2018generative}.
    
    \paragraph{\textbf{Variational Auto-encoder (VAEs):}} Variational Auto-encoders are a type of generative model that leverage the architecture of auto-encoders to create high-dimensional data~\cite{wan2017variational}. They consist of an encoder that compresses the input data into a lower-dimensional representation and a decoder that reconstructs the data from this representation. However, unlike traditional auto-encoders, VAEs introduce a probabilistic approach to the encoding process, making them capable of generating new data points by sampling from the learned distribution in the latent space~\cite{acs2018differentially}. 
    
    \paragraph{\textbf{Diffusion Models:}} Diffusion models are a class of generative models that simulate the gradual process of adding and then removing noise to generate data~\cite{kingma2021variational,jothiraj2023phoenix}. The model starts with a distribution of pure noise and gradually refines this into samples from a target distribution through a reverse diffusion process. This approach generates highly detailed and coherent samples, making Diffusion Models particularly effective in high-quality image and audio generation. The Diffusion Model is a probabilistic generative model that uses a parameterized Markov chain. It is trained through variational inference to learn the process of reversing the gradual degradation of training data structures~\cite{wang2023fda}. ~\cite{ho2020denoising} introduced the Denoising Diffusion Probabilistic Models (DDPM), which improve the original Diffusion Model's mathematical framework.\\

This survey mainly focuses on using DGMs for data generation and synthetic data. There are several reasons why synthetic data is used in machine learning. 

The first and most important, real-world data may contain sensitive information. Synthetic data provides a way to create realistic data without revealing important details, making it useful for training models without privacy concerns.
Second, synthetic data can augment existing datasets. By generating additional samples, models can be trained on more diverse examples, improving generalization and robustness. Also, in situations where the distribution of classes in a dataset is imbalanced, synthetic data generation can help balance the classes.
Finally, When a model is trained on data from one domain but needs to perform well in a different domain, synthetic data can be generated to simulate the target domain~\cite{oussidi2018deep, de2024synthetic}.

\subsection{Differential Privacy}
%DP Definition
DP is the established framework to define algorithms resilient to adversarial inferences. It provides an unconditional upper bound on the privacy loss of individual data subjects from the output of an algorithm by introducing statistical noise~\cite{dwork2006calibrating}.
In other words, It aims to protect sensitive information by adding noise into the data generation process or data sharing from client to server while maintaining the overall statistical properties of the original dataset. It has become the state-of-the-art paradigm for protecting individual privacy in statistical databases~\cite{dwork2006calibrating}.  

\begin{definition}[Differential Privacy~\cite{differntialprivacy2014dwork}]
A randomized mechanism $\mathcal{M}:\mathcal{D}\rightarrow\mathcal{R}$ is $(\epsilon, \delta)$-differentially private if for any two neighboring datasets set ${d},d^{\prime}\in\mathcal{D}$ and $S\subseteq\mathcal{R}$
\begin{equation}
\mathrm{P}(\mathcal{M}(d) \in S) \leq \mathrm{e}^\epsilon \mathrm{P}\left(\mathcal{M}\left(d^{\prime}\right) \in S\right)+\delta
\end{equation}
\label{def:DP}
\end{definition}

According to the Definition \ref{def:DP}, the definition of neighboring datasets depends on the setting, and thus, it can vary. The $\epsilon$ parameter (a.k.a privacy budget) is a metric of privacy loss (in a range from $0$ to $\infty$). It also controls the privacy-utility trade-off, i.e., lower $\epsilon$ values indicate higher levels of privacy but likely reduce utility too. The $\delta$ parameter accounts for a (small) probability on which the upper bound $\epsilon$ does not hold. The amount of noise
needed to achieve DP is proportional to the sensitivity of the output; this measures the maximum change in the output due to the inclusion or removal of a single record~\cite{annamalai2023fp, naseri2020local}. 

To provide individual privacy in case of $\epsilon$-DP failure, the recommended value for $\delta$ should be smaller than the inverse of the database size, i.e., $\frac{1}{|\mathcal{D}|}$. 
% Recently, another relaxation of $\epsilon$-DP, Rényi Differential Privacy (RDP)~\cite{mironov2017renyi} has been proposed with a stronger privacy notion than $(\epsilon, \delta)$-DP.

\subsection{Differential Privacy in Federated Learning}
As mentioned, in the context of FL, one can use one of two variants of DP, namely, local and central.

\subsubsection{Local Differential Private (LDP)}
With LDP, each client performs the noise addition required for DP locally.
Each client runs a random perturbation algorithm $M$ and sends the results to the server. The perturbed result is guaranteed to protect an individual’s data according to the $\epsilon$ value. This is formally defined next~\cite{naseri2020local}.

\begin{definition}[Locally Differential Private~\cite{truex2020ldp}]
Let X be a set of possible values and Y the noisy values. $M$ is ($\epsilon$, $\delta$)-locally differentially private ($\epsilon$-LDP) if for all $x1, x2 \in X$ and for all $y \in Y$:
\begin{equation}
\mathrm{P}(\mathcal{M}(x)=y) \leq \mathrm{e}^\epsilon \mathrm{P}(\mathcal{M}\left(x^{\prime}\right) = y)+\delta
\end{equation}
\end{definition}

\subsubsection{Central Differential Private (CDP)} 
With CDP, the FL aggregation function is perturbed by the server, and this provides client-level DP. This guarantees that the output of the aggregation function is indistinguishable, with probability bounded by $\epsilon$, to whether or not a given client is part of the training process. In this setting, clients need to trust the server: 1) with their model updates and 2) to correctly perform perturbation by adding noise, etc. While some degree of trust in the server is needed, this is a much weaker assumption than entrusting the server with the data itself~\cite{geyer2017differentially, mcmahan2017learning}. If anything, inferring training set membership or properties from the model updates is much less of a significant privacy threat than having data in the clear. Moreover, in FL, clients do not share entire datasets for efficiency reasons and because they might be unable to for policy or legal reasons~\cite{naseri2020local}.

In Algorithm \ref{Alg:dpfedavg}, we elaborate on the details of the Federated Learning with Average Aggregation (FedAVG) and Central Differential Private~\cite{geyer2017differentially, mcmahan2017learning, annamalai2023fp}.

\begin{algorithm}[t]
\DontPrintSemicolon
\SetAlgoLined
\SetKwInOut{Input}{Input}
\SetKwInOut{Output}{Output}
\SetKwFunction{FMain}{MAIN}
\SetKwFunction{FLocalUpdate}{LOCAL\_UPDATE}
\SetKwProg{Fn}{Function}{:}{}

\Input{ initial model $\theta_0$, number of rounds $R$, number of clients $W$, sampling probability $q$, noise scale $z$, clipping parameter $S$, optimizer OPT}
\Output{ trained model $\theta^{R+1}_{\text{global}}$}
\Fn{\FMain{$\theta_0$, $R$, $W$, $q$, $z$, $S$, OPT}}{
    $\theta^1_{\text{global}} \leftarrow \theta_0$\;
    $\sigma \leftarrow \frac{zS}{qW}$\;
    \For{round $r \leftarrow 1$ \KwTo $R$}{
        $P_r \leftarrow$ randomly select clients with probability $q$\;
        \For{client $k \in P_r$}{
            $\Delta^{r+1}_k \leftarrow$ \FLocalUpdate{$k$, $\theta^r_{\text{global}}$, $S$, OPT}\;
        }
        $\theta^{r+1}_{\text{global}} \leftarrow \theta^r_{\text{global}} + \frac{1}{qW} \sum_{k \in P_r} \Delta^{r+1}_k + \mathcal{N}(0, \sigma^2 I)$\;
    }
    \KwRet{$\theta^{R+1}_{\text{global}}$}\;
}
\Fn{\FLocalUpdate{$k$, $\theta^r_{\text{global}}$, $S$, OPT}}{
    $\theta \leftarrow \theta^r_{\text{global}}$\;
    \For{local epoch $i \leftarrow 1$ \KwTo $E$}{
        $\theta \leftarrow \text{OPT}(\theta, D_k)$ \tcp*{local update with OPT}
        $\Delta \leftarrow \theta - \theta^r_{\text{global}}$\;
        $\theta \leftarrow \theta^r_{\text{global}} + \min(1, \frac{S}{\| \Delta \|_2}) \cdot \Delta$ \tcp*{already clipped}
    }
    \KwRet{$\theta - \theta^r_{\text{global}}$}\;
}
\caption{DP-FedAvg Algorithm}
\label{Alg:dpfedavg}
\end{algorithm}

\section{Attacks against Federated Learning}
Prior work shows that FL may be vulnerable to attacks during and after the learning phase, targeting robustness and privacy.
Especially for sensitive datasets such as Personal Information, Medical Records, Financial Reports,etc~\cite{Bouacidea2021VulnerabilitiesFL, mamun2023deepmem}. %here

\begin{figure}[t]
    \centering
    \includegraphics[width=.9\textwidth]{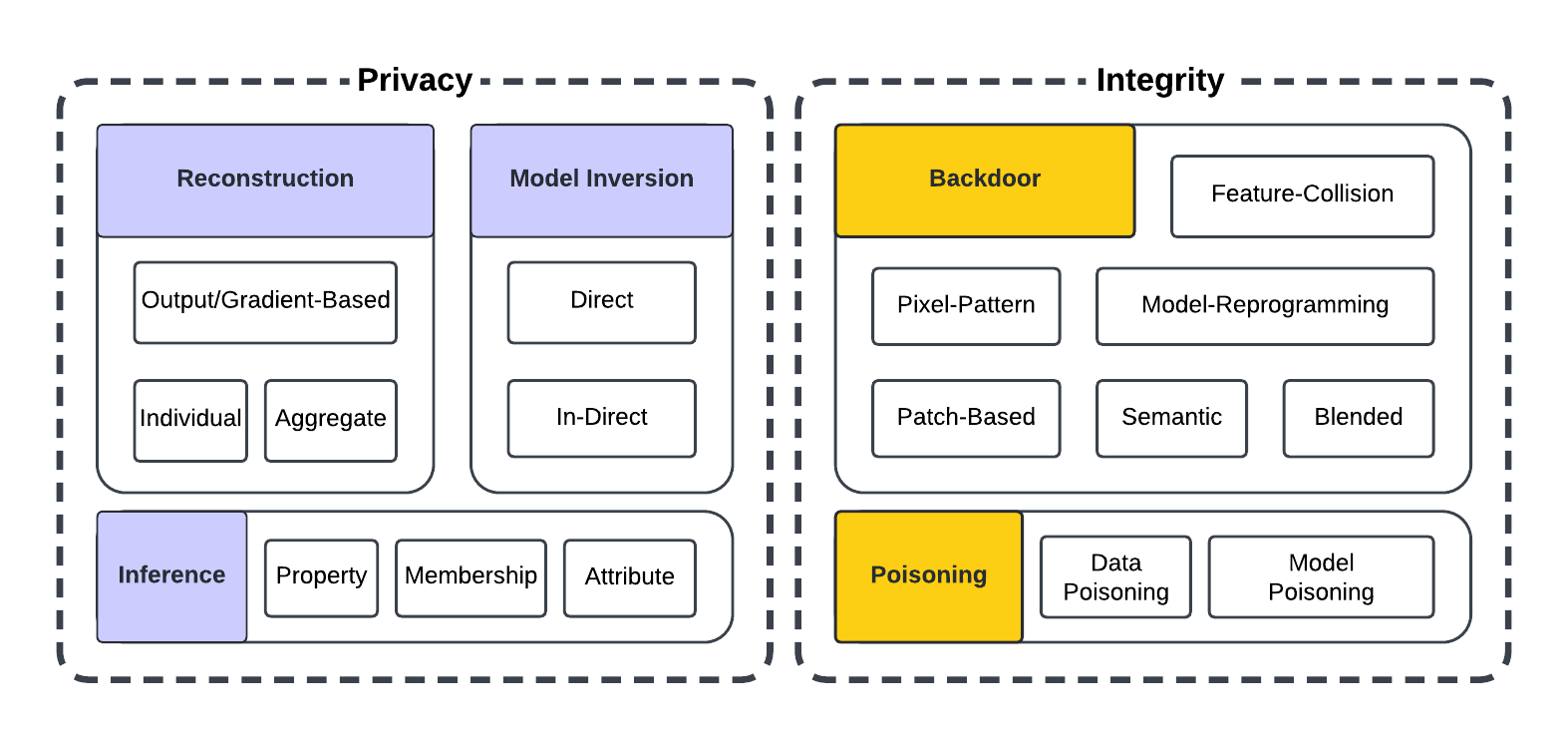}
    \caption{Overview of Different Types Common Attacks and Defences in Federated Learning and Generative Models}
    \label{fig:fl-attacks}
\end{figure}

We can split the type of attacks into two main groups: (i) privacy (e.g., Reconstruction, Inference Attacks, Model Inversion) and (ii) integrity (e.g., Backdoor, Model Poisoning) attacks. Here, we list the common attacks that happen in Federated Learning and Generative Models~\cite{sikandar2023detailed} based on privacy and integrity, as shown in Figure \ref{fig:fl-attacks}:

\subsection{Privacy}
\subsubsection{\textbf{Inference Attacks}} 
Although the raw user data does not leave the local device, many ways exist to infer the training data used in FL. These aim to exploit model updates exchanged between the clients and the central server to extract information about training data points. The goal is to infer properties of these points that may be even uncorrelated with the main task or training set membership by these updates.

These updates might accidentally give away more information than intended. We have different inference attacks, such as Property Inference, Attribute Inference, and Membership Inference~\cite{nasr2019comprehensive, gu2022cs}.

\paragraph{Membership Inference Attacks} Membership Inference attacks try to discover whether a data sample belongs to the training data~\cite{nasr2019comprehensive}. For instance, a membership inference attack could reveal whether data from a specific patient was used in training the model for predicting Alzheimer’s disease. If an attacker can find that a particular patient's data was included in the model's training dataset, it could use that the patient is either at risk of or is already diagnosed with Alzheimer's. This information could lead to privacy violations against the patient.

\paragraph{Attribute Inference Attack} 
An attribute inference attack seeks to uncover specific attributes or confidential information about individuals in a dataset. For example, consider a dataset with anonymized user data featuring details such as age, location, and browsing habits. An attacker might analyze observable patterns in this data to infer private attributes like political preferences or health conditions. The primary objective of such an attack is to predict sensitive information about individuals using the data that is available~\cite{gong2018attribute}.

\paragraph{Property Inference Attacks} Gradient exchange leakage can infer when a property appears and disappears in the dataset during training (e.g., identifying when a person first appears in the photos used to train a face recognition classifier). Property inference attacks assume that the adversary has auxiliary training data correctly labeled with the property he wants to infer~\cite{nasr2019comprehensive, gu2022cs}.

There is a slight difference between Attribute and Membership inference. Attribute inference attacks target individual records' attributes, whereas property inference attacks target collective properties of datasets or characteristics inherent to models.

\subsubsection{\textbf{Reconstruction Attacks}}
Where an adversary attempts to reconstruct clients' private data from shared model updates, an attacker uses inference techniques and statistical analysis to get properties of the original data from these updates (e.g., gradients carry substantial information about the input data and output labels). An attacker can potentially reconstruct the original input data by analyzing the pattern and nature of these gradients or model updates. This might involve making guesses about the data that would most likely produce such gradients ~\cite{yang2022using, lyu2021novel}. 
% The concept of differential privacy in FL, which aims to protect personal data by adding noise to the data or the model updates, came about because of concerns over these reconstruction attacks~\cite{bhowmick2019protection}.

\subsubsection{\textbf{Model Inversion Attacks}}
The core idea behind model inversion attacks is to take advantage of how machine learning models capture and store information about the training data. An attacker can derive the properties of the data on which it was trained by analyzing the model's behavior (e.g., how it responds to various inputs). Model Inversion Attack is a common attack with models with high memorization capacity and overly fitted to the training data~\cite{huang2021evaluating, hatamizadeh2023gradient}. For instance, by carefully tracking input queries and observing the model's outputs, an attacker can infer the characteristics of the data used to train the model. Also, in some cases, the attacker might reconstruct the data by the inferred characteristics of the data.

\subsection{Integrity}
\subsubsection{\textbf{Poisoning Attacks}}
These aim to make the target model misbehave; In general, we have two types of poisoning attacks: (i) Data Poisoning and (ii) Model Poisoning~\cite{tolpegin2020data}.

\textit{Data poisoning}: Attackers are messing with the training data of a machine learning model during the local data collection. An attacker adds misleading data to the ML database. As the model learns from this bad data, it starts making harmful decisions~\cite{naseri2022cerberus}. 

\textit{Model poisoning}: Attackers might change the local updates of the model during the model training, like changing the gradients, which can reduce the accuracy of the whole system~\cite{fang2020local}. 

Poisoning attacks can be random or targeted; random ones reduce the utility of the aggregated FL model, while targeted attacks make the aggregated FL model output predefined labels ~\cite{naseri2022cerberus, yang2023model}.

\subsubsection{\textbf{Backdoor Attacks}}
A subclass of poisoning attacks, namely backdoor attacks, has recently attracted much attention from the research community~\cite{naseri2022cerberus, bhagoji2019analyzing, shayegani2023survey}. Backdoor attacks are \textit{targeted} model poisoning attacks where a malicious client injects a backdoor task into the final model, typically using a \textit{model-replacement} methodology~\cite{naseri2022cerberus}. Like Poisoning Attacks, these attacks are tricky because they do not make the model wrong. Instead, they keep the model doing its main job right but sneak in secret weaknesses. In Federated Learning (FL), some clients that help train the model might be set up to teach these hidden backdoors using particular data. They adjust the model's performance based on their target while ensuring their effect remains unnoticed.

In Federated Learning, attackers can create these adjustments using models similar to the target. FL makes it easier for attackers because they can see how the model works and what makes it fail. A common trick is to change an image's pixels slightly so the system cannot recognize it correctly.

\section{Review Criteria}
\label{sec:criteria}
As mentioned in Section \ref{sec:model-review}, in this survey, our objective is to comprehensively compare existing papers that discuss the combination of Federated Learning and Generative Models with a primary focus on Federated Generative Models. We provide tables summarizing critical information from each paper based on various essential features. We have three main tables for those papers that propose new models: (i) GAN-based FL, (ii) Diffusion Model and VAE-based FL, and (iii) Other methods. Moreover, many papers consider security and privacy for generative models based on FL: (iv) Consider Privacy Attacks, and (v) Consider Integrity Attacks.

% define the purpose of each column of the table
To comprehensively review all related papers, each review criterion will hold important information, including details about generative models, where the papers were published, the federated algorithms used, types of datasets, and whether they support Differential Privacy. Additionally, these criteria include details about how the papers were evaluated, the number of clients considered, and if the code is available. This organized layout ensures that the table not only summarizes the key features of each paper but also helps in understanding how Federated Learning and Generative Models interact in various ways. In the following paragraph, we are going to elaborate on the criteria of our review:

\paragraph{Generative Model}
This section is a focal point of our survey, detailing each paper's specific generative approach, such as WGAN, DP-CGAN, ACGAN, Vanilla GAN, Synthetic Tabular Data, VAE, Diffusion Models, etc.

\paragraph{Federated Algorithm}
As mentioned earlier, various federated algorithms exist based on different aggregation approaches. Papers may focus on FedAVG, FedSGD, and MMD scores. We cover them in the second column to compare various models.

\paragraph{Data Type and Datasets}
Understanding the data type used in experiments is essential. While many papers concentrate on image datasets (e.g., MNIST, CIFAR10, FMNIST), a few articles explore tabular data (e.g., Adult, Cover-type, Credit), providing valuable diversity. 

\paragraph{Number of clients}
Considering the distributed dataset, knowing the number of clients in each paper's model is essential for assessing scalability and comparisons. For example, a model that only works with a maximum of 10 clients might not address a Cross-Devices setup where the data is distributed above 100 clients. We find the number of clients according to the experimental results of each work they reported. Many papers focus on analyzing the scalability; others only show their models' performance on a fixed number of clients.

\paragraph{Differential Privacy}
As mentioned in Section \ref{def:DP}, Differential Privacy is a privacy-preserving definition. We explore whether papers satisfy Differential Privacy in their architectures or not.

\paragraph{Privacy/Integrity Attacks Evaluation}
Federated Learning is a method used to protect data. If we introduce or modify a model that uses Federated Learning, we have to make a lot of effort to check its privacy and robustness. A paper in this area needs to evaluate privacy (e.g., Reconstruction, Inference Attacks, Model Inversion) or integrity (e.g., Backdoor, Model Poisoning) attacks.

\paragraph{Non-IID Support}
As we mentioned earlier, there are two different settings for FL problems: (i) Non-IID and (ii) IID settings. Handling Non-IID settings in Federated Learning is one of the most important details in every model. Many papers modify models that only support the IID setting to accommodate Non-IID settings.

\paragraph{Code Availability}
Code availability is noted, facilitating the re-running and modification of experiments and distinguishing papers that only provide algorithms.

\section{Models Review}
\label{sec:model-review}
% introduction
% As for all other deep neural networks, DGMs require a large training dataset to fit the target application. In recent years, distributed machine learning methods were proposed where the data is stored at several data centers and learned in a Federated way. Many prior works have tried to bring generative models into Federated Deep Learning setups rather than collect large amounts of data into a central server such as MD-Gan~\cite{mdgan2019hardy}, Private FL-GAN~\cite{privateFLGAN2020Xin}, PerFed-GAN~\cite{perfedGan2022cao}, GDTS~\cite{zhao2023gdts}, Fed-TGAN~\cite{zhao2021fed}, and more. Section \ref{sec:model-review} tries to elaborate on different Federated Generative Models proposed in recent years. 

The main reason for bringing generative models into FL was to protect individual data by preventing sensitive data sharing and communication overhead by keeping it in their devices. On the other hand, some works focus on using generative models independently in client or server to utilize the applications of generative models such as anomaly detection, addressing Non-IID and data heterogeneity challenges, inference attacks, and protection against reconstruction attacks.

In summary, we can group the research that combines generative models and Federated Learning into (i) Client-Generator, where synthetic data is generated locally and then communicated to the server for the server model updating; (ii) Server-Generator, where a centralized generator is used to assist in updating the server and client model, (iii) Federated-Generator, both clients and server working together to generate new dataset. However, our primary focus will be on the \textit{Federated-Generator} category.

In this section, we will review papers that discuss the combination of Federated Models with generative models, organizing them into three sub-sections. These are (i) those focusing on Generative Adversarial Networks (GANs) based FL, (ii) those focusing on Diffusion Models and Variational Autoencoders (VAEs) based FL, and (iii) those that do not fit within the first two categories.

% GAN done
\subsection{Generative Adversarial Networks based FL}
As summarized in Table \ref{tab:overview-models-GAN}, this section reviews the contributions of Federated GANs. Since the majority of papers focus exclusively on this area, we have dedicated a single section to review all related papers that integrate GANs with FL.

\paragraph{\textbf{How did it start?}}
MD-GAN \cite{mdgan2019hardy}, or Multi-Discriminator GAN, introduced the concept of using GANs in distributed systems. This model is characterized by a unique architecture that includes a single generator on the server side, multiple discriminators distributed across client sides, and a peer-to-peer communication pattern. Additionally, the performance of MD-GAN was evaluated against an adapted model, FL-GAN, which implements GANs on both the clients and the server. 

\paragraph{\textbf{Guarantees some intermediate Privacy}}
DP-FedAVG-GAN \cite{augenstein2019generative} show the efficacy of employing GANs in a Federated Learning context with user-level Differential Privacy (DP) guarantees. DP-FedAVG-GAN has effectively addressed many data issues when the data cannot be directly inspected. 
\cite{privateFLGAN2020Xin} introduce a privacy-centric approach to Federated Learning GANs (Private FL-GANs) by using Differential Privacy in the training process. Their method involves serialized training sessions among clients, where noise is added during training to ensure privacy protection. Specifically, they apply this technique within the Wasserstein GANs (WGAN) framework to generate synthetic data, enhancing data privacy without losing the quality of the generated data.

\cite{perfedGan2022cao} demonstrate the effect of DP-CGAN (Differentially Private Conditional GAN) on each client independently, focusing on aggregating sampled data on the server side before redistributing it back to each client.% redistributing: send them back to each client
\cite{behera2022fedsyn} create a model capable of generating synthetic data in a federated setup that reflects the data distribution of all clients using a DP GAN. This is particularly beneficial for clients with limited data resources.
% \cite{Wu2022FedCG} propose a mechanism that utilizes GANs for protecting clients by sharing only the generators within the server against DLG attack. It shares clients’ generators in the place of extractors with the server to aggregate clients’ shared knowledge, aiming to enhance model
% performance.

Additionally, \cite{pejic2022effect} explores using homomorphic encryption to train FL-GANs, achieving a fully privacy-preserving model. This approach allows data encryption during training, ensuring that sensitive information remains secure while still having practical model training. \cite{fedganids2022tabassum} explore using FedGAN to protect IoT and edge devices (as clients) by employing one global and many local GANs. This approach ensures the secure sharing of local gradients and global updates to enhance the integrity of IDS.

\paragraph{\textbf{Reduce Communication Overhead}}
\cite{zhang2021dance} proposed a distributed learning architecture for GANs that adapts communication compression in response to available bandwidth and supports data and model parallelism in GAN training. This architecture uses the gossip mechanism and Stackelberg game to introduce AC-GAN, a model that enhances training effectiveness while reducing communication overhead compared to FL-GAN and MD-GAN. In other words, AC-GAN extends the MD-GAN into a multi-server-multi-device framework, strategically deploying GANs at the network's edge. This approach effectively reduces the reliance on a single server\cite{zhang2022flschemeGAN}.
Additionally, \cite{zhang2022flschemeGAN} proposes a Collaborative Game Parallel Learning (CAP) strategy, where each client operates as a separate information island, using a synthesis score for feedback aggregation and a Mix-Generator to address fully non-IID sources.

\cite{wijesinghe2023ufed} introduce an unsupervised federated GAN, UFed-GAN, which adeptly captures the data distribution at the user level without necessitating local classification training. 
\cite{ekblom2022effgan} addresses client drift by generating data from an ensemble of locally fine-tuned GANs, providing a solution to one of the key challenges in FL.

% non-IID
\paragraph{\textbf{Combine GANs with FL to address non-IID and data heterogeneity challenges}}
The idea of combining GANs with FL to address non-independent and identically distributed (non-IID) data challenges is extensive and varied. Several innovative approaches have been proposed.

FedGAN \cite{fedgan2020Rasouli} is one such paper that extends the application of GANs to FL settings, particularly when data across different sources are non-IID. FedGAN introduces an architecture that employs local generators and discriminators, which are periodically synchronized through an intermediary. This intermediary is responsible for averaging and sharing the parameters of the generator and discriminator, ensuring the model's convergence, especially in non-IID scenarios. 
FeGAN \cite{guerraoui2020fegan}, another GAN-based FL, innovates by determining the aggregation weight to address non-IID issues in distributed GAN setups. It uses the Kullback-Leibler distance between the local and global label categories to fine-tune the aggregation process. However, both FedGAN and FeGAN have been noted to require significant energy and computational resources.

An additional noteworthy contribution is from Improved FL-GAN \cite{IFLGAN2023Li}, which proposes using Maximum Mean Discrepancy (MMD) for aggregation instead of the weight or parameter averaging methods to address non-IID sources. 

\cite{zhang2021training} introduces a novel approach that simulates a centralized discriminator by aggregating a mixture of all private discriminators, offering a unique solution to the challenges of Federated Learning, named Universal Aggregation. \cite{wijesinghe2023pfl} proposes a sharing and aggregation strategy using one local conditional GAN (cGAN) at each client and two cGANs on the server, addressing client heterogeneity in GAN-based FL.

% tabular
\paragraph{\textbf{Using GAN-based FL for Tabular Data}}
While the main focus of the research in this area is addressing image datasets, there is a growing interest in exploring other data types, such as tabular and time series data. ~\cite{chiaro2023flenhance} is one of the papers focusing on image and tabular non-IID data. They enhance the non-IID issue using established solutions like data selection, compression, and augmentation. In other words, they use CGANs trained on the server level. 

GDTS~\cite{zhao2023gdts} and Fed-TGAN~\cite{zhao2021fed} are two other papers focusing only on Tabular data. They introduce two distinct approaches: (i) deploying one GAN for each client and (ii) adapting the MD-GAN framework specifically for tabular data. They also show that the first version outperforms the second version. ~\cite{zhao2023gtv} propose a method for generating tabular data with Vertical Federated Learning using GANs with a training-with-shuffling mechanism. They evaluate their mechanism in synthetic data quality and training scalability.

\cite{maliakel2024fligan} explores the capabilities of GANs in processing tabular data within a federated setup. Their method focuses on privacy preservation, synthetic data generation, and handling incomplete data simultaneously, which outperforms all other Tabular-based GANs. To achieve their objectives, they utilize a Federated Wasserstein GAN (WGAN) with class-wise sampling and node grouping.

\paragraph{\textbf{Using GAN-based FL for Time Series data and Anomaly Detection}}
% anomaly detection in time-series
~\cite{chen2023fedlgan} propose a method for anomaly detection and repair of hydrological telemetry data using Federated Learning, GANs, and utilizing the advantages of long short-term memory (LSTM). 
% The first paper discusses data repair and anomaly detection in time series with GAN-based FL.

\paragraph{\textbf{Using GAN-based FL for Clinical Application}}
\cite{rehman2024fedcscd} explore the use of FL-based GAN for cancer diagnosis, emphasizing the accuracy of diagnosis while protecting patient confidentiality. This approach combines DP and Quasi Identifiers within Federated Analytics to improve data confidentiality.

~\cite{zhang2021feddpgan, nguyen2021federated} are proposed for COVID-19 recognition. ~\cite{nguyen2021federated} combined the FedGAN design with Blockchain networks on cloud devices for COVID-19 detection. 
FedDPGAN~\cite{zhang2021feddpgan} tried to resolve the non-IID issue using data augmentation and their methods for training COVID-19 models. They used a distributed DPGAN by using the FL framework.
~\cite{WANG2023FedMedGAN} implement a new benchmark for cross-modality brain image synthesis in a federated setup and greatly facilitate the development of medical GAN with DP guarantees.

% GAN models:
\begin{table*}[t]
\centering
\setlength\tabcolsep{2pt}
\caption{Federated GANs Summary}
\label{tab:overview-models-GAN}
\resizebox{\textwidth}{!}{%
\begin{tabular}{|c|c|c|c|c|c|c|c|c|c|c|}
\hline
\textbf{Paper} &
  \textbf{Model} &
  \textbf{Federate Algorithm} &
  \textbf{Data Type} &
  \textbf{Datasets} &
  \textbf{DP} &
  \textbf{\#Clients} &
  \textbf{non-IID} &
  \textbf{Code} \\ \hline
\textbf{
    \begin{tabular}[c]{@{}c@{}}
Private FL-GAN\\~\cite{privateFLGAN2020Xin}
    \end{tabular}
} &
  WGAN &
  FedAVG &
  Image &
  \begin{tabular}[c]{@{}c@{}}MNIST, CelebA\end{tabular} &
  \cmark &
  $\sim$20 &
  \xmark &
  \xmark \\  \hline
\textbf{
    \begin{tabular}[c]{@{}c@{}}
PerFED-GAN\\~\cite{perfedGan2022cao}
    \end{tabular}
    } &
  DP-CCGAN &
  \xmark &
  Image &
  \begin{tabular}[c]{@{}c@{}}CIFAR10, CIFAR100, FMNIST\end{tabular} &
  \cmark &
  $\sim$100 &
  \cmark &
  \xmark \\ \hline
\textbf{
    \begin{tabular}[c]{@{}c@{}}
IFL-GAN\\~\cite{IFLGAN2023Li}
    \end{tabular}
} &
  \begin{tabular}[c]{@{}c@{}}DCGAN\end{tabular} &
  MMD Score &
  Image &
  \begin{tabular}[c]{@{}c@{}}CIFAR10, MNIST, SVHN\end{tabular} &
  \cmark &
  $\sim$10 &
  \cmark &
  \xmark \\ \hline
\textbf{
    \begin{tabular}[c]{@{}c@{}}
MD-GAN\\~\cite{mdgan2019hardy}
    \end{tabular}
} &
  GAN &
  Separate G/D &
  Image &
  \begin{tabular}[c]{@{}c@{}}CIFAR10, MNIST, CelebA\end{tabular} &
  \xmark &
  $\sim$50 &
  \xmark &
  \xmark \\ \hline
\textbf{
    \begin{tabular}[c]{@{}c@{}}
FedGAN\\~\cite{fedgan2020Rasouli}
    \end{tabular}
} &
  CGAN &
  FedAVG &
  Image &
  \begin{tabular}[c]{@{}c@{}}Toy Examples, MNIST\\ CIFAR10, CelebA\end{tabular} &
  \xmark &
  $\sim$10 &
  \cmark &
  \xmark \\ \hline
\textbf{
    \begin{tabular}[c]{@{}c@{}}
FedDPGAN\\~\cite{zhang2021feddpgan}
    \end{tabular}
}
 &
  DP-GAN &
  FedAVG &
  Image &
  COVID-19 &
  \cmark &
  $\sim$100 &
  \cmark &
  \xmark \\ \hline
\textbf{
 \begin{tabular}[c]{@{}c@{}}
Fed-TGAN\\~\cite{zhao2021fed}
    \end{tabular}
} &
  GAN &
  Adopt FL-GAN  &
  Tabular &
  \begin{tabular}[c]{@{}c@{}}Adult, Covertype\\ Credit, Intrusion \end{tabular}&
  \cmark &
  $\sim$10 &
  \cmark &
  \cmark \\ \hline
\textbf{
    \begin{tabular}[c]{@{}c@{}}
GDTS\\~\cite{zhao2023gdts}
    \end{tabular}
} &
  GAN &
  FedAVG &
  Tabular &
  \begin{tabular}[c]{@{}c@{}}Adult, Covertype\\ Credit, Intrusion \end{tabular}&
  \cmark &
  $\sim$10 &
  \cmark &
  \xmark \\ \hline
\textbf{
    \begin{tabular}[c]{@{}c@{}}
GTV\\~\cite{zhao2023gtv}
    \end{tabular}
} &
  \begin{tabular}[c]{@{}c@{}}CT-GAN\\ CTAB-GAN\end{tabular} &
  \begin{tabular}[c]{@{}c@{}}VFL\end{tabular} &
  Tabular &
  \begin{tabular}[c]{@{}c@{}}Adult, Covtype, Loan\\ Intrusion, Credit\end{tabular} &
  \xmark &
  2-5 &
  \cmark &
  \xmark \\ \hline
\textbf{
    \begin{tabular}[c]{@{}c@{}}
FL-Enhance\\~\cite{chiaro2023flenhance}
    \end{tabular}
} &
  CGAN &
  \begin{tabular}[c]{@{}c@{}}FedAVG, Fednova, FedOPT\end{tabular} &
  \begin{tabular}[c]{@{}c@{}}Image\\ Tabular\end{tabular} &
  \begin{tabular}[c]{@{}c@{}}MNIST, FMNIST\\ CA Housing, Beijing PM2.5\end{tabular} &
  \xmark &
  $\sim$100 &
  \cmark &
  \xmark \\ \hline
\textbf{
    \begin{tabular}[c]{@{}c@{}}
FeGAN\\~\cite{guerraoui2020fegan}
    \end{tabular}
} &
  CGAN &
  \begin{tabular}[c]{@{}c@{}}FedSGD\end{tabular} &
  Image &
  \begin{tabular}[c]{@{}c@{}}MNIST, FMNIST\\ ImageNET\end{tabular} &
  \xmark &
  $\sim$128 &
  \cmark &
  \xmark \\ \hline
\textbf{
    \begin{tabular}[c]{@{}c@{}}
UFed-GAN\\ \cite{wijesinghe2023ufed}
    \end{tabular}
} &
  DCGAN &
  Share Gradients Update &
  Image &
  \begin{tabular}[c]{@{}c@{}}FMNIST, SVHN\\ CIFAR10\end{tabular} &
  \cmark &
  $\sim$30 &
  \cmark &
  \xmark \\ \hline
\textbf{
    \begin{tabular}[c]{@{}c@{}}
        DANCE (AC-GAN) \\ ~\cite{behera2022fedsyn}
    \end{tabular}} &
  GAN &
  \begin{tabular}[c]{@{}c@{}}FedSGD\end{tabular} &
  Image &
  \begin{tabular}[c]{@{}c@{}}MNIST, FMNIST\\ CIFAR10\end{tabular} &
  \xmark &
  $\sim$10 &
  \xmark &
  \xmark \\ \hline
\textbf{
\begin{tabular}[c]{@{}c@{}}
        Universal Aggregation \\ ~\cite{zhang2021training}
    \end{tabular}
} &
  CGAN &
  Discriminator Aggregation &
  Image &
  \begin{tabular}[c]{@{}c@{}}MNIST, FMNIST\end{tabular} &
  \xmark &
  $\sim$10 &
  \cmark &
  \cmark \\ \hline
\textbf{\begin{tabular}[c]{@{}c@{}}
        FedSyn\\~\cite{behera2022fedsyn}
    \end{tabular}} &
  GAN &
  \begin{tabular}[c]{@{}c@{}}Average Noisy Parameters\\ HFL\end{tabular} &
  \begin{tabular}[c]{@{}c@{}}Image\end{tabular} &
  \begin{tabular}[c]{@{}c@{}}MNIST, CIFAR10\end{tabular} &
  \cmark &
  3 &
  \cmark &
  \xmark \\ \hline
\textbf{
    \begin{tabular}[c]{@{}c@{}}
        FedCSCD-GAN\\~\cite{rehman2024fedcscd}
    \end{tabular}
} &
  Cramer GAN &
  Secure Aggregation &
  Image &
  \begin{tabular}[c]{@{}c@{}}Prostate/Lung/Breast Cancer\end{tabular} &
  \cmark &
  5 &
  \xmark &
  \xmark \\ \hline
\textbf{
    \begin{tabular}[c]{@{}c@{}}
        DP-FedAvg-GAN\\~\cite{augenstein2019generative}
    \end{tabular}
} &
  GAN &
  FedAVG &
  Image &
  \begin{tabular}[c]{@{}c@{}}MNIST, EMNIST\end{tabular} &
  \cmark &
  10 &
  \xmark &
  \cmark \\ \hline
\textbf{
    \begin{tabular}[c]{@{}c@{}}
        FedLGAN\\~\cite{chen2023fedlgan}
    \end{tabular}
} &
  LSTM + GAN &
  FedAvg &
  Time Series &
  Hydrological Data &
  \xmark &
  \xmark &
  \cmark &
  \cmark \\ \hline

  \textbf{
    \begin{tabular}[c]{@{}c@{}}
        FL-GAN + HE\\~\cite{pejic2022effect}
    \end{tabular}
}&
  GAN &
  FedAvg &
  Image &
  CIFAR10 &
  \xmark&
  $\sim$3 &
  \xmark &
  \xmark\\ \hline

\textbf{
    \begin{tabular}[c]{@{}c@{}}
        FLIGAN\\~\cite{maliakel2024fligan}
    \end{tabular}
}&
  GAN &
  FedAVG &
  Tabular &
  \begin{tabular}[c]{@{}c@{}}Albert, Bank, Adult\\
Creditcard, Intrusion\end{tabular} &
  \xmark&
  4-8 &
  \cmark &
  \xmark\\ \hline

\textbf{
    \begin{tabular}[c]{@{}c@{}}
        FedMed-GAN\\~\cite{WANG2023FedMedGAN}
    \end{tabular}
}&
  GAN &
  FedAVG &
  Image &
  \begin{tabular}[c]{@{}c@{}}IXI, BRATS2021\end{tabular} &
  \cmark&
  2,4,8 &
  \cmark &
  \cmark\\ \hline 

    \textbf{
    \begin{tabular}[c]{@{}c@{}}
        EFFGAN\\~\cite{ekblom2022effgan}
    \end{tabular}
} &
\begin{tabular}[c]{@{}c@{}}
    DCGAN \\ Ensemble Model
\end{tabular} &
  \begin{tabular}[c]{@{}c@{}}FedAVG\\ Fine-Tuning\end{tabular} &
  Image &
  \begin{tabular}[c]{@{}c@{}}MNIST, FMNIST\end{tabular} &
  \xmark &
  100 &
  \cmark &
  \xmark\\ \hline
\textbf{~\cite{nguyen2021federated}} &
  GAN &
  Block-Chain Based &
  Image &
  \begin{tabular}[c]{@{}c@{}}COVID-19 X-ray\\ DarkCOVID\end{tabular} &
  \cmark &
  5 &
  \cmark &
  \cmark \\ \hline
  
  \textbf{
    \begin{tabular}[c]{@{}c@{}}
        PFL-GAN\\~\cite{wijesinghe2023pfl}
    \end{tabular}
}&
  GAN &
  Weighted Collaborative Aggregation &
  Image &
  \begin{tabular}[c]{@{}c@{}}MNIST, FMNIST, EMNIST\end{tabular} &
  \xmark&
  10, 20 &
  \cmark &
  \xmark\\ \hline
\end{tabular}%
}
\end{table*}

% ddpm and VAE done:
\subsection{Diffusion Models and VAEs based FL}
As summarized in Table \ref{tab:overview-models-VAE}, this section reviews contributions to combine VAEs/Diffusion Models with FL, highlighting advancements in addressing data heterogeneity and non-iid challenges and enhancing recommender systems.

%VAE:
\paragraph{\textbf{VAEs-based FL}}
\cite{pfitzner2022dpdfvae} introduce a VAE-based FL with a differentially private decoder to synthesize labeled datasets while synchronizing the decoder component with FL. Their method is evaluated in both private and non-private FL settings.
\cite{dugdale2023federated} focuses on employing FL with VAEs to generate handwritten digits, specifically trained on the MNIST dataset.

%Non-IID
\paragraph{\textbf{Combine VAEs with FL to Address Non-IID and data Heterogeneity challenges}}
% \cite{duan2021feddna} use VAEs in FedDNA to aggregate gradient parameters using federated averaging and statistical parameters using a weighting method. This approach aims to reduce divergence and optimize collaborative learning by addressing cross-model distribution covariate shift (CDCS). 
% \cite{duan2023federated} proposes using VAEs to cluster heterogeneous client data distributions by learning optimal parameters for virtual fusion components without risking privacy leakage.
\cite{chen2023federated} enrich client datasets with DP synthetic data using VAEs and adding Gaussian noise and filtering mechanisms for efficient data synthesis to reduce the effects of Non-IID data distributions.
\cite{chen2023label} introduce cVAE with label augmentation at the client side to reduce divergence in label-wise distribution adaptive FL.

\paragraph{\textbf{Using VAE-based FL for Time Series, Tabular, and Trajectory data}}

\cite{kaspour2023variational} propose a VAE model for non-intrusive load monitoring (NILM) in smart homes. This is the first paper that works on energy time-series data by using VAE in a federated setup. 
\cite{jiang2023fedvae} focus on using Federated VAEs with DP for decentralized trajectory data analysis.
\cite{lomurno2022sgde} introduces a VAE-based method for tabular data in FL by sharing the generator for aggregation.

% Diffusion Models
\paragraph{\textbf{Diffusion Models based FL}}
~\cite{de2023training} adapt the FedAvg algorithm to train a Denoising Diffusion Model (DDPM).
\cite{jothiraj2023phoenix} introduce training Denoising Diffusion Probabilistic Models across distributed data sources using FL.
\cite{li2023synthetic} propose DDPM as a local generator for synthetic data shuffling to improve FL convergence under data heterogeneity. They also compare the efficacy of using $\beta$-VAE as an alternative generative model.
~\cite{tun2023fedDM} combine Diffusion Models and FL for privacy-sensitive data in vision-related tasks. ~\cite{sattarov2024fedtabdiff} is the only recent study on FL-based Denoising Diffusion Probabilistic Models (DDPM) to address the inherent complexities in tabular data, such as mixed attributes, implicit relationships, and distribution imbalance.

\paragraph{\textbf{Combine Diffusion Models with FL to Address Non-IID and Data Heterogeneity Challenges}}
\cite{ahn2023communicationefficient} propose FedDif, a communication-efficient strategy for improving FL performance with Non-IID clients by reducing weight divergence. They designed the diffusion strategy based on auction theory and provided both theoretical and empirical analyses of FedDif to show its success in mitigating weight divergence.

\paragraph{\textbf{One-Shot Federated Learning}}
One-shot Federated Learning (OSFL) has recently gained attention due to its low communication cost, leading many researchers to explore its design advantages. 
~\cite{yang2023one} introduce Diffusion Models to OSFL, proposing a new large-scale mechanism to use client guidance for generating data that complies with clients' distributions. 
~\cite{heinbaugh2023datafree} address the challenge of high statistical heterogeneity in one-shot FL using VAEs, showing the performance of their mechanism with varying numbers of clients. They propose two methods that ensemble the decoders from clients: (i) FedCVAE-Ens and (ii) FedCVAE-KD, with the latter also using knowledge distillation to compress the ensemble of client decoders into a single decoder. 
~\cite{yang2024exploring} addresses the semi-FL problem with Non-IID clients by introducing Diffusion Models into semi-FL for one-shot semi-supervised FL.

% VAE and DDPM:
\begin{table*}[]
\centering
\caption{Federated VAEs and Diffusion Models Summary}
\label{tab:overview-models-VAE}
\setlength\tabcolsep{2pt}
\resizebox{\textwidth}{!}{%
\begin{tabular}{|c|c|c|c|c|c|c|c|c|c|c|}
\hline
\textbf{Paper} &
  \textbf{Model} &
  \textbf{Federate Algorithm} &
  \textbf{Data Type} &
  \textbf{Datasets} &
  \textbf{DP} &
  \textbf{\#Clients} &
  \textbf{Non-IID} &
  \textbf{Code} \\ \hline
\textbf{
    \begin{tabular}[c]{@{}c@{}}
        DPD-fVAE\\
        ~\cite{pfitzner2022dpdfvae}
    \end{tabular}
} &
  VAE &
  \begin{tabular}[c]{@{}c@{}}Average of Probabilities\\ Average of Decoder Weights\end{tabular} &
  Image &
  \begin{tabular}[c]{@{}c@{}}MNIST, FMNIST, CelebA\end{tabular} &
  \cmark &
  $\sim$100-10,000 &
  \xmark &
  \cmark \\ \hline
\textbf{
    \begin{tabular}[c]{@{}c@{}}
        FedDPMS\\
        ~\cite{chen2023federated}
    \end{tabular}
} &
  VAE &
  \begin{tabular}[c]{@{}c@{}}Uploading Usable Noisy Latent Means\end{tabular} &
  Image &
  \begin{tabular}[c]{@{}c@{}}CIFAR10, CIFAR100\\ FMNIST\end{tabular} &
  \cmark &
  $\sim$10-50 &
  \cmark &
  \xmark \\ \hline

\textbf{
    \begin{tabular}[c]{@{}c@{}}
        FedVAE\\
        ~\cite{jiang2023fedvae}
    \end{tabular}
} &
  VAE &
  \begin{tabular}[c]{@{}c@{}}FedAVG\\ Joint-Announcement Protocol\end{tabular} &
  Trajectory &
  Microsoft’s Geolife &
  \xmark &
  70 &
  \xmark &
  \xmark\\ \hline

\textbf{
    \begin{tabular}[c]{@{}c@{}}
        LEDA-FL\\
        ~\cite{chen2023label}
    \end{tabular}
} &
  CVAE &
  \begin{tabular}[c]{@{}c@{}}Augment the Label-wise Features\\ Interaction Aggregation\end{tabular} &
  Image &
  CIFAR100 &
  \xmark &
  50 &
  \cmark &
  \xmark\\ \hline
\textbf{
    \begin{tabular}[c]{@{}c@{}}
        ~\cite{heinbaugh2023datafree}
    \end{tabular}
    } &
  CVAE &
  \begin{tabular}[c]{@{}c@{}}
  One-shot FL\\
Knowledge Distillation\\
Decoder Aggregation\\\end{tabular} &
  \begin{tabular}[c]{@{}c@{}}Image\end{tabular} &
  \begin{tabular}[c]{@{}c@{}}MNIST, FMNIST, SVHN\end{tabular} &
  \cmark &
  5,10,20,50 &
  \cmark &
  \xmark \\ \hline
\textbf{~\cite{kaspour2023variational}} &
  \begin{tabular}[c]{@{}c@{}}
        VAE\\
        NILM
    \end{tabular} &
  FedAVG &
  Energy Time-Series &
  UK-DALE &
  \cmark &
  \xmark &
  \xmark &
  \xmark\\ \hline
\textbf{
    \begin{tabular}[c]{@{}c@{}}
        SGDE\\
        ~\cite{lomurno2022sgde}
    \end{tabular}
    } &
  $\beta$-VAE &
  \begin{tabular}[c]{@{}c@{}}Cross-Silo FL\\ Generator Aggregation\end{tabular} &
  \begin{tabular}[c]{@{}c@{}}Image\\ Tabular\end{tabular} &
  \begin{tabular}[c]{@{}c@{}}UCI ML Repo\\ MNIST, FMNIST\end{tabular} &
  \cmark &
  \textless{}100 &
  \cmark &
  \xmark \\ \hline
\textbf{~\cite{dugdale2023federated}} &
  $\beta$-VAE &
  FedAVG &
  Image &
  MNIST &
  \xmark &
  4 &
  \xmark&
  \cmark \\ \hline

\textbf{
    \begin{tabular}[c]{@{}c@{}}
        Phoenix\\
        ~\cite{jothiraj2023phoenix}
    \end{tabular}
} &
  Diffusion Model &
  Weight Average after 100 Local Epochs &
  Image &
  CIFAR10 &
  \xmark &
  10 &
  \cmark &
  \xmark \\ \hline
\textbf{
    \begin{tabular}[c]{@{}c@{}}
        FedDif\\
        ~\cite{ahn2023communicationefficient}
    \end{tabular}
    } &
  Diffusion Model &
  \begin{tabular}[c]{@{}c@{}}
  Cooperative FL\\ D2D Communication \end{tabular} &
  \begin{tabular}[c]{@{}c@{}}Image\end{tabular} &
  \begin{tabular}[c]{@{}c@{}}MNIST, FMNIST, CIFAR10\end{tabular} &
  \xmark &
  \xmark &
  \cmark &
  \cmark \\ \hline

\textbf{
    \begin{tabular}[c]{@{}c@{}}
        FedDM\\
        ~\cite{tun2023fedDM}
    \end{tabular}
    } &
  Diffusion Model &
  FedAvg &
  \begin{tabular}[c]{@{}c@{}}Image\end{tabular} &
  \begin{tabular}[c]{@{}c@{}}CIFAR10, MNIST, SVHN\\ SARS-CoV-2\end{tabular} &
  \xmark &
  10,30,50 &
  \xmark &
  \cmark \\ \hline

\textbf{
    \begin{tabular}[c]{@{}c@{}}
        ~\cite{de2023training}
    \end{tabular}
    } &
  Diffusion Model &
  \begin{tabular}[c]{@{}c@{}}
  FedAVG, Federated UNET \\ Cross-Silo FL \end{tabular} &
  \begin{tabular}[c]{@{}c@{}}Image\end{tabular} &
  \begin{tabular}[c]{@{}c@{}}CelebA, FMNIST\end{tabular} &
  \xmark &
  1-10 &
  \xmark &
  \xmark \\ \hline

\textbf{
    \begin{tabular}[c]{@{}c@{}}
        FedCADO\\
        ~\cite{yang2023one}
    \end{tabular}
    } &
  Diffusion Model &
  \begin{tabular}[c]{@{}c@{}}One-Shot FL\end{tabular}&
  \begin{tabular}[c]{@{}c@{}}Image\end{tabular} &
  \begin{tabular}[c]{@{}c@{}}OpenImage, NICO++\\ Domain-Net\end{tabular} &
  \xmark &
  20 &
  \cmark &
  \cmark \\ \hline

  \textbf{
    \begin{tabular}[c]{@{}c@{}}
        FedDISC\\
        ~\cite{yang2024exploring}
    \end{tabular}
    } &
  Diffusion Model &
  \begin{tabular}[c]{@{}c@{}}One-Shot FL\\ FedAvg\end{tabular} &
  \begin{tabular}[c]{@{}c@{}}Image\end{tabular} &
  \begin{tabular}[c]{@{}c@{}}OpenImage, NICO++\\Domain-Net\end{tabular} &
  \xmark &
  5 &
  \cmark &
  \xmark \\ \hline
\textbf{
    \begin{tabular}[c]{@{}c@{}}
        FedTabDiff\\
        ~\cite{sattarov2024fedtabdiff}
    \end{tabular}
    } &
  Diffusion Model (DDPM) &
  \begin{tabular}[c]{@{}c@{}}
  Weighted Averaging \end{tabular} &
  \begin{tabular}[c]{@{}c@{}}Tabular\end{tabular} &
  \begin{tabular}[c]{@{}c@{}}Philadelphia City Payments\\
Diabetes Hospital\end{tabular} &
  \xmark &
  5 &
  \cmark &
  \cmark \\ \hline
\textbf{
    \begin{tabular}[c]{@{}c@{}}
        Fedssyn\\
        ~\cite{li2023synthetic}
    \end{tabular}
} &
  \begin{tabular}[c]{@{}c@{}}
  Diffusion Model (DDPM)\\ VAE
  \end{tabular} &
  \begin{tabular}[c]{@{}c@{}}FedAvg, FedProx, FedPVR\\ SCAFFOLD, FedDyn\end{tabular} &
  \begin{tabular}[c]{@{}c@{}}Image\\ dSprites\end{tabular} &
  \begin{tabular}[c]{@{}c@{}}CIFAR10, CIFAR100\end{tabular} &
  \cmark &
  10, 40, 100 &
  \cmark &
  \cmark \\ \hline
\end{tabular}%
}
\end{table*}

% others done:
\subsection{Other Application of FL and Generative Models}
As mentioned in Table \ref{tab:overview-models-ther}, we summarize a collection of studies that use different Federated Learning (FL) methods to address challenges not specifically aligned with the previously mentioned groups. These papers introduce approaches to data heterogeneity, Non-IID distributions, knowledge distillation, and data sharing to enhance the performance and efficiency of FL systems across various applications.

\paragraph{\textbf{Solving Data Heterogeneity and Non-IID}}
\cite{guo2023enhancefl} propose a data augmentation method to solve data heterogeneity in FL. This method to balance the Non-IID source effectively uses a generative model locally on the client side to synthesize data reflective of the distribution.
\cite{duan2022fedTDA} introduce a federated tabular data augmentation strategy designed for Non-IID data distributions. The approach includes federated Gaussian Mixture Models (GMM), Inverse Distance Matrix (IDM) frequency, and global covariance for aggregating global statistics.
~\cite{yoon2021fedmix} propose a simple Mean Augmented FL, where clients send and receive averaged augmented local data under highly Non-IID federated settings.

\cite{li2022federated} introduces a framework that handles the non-IID issue by sharing differentially private (DP) synthetic data. Each client pre-trains a local DP GAN to generate synthetic data in this setup. They are combined with a parameter server to construct synthetic datasets with a pseudo-label update mechanism. These data can be applied in supervised and semi-supervised learning settings.
\cite{ma2023FLGAN} use GANs to reduce local biases through synthesized samples for improving non-IID FL performance. This model uses Fully Homomorphic Encryption (FHE) to ensure privacy. 

\cite{duan2021feddna} use VAEs in FedDNA to aggregate gradient parameters using federated averaging and statistical parameters using a weighting method. This approach aims to reduce divergence and optimize collaborative learning by addressing cross-model distribution covariate shift (CDCS). 
\cite{duan2023federated} proposes using VAEs to cluster heterogeneous client data distributions by learning optimal parameters for virtual fusion components without risking privacy leakage.

\cite{zhao2023federated} introduce a Diffusion-based FL Model to handle Non-IID data. They use Diffusion Models for data augmentations to address Non-IID data.
~\cite{wang2023fda} introduces a data augmentation framework for handling Non-IID data using Diffusion Models on the central server.

\paragraph{\textbf{Knowledge, Feature, and Model Distillation}}
To improve the performance of the smaller dataset, new approaches include learning a synthetic set of original data or data distillation. Different methods are proposed, such as meta-learning, gradient matching, distribution matching, neural kernels, or generative models.  
\cite{zhu2021data} addresses the issue of data heterogeneity in FL through data-free knowledge distillation, also known as a teacher-student paradigm. Their strategy enhances privacy and efficiency since it involves the server learning a lightweight generator to ensemble user information without direct data access.
\cite{zhang2022dense} propose a two-stage, data-free, one-shot FL framework focused on data generation and model distillation. The initial stage trains a generator considering factors like similarity, stability, and transferability, followed by employing ensemble models and generated data to train the global model.

\cite{yang2024fedfed} introduce a VAE-based framework that addresses data heterogeneity by sharing partial features in the data. The framework also elaborates on methods to reduce the risks of model inversion and membership inference attacks.

\cite{Wu2022FedCG} propose a method that employs GANs to protect clients by exclusively sharing their generators with the server, thus strengthening against DLG attacks. This approach entails sharing clients' generators with the server instead of extractors to merge clients' collective knowledge to enhance model performance.

\paragraph{\textbf{Sharing (Synthetic) Data and Distribution Matching}}
\cite{xiong2023feddm} introduce a communication-efficient FL, an iterative distribution matching approach that learns a surrogate function by sending differentially private synthesized data to the server.
\cite{prediger2023collaborative} propose a collaborative framework in that every client creates synthetic twin data to share with the global server. Clients then decide whether to combine this synthetic data with their real data for model training.
\cite{zhou2023communicationefficient} introduce a new communication-efficient Federated Learning method with a single-step synthetic features compressor. Their method transmits only a tiny set of model inputs and labels.
% recommender systems:
\paragraph{\textbf{Using VAE-based FL for Recommender Systems}}
\cite{ding2023combining} use VAEs with AdaptiveDP in federated settings for privacy-safe recommendations. Their method also analyzes the privacy risks of VAE in federated collaborative filtering.
\cite{li2023distvae} proposes a Distributed VAE for a sequential recommendation that clusters clients into virtual groups for local model training and server aggregation by GMM.

\paragraph{\textbf{Federated Imputation of Incomplete Data}}
~\cite{FedTMI2022Yao} use a GAN at each client to address data imputation challenges within the Federated Transfer Learning problem. 
~\cite{Zhou2021Fed} design a GAN imputation mechanism under an FL framework to address datasets originating from various data owners without necessitating data sharing.
~\cite{balelli2023fed} propose a federated version of MIWAE as a deep latent variable model for missing data imputation with variational inference.

% VAE [not used]:
% \cite{mou2023pfedv} explore horizontal Federated Learning with supervised learning tasks, employing variational distribution constraints to reduce feature distribution skewness.

% \paragraph{\textbf{Federated Imputation of Incomplete Data}}

% ~\cite{balelli2023fed} propose a federated version of MIWAE as a deep latent variable model for missing data imputation with variational inference. 

% other models:
\begin{table*}[]
\centering
\caption{Other Uses of Generative Models and Federated Learning}
\label{tab:overview-models-ther}
\setlength\tabcolsep{2pt}
\resizebox{\textwidth}{!}{%
\begin{tabular}{|c|c|c|c|c|c|c|c|c|c|c|}
\hline
\textbf{Paper} &
  \textbf{Model} &
  \textbf{Federate Algorithm} &
  \textbf{Data Type} &
  \textbf{Datasets} &
  \textbf{DP} &
  \textbf{\#Clients} &
  \textbf{Non-IID} &
  \textbf{Code} \\ \hline
\textbf{~\cite{prediger2023collaborative}} &
  \xmark &
  \begin{tabular}[c]{@{}c@{}}Share Twin Data\end{tabular} &
  Image &
  UK biobank &
  \cmark &
  16 &
  \xmark &
  \cmark \\ \hline
\textbf{~\cite{guo2023enhancefl}} &
  \begin{tabular}[c]{@{}c@{}}Data Augmentation \end{tabular} &
  FedAVG &
  Image &
  Covid-19 Xray &
  \xmark &
  6-10 &
  \cmark &
  \xmark \\ \hline
\begin{tabular}[c]{@{}c@{}}
        \textbf{Fed-TDA}\\
        \textbf{~\cite{duan2022fedTDA}}
    \end{tabular} &
  \begin{tabular}[c]{@{}c@{}}Data Augmentation\end{tabular} &
  \begin{tabular}[c]{@{}c@{}}Fed-VB-GMM\\ FedAvg, FedProx\end{tabular} &
  Tabular &
  \begin{tabular}[c]{@{}c@{}}Clinical, Tuberculosis\\CovType, Intrusion, Body\end{tabular} &
  \cmark &
  60 &
  \cmark &
  \cmark \\ \hline

 \begin{tabular}[c]{@{}c@{}}
        \textbf{FedMix}\\
        \textbf{~\cite{yoon2021fedmix}}
\end{tabular}&
  Mixup &
  \begin{tabular}[c]{@{}c@{}}
        Weight Average
\end{tabular}&
  Image &
  \begin{tabular}[c]{@{}c@{}}FMNIST, Shakespeare\\ CIFAR10, CIFAR100\end{tabular} &
  \cmark &
  20, 40, 60 &
  \cmark &
  \cmark \\ \hline

   \begin{tabular}[c]{@{}c@{}}
        \textbf{3SFC}\\
        \textbf{~\cite{zhou2023communicationefficient}}
\end{tabular}&
  Synthetic Features &
  \begin{tabular}[c]{@{}c@{}}
        Synthetic Features Averaging
\end{tabular}&
  Image &
  \begin{tabular}[c]{@{}c@{}}MNIST, FMNIST, EMNIST\\ CIFAR10, CIFAR100\end{tabular} &
  \cmark &
  20 &
  \cmark &
  \xmark \\ \hline

    \begin{tabular}[c]{@{}c@{}}
        \textbf{FedDM}\\
        \textbf{~\cite{xiong2023feddm}}
    \end{tabular}&
  Synthetic Data (MMD) &
  Surrogate Function &
  Image &
  \begin{tabular}[c]{@{}c@{}}MNIST, CIFAR10, CIFAR100\end{tabular} &
  \cmark &
  10 &
  \cmark &
  \xmark \\ \hline
    \begin{tabular}[c]{@{}c@{}}
        \textbf{FEDGEN}\\
        \textbf{~\cite{zhu2021data}}
    \end{tabular} &
  \begin{tabular}[c]{@{}c@{}}Ensemble Model\end{tabular} &
  Knowledge Distillation &
  Image &
  \begin{tabular}[c]{@{}c@{}}MNIST, EMNIST, CelebA\end{tabular} &
  \xmark&
  20 &
  \cmark &
  \cmark \\ \hline
    \begin{tabular}[c]{@{}c@{}}
        \textbf{DENSE}\\
        \textbf{~\cite{zhang2022dense}}
    \end{tabular}&
  \begin{tabular}[c]{@{}c@{}}2-Stage Data Generation\\ Ensemble Model\end{tabular} &
  \begin{tabular}[c]{@{}c@{}}One Shot FL\\  Knowledge Distillation\end{tabular} &
  Image &
  \begin{tabular}[c]{@{}c@{}}MNIST, FMNIST, SVHN\\ CIFAR10, CIFAR100\\ Tiny-ImageNet\end{tabular} &
  \xmark&
  5-100 &
  \cmark &
  \xmark\\ \hline
   \textbf{
    \begin{tabular}[c]{@{}c@{}}
        Fedtmi\\~\cite{FedTMI2022Yao}
    \end{tabular}
}&
  GAN &
  Federated Transfer Learning &
  Image &
  \begin{tabular}[c]{@{}c@{}}Steam-driven Water Pumps\end{tabular} &
  \xmark&
  3 &
  \cmark &
  \xmark\\ \hline

\textbf{
    \begin{tabular}[c]{@{}c@{}}
        FCGAI\\~\cite{Zhou2021Fed}
    \end{tabular}
}&
  GAN &
  FedAVG &
  Image &
  \begin{tabular}[c]{@{}c@{}}Air Pollutants\end{tabular} &
  \xmark&
  3 &
  \cmark &
  \cmark\\ \hline

  \textbf{
    \begin{tabular}[c]{@{}c@{}}
        FedCG\\~\cite{Wu2022FedCG}
    \end{tabular}
}&
  DCGAN &
  \begin{tabular}[c]{@{}c@{}}Share Generators \\ Knowledge Distillation \end{tabular}&
  Image &
  \begin{tabular}[c]{@{}c@{}}FMNIST, CIFAR10\\ Difit5, DomainNet\\ Office-Caltech10\\
\end{tabular} &
  \xmark&
  4-8 &
  \cmark &
  \xmark\\ \hline
  
  \textbf{~\cite{li2022federated}} &
  \begin{tabular}[c]{@{}c@{}}WGAN-GP\\ AC-WGAN-GP\end{tabular} &
  Pseudo Label Update &
  Image &
  \begin{tabular}[c]{@{}c@{}}MNIST, FMNIST\\ CIFAR10, SVHN\end{tabular} &
  \cmark &
  $\sim$10 &
  \cmark &
  \xmark \\ \hline

  \textbf{
    \begin{tabular}[c]{@{}c@{}}
FLGAN\\~\cite{ma2023FLGAN}
    \end{tabular}
} &
  FHE-based GAN &
  \begin{tabular}[c]{@{}c@{}}FedAVG\\ FedSGD\end{tabular} &
  Image &
  \begin{tabular}[c]{@{}c@{}}MNIST, FMNIST, USPS\\ CIFAR10, CIFAR100\end{tabular} &
  \xmark &
  $\sim$10 &
  \cmark &
  \xmark\\ \hline

  \textbf{
    \begin{tabular}[c]{@{}c@{}}
        FedDDA\\
        ~\cite{zhao2023federated}
    \end{tabular}
    } &
  Diffusion Model &
  FedAvg &
  \begin{tabular}[c]{@{}c@{}}Image\end{tabular} &
  \begin{tabular}[c]{@{}c@{}}CIFAR10, FMNIST\end{tabular} &
  \xmark &
  2-4 &
  \cmark &
  \xmark \\ \hline

\textbf{
    \begin{tabular}[c]{@{}c@{}}
        FDA-CDM\\
        ~\cite{wang2023fda}
    \end{tabular}
    } &
  Diffusion Model &
  Share Data &
  \begin{tabular}[c]{@{}c@{}}Image\end{tabular} &
  \begin{tabular}[c]{@{}c@{}}CIFAR10, MNIST\end{tabular} &
  \xmark &
  20 &
  \cmark &
  \xmark \\ \hline

  \textbf{
    \begin{tabular}[c]{@{}c@{}}
        pFedV\\
        ~\cite{mou2023pfedv}
    \end{tabular}
} &
  VAE &
  FedAdam &
  Image &
  \begin{tabular}[c]{@{}c@{}}MNIST, MNIST-M, USPS\\ SVHN, Synthetic Digits\end{tabular} &
  \xmark&
  \xmark &
  \cmark &
  \xmark\\ \hline

  \textbf{
    \begin{tabular}[c]{@{}c@{}}
        ADPFedVAE\\
        ~\cite{ding2023combining}
    \end{tabular}
} &
  VAE &
  \begin{tabular}[c]{@{}c@{}}Randomly Select k Elements\\ FedAdam\end{tabular} &
  Tabular &
  \begin{tabular}[c]{@{}c@{}}MovieLens-1M2, \\ Lastfm-2K3, Steam\end{tabular} &
  \cmark &
  250 &
  \xmark &
  \xmark \\ \hline

    \textbf{
    \begin{tabular}[c]{@{}c@{}}
        Fed-MIWAE\\
        ~\cite{balelli2023fed}
    \end{tabular}
    } &
  VAE &
  \begin{tabular}[c]{@{}c@{}}FedAvg, FedPROX, Scaffold\end{tabular} &
  \begin{tabular}[c]{@{}c@{}}Image\end{tabular} &
  \begin{tabular}[c]{@{}c@{}}ADNI\end{tabular} &
  \xmark &
  3 &
  \cmark &
  \xmark \\ \hline

% from privacy/integrity
\textbf{
   \begin{tabular}[c]{@{}c@{}}
        FedFed\\
        ~\cite{yang2024fedfed}
    \end{tabular}
    } &
  VAE &
  \begin{tabular}[c]{@{}c@{}}
    FedAvg, FedProx\\ SCAFFOLD, FedNova\\ +  Information-Sharing\end{tabular} &
  Image &
  \begin{tabular}[c]{@{}c@{}}FMNIST\\ SVHN\\ CIFAR10\\CIFAR100\end{tabular} &
  \cmark &
  10, 100 &
  \cmark &
  \xmark\\ \hline

    \textbf{
    \begin{tabular}[c]{@{}c@{}}
        FedDNA\\
        ~\cite{duan2021feddna}
    \end{tabular}
} &
  VAE &
  Decoupled Parameter Aggregation &
  \begin{tabular}[c]{@{}c@{}}Image\\ NLP Prompt\end{tabular} &
  \begin{tabular}[c]{@{}c@{}}MNIST, FMNIST, CIFAR10\\ Sentiment140\end{tabular} &
  \xmark&
  5-20 &
  \cmark &
  \xmark \\ \hline
  
  \textbf{~\cite{duan2023federated}} &
  VAE &
  \begin{tabular}[c]{@{}c@{}}Distribution Fusion FL\\Interaction Aggregation \end{tabular} &
  Image &
  \begin{tabular}[c]{@{}c@{}}MNIST, FMNIST, CIFAR10\end{tabular} &
  \xmark&
  20-100 &
  \cmark &
  \cmark \\ \hline
\textbf{
    \begin{tabular}[c]{@{}c@{}}
        DistVAE\\
        ~\cite{li2023distvae}
    \end{tabular}
} &
    AC-VAE &
  \begin{tabular}[c]{@{}c@{}}Average Pooling\\ FedAVG (Virtual Groups)\end{tabular} &
  \begin{tabular}[c]{@{}c@{}}Tabular\end{tabular} &
  \begin{tabular}[c]{@{}c@{}}MovieLens Latest1\\ MovieLens-1m2 \\ MovieTweet-ings3, PEEK\end{tabular} &
  \xmark &
  4, 8, 16, 32, 64, 128 &
  \cmark &
  \xmark\\ \hline

\end{tabular}%
}
\end{table*}

\section{Security and Privacy Review}
\label{sec:security}
As we mentioned earlier, we can split the type of attacks into two groups: (i) Privacy Attacks (Table \ref{tab:overview-security-privacy}) and (ii) Integrity Attacks (Table \ref{tab:overview-security-integrity}). This section will review papers discussing the combination of FL and Generative Models with Privacy and Integrity evaluation and consideration. Although our primary focus is on Federated Generative Models, we also try to summarize some Generative Models based on attack and defense mechanisms in federated setup. 

\subsection{Consider Privacy Attacks}
% only privacy
\begin{table*}[]
\centering
\caption{Privacy Attacks Summary}
\label{tab:overview-security-privacy}
\setlength\tabcolsep{2pt}
\resizebox{\textwidth}{!}{%
\begin{tabular}{|c|c|c|c|c|c|c|c|c|c|c|}
\hline
\textbf{Paper} &
  \textbf{Model} &
  \textbf{Federate Algorithm} &
  \textbf{Data Type} &
  \textbf{Datasets} &
  \textbf{DP?} &
  \textbf{\#Clients} &
  \textbf{\begin{tabular}[c]{@{}c@{}}Explicitly evaluate \\ privacy attacks\end{tabular}} &
  \textbf{Non-IID} &
  \textbf{Code} \\ \hline
\textbf{~\cite{federatedDP2022xin}} &
  WGAN &
  FedAVG &
  Image &
  \begin{tabular}[c]{@{}c@{}}CelebA\\ MNIST\\ Fashion MNIST\end{tabular} &
  \cmark &
  $\sim$20 &
  Membership Inference &
  \cmark &
  \xmark \\ \hline
\textbf{
    \begin{tabular}[c]{@{}c@{}}
        PP-FedGAN\\
        ~\cite{ghavamipour2023federated}
    \end{tabular} 
} &
  WGAN &
  FedAVG &
  Image &
  \begin{tabular}[c]{@{}c@{}}MNIST\\ SVHN\end{tabular} &
  \cmark &
  $\sim$10 &
  \begin{tabular}[c]{@{}c@{}}Membership Inference \\ Property Inference\\ Reconstruction\end{tabular} &
  \xmark &
  \xmark \\ \hline
\textbf{
    \begin{tabular}[c]{@{}c@{}}
        HT-Fed-GAN\\
        ~\cite{duan2022htfedgan}
    \end{tabular} 
} &
  CGAN &
  Fed-VB-GMM &
  Tabular &
  \begin{tabular}[c]{@{}c@{}}Adult\\Health\\CovType\\ Intrusion\\Body\end{tabular} &
  \cmark &
  3 &
  Membership Inference &
  \cmark &
  \cmark \\ \hline
\textbf{
    \begin{tabular}[c]{@{}c@{}}
        GAFM\\
        ~\cite{han2023gan}
    \end{tabular} 
    } &
  GAN &
  \begin{tabular}[c]{@{}c@{}}VFL\\ Vanilla SplitNN\end{tabular} &
  Tabular &
  \begin{tabular}[c]{@{}c@{}}Spambase\\ IMDB\\ Criteo\\ ISIC\end{tabular} &
  \xmark &
  $\sim$10 &
  Label Leakage &
  \xmark &
  \xmark \\ \hline
\textbf{
    \begin{tabular}[c]{@{}c@{}}
        PS-FedGAN\\
        ~\cite{wijesinghe2023psfedgan}
    \end{tabular} 
} &
  CGAN &
  Sharing Discriminator &
  Image &
  \begin{tabular}[c]{@{}c@{}}MNIST\\ FMNIST\\ SVHN\\ CIFAR10\end{tabular} &
  \cmark &
  $\sim$30 &
  Reconstruction &
  \cmark &
  \xmark \\ \hline
\textbf{~\cite{sun2021information}} &
  GAN &
  Weight Average &
  Image &
  \begin{tabular}[c]{@{}c@{}}CIFAR10\\ MNIST\\ FMNIST\end{tabular} &
  \xmark&
  $\sim$10 &
  Reconstruction &
  \xmark&
  \xmark\\ \hline
\textbf{~\cite{ha2022inference}} &
  \begin{tabular}[c]{@{}c@{}}GAN\end{tabular} &
  \xmark &
  Image &
  \begin{tabular}[c]{@{}c@{}}LFW\\ CIFAR10\end{tabular} &
  \xmark &
  \xmark &
  Inference &
  \xmark &
  \xmark \\ \hline
  
% \textbf{
%    \begin{tabular}[c]{@{}c@{}}
%         ~\cite{Zhang2020GANmembership}
%     \end{tabular}
%     } &
%   GAN &
%   \xmark &
%   Image &
%   \begin{tabular}[c]{@{}c@{}}MNIST\end{tabular} &
%   \xmark &
%   100 &
%   \begin{tabular}[c]{@{}c@{}}Membership Inference\end{tabular} &
%   \xmark &
%   \xmark\\ \hline

%   \textbf{
%     \begin{tabular}[c]{@{}c@{}}
%         ~\cite{zhang2023federatedFoundation}
%     \end{tabular}
%     } &
%   Diffusion Model &
%   Prompt Aggregation &
%   \begin{tabular}[c]{@{}c@{}}Image\\NLP\end{tabular} &
%   \begin{tabular}[c]{@{}c@{}}ImageNette\\
% ImageFruit\\
% ImageYellow\\
% ImageSquawk\\
% ImageNet100\end{tabular} &
%   \xmark &
%   5 &
%   \begin{tabular}[c]{@{}c@{}}Membership Inference\end{tabular} &
%   \cmark &
%   \xmark \\ \hline
  
\end{tabular}%
}
\end{table*}

% GANs
\paragraph{\textbf{Federated GAN with Privacy Attacks Evaluation and Optimization}}
\cite{federatedDP2022xin} propose an extended version of private FL-GAN with DP, specifically considering membership inference attacks in a cross-silo setting of FL.
\cite{ghavamipour2023federated} proposes a FL-based GAN with DP and CKKS homomorphic encryption to counter membership inference attacks, optimizing the private FL-GAN approach from \cite{privateFLGAN2020Xin}.
\cite{duan2022htfedgan} focuses on preventing membership inference attacks within tabular GAN-based FL, marking the first effort towards privacy-preserving data synthesis on decentralized tabular datasets through a volitional Bayesian Gaussian mixture model.

% defence gan
\paragraph{\textbf{Using GANs for Defence Purpose in FL}}
\cite{wijesinghe2023psfedgan} propose a GAN-related FL framework to prevent reconstruction attacks and handle heterogeneous data distributions across clients. This method trains a generator at the server to mimic the data distribution of the clients' local GANs while sharing only the local discriminator, not the generator.
\cite{han2023gan} introduce GANs in vertical Federated Learning for label protection during binary classification. This approach improves label privacy protection by integrating splitNN with GANs, aiming to reduce Label Leakage from Gradients (LLG).

% attack gan
\paragraph{\textbf{Using GANs for Attack Purpose in FL}}
\cite{Zhang2020GANmembership, ha2022inference} propose an attack methodology where the attacker, an "honest but curious" client, uses GANs to infer information. Their study experiments with different GAN models: conditional GAN, control GAN, and WGAN.
\cite{sun2021information} use GANs to affect the learning process in FL and reconstruct clients' private data, focusing on a target-oriented data reconstruction attack.

\subsection{Consider Integrity Attacks}
% GANs
\paragraph{\textbf{Federated GANs with Integrity Attacks Evaluation and Optimization}}
\cite{Veeraragavan2023SecuringFedGan} addresses security weaknesses in GAN-based FL using Federated GANs, consortium blockchains, and Shamir Secret Sharing algorithms to reduce model poisoning attacks.
\cite{jin2022backdoor, jin2023backdoor} outline a backdoor attack strategy targeting GAN-based medical image synthesis, specifically within FedGAN frameworks. They implement a data poisoning strategy against discriminators and introduce FedDetect, a defensive mechanism to counter backdoor attacks.
\cite{Wang2023Poisoning} explores the potential for fake gradient attacks within distributed GAN settings, showing how poisoned gradients can disrupt the training process of generators.
\cite{fedganids2022tabassum} introduce an IDS using GAN, inspired by the FL-GAN architecture, to detect cyber threats in IoT systems, including backdoor classification efforts.

\paragraph{\textbf{FL-based VAEs with Privacy Attacks Evaluation and Optimization}}
VAE-based Anomaly Detection Framework: \cite{gu2021frepd} presents a VAE-based framework for anomaly detection, using reconstruction error probability distributions to identify benign model updates. It works with IID and non-IID data sources and evaluates Byzantine and Backdoor Attacks' robustness. 
\cite{gu2021detecting} further extends this concept with a CVAE-based framework for detecting and removing malicious model updates to reduce the negative impact of targeted model poisoning attacks.

\paragraph{\textbf{Using GANs for Attack Purpose in FL}}
\cite{zhang2019poisoning, psychogyios2023gan} introduces GAN-driven data poisoning and label flipping attacks, showcasing how attackers can stealthily generate and utilize poisoned data to compromise the FL process, proposing methods to reduce such attacks.

\paragraph{\textbf{Using GANs for Defence Purpose in FL}}
\cite{zhao2022detecting} propose a mechanism to mitigate and detect poisoning attacks in FL using GANs in the server. 
\cite{zhao2020pdgan} proposes a defense mechanism against poisoning in FL using GANs to reconstruct training data from model updates and evaluate client contributions for authenticity by filtering out attackers based on a predefined accuracy threshold.
% ~\cite{zhao2020pdgan} is a poisoning defense mechanism in FL using GAN to reconstruct training data from the model update and calculate the accuracy for each client. Then, the accuracy of the client will be compared with a threshold. If it is smaller than the threshold, it will be identified as an attacker, and the model parameters of the attacker will be removed from the training process.

\paragraph{\textbf{Using VAEs for Defence Purpose in FL}}
\cite{chelli2023fedguard} employ CVAE and selective parameter aggregation to defend against poisoning attacks, including sign flipping and label flipping.

% only integrity:
\begin{table*}[]
\centering
\caption{Integrity Attack Summary}
\label{tab:overview-security-integrity}
\setlength\tabcolsep{2pt}
\resizebox{\textwidth}{!}{%
\begin{tabular}{|c|c|c|c|c|c|c|c|c|c|c|}
\hline
\textbf{Paper} &
  \textbf{Model} &
  \textbf{Federate Algorithm} &
  \textbf{Data Type} &
  \textbf{Datasets} &
  \textbf{DP?} &
  \textbf{\#Clients} &
  \textbf{\begin{tabular}[c]{@{}c@{}}Explicitly evaluate \\ integrity attacks\end{tabular}} &
  \textbf{Non-IID} &
  \textbf{Code} \\ \hline
\textbf{
    \begin{tabular}[c]{@{}c@{}}
        FEDGAN-IDS\\
        ~\cite{fedganids2022tabassum}
    \end{tabular} 
    }&
  ACGAN &
  FedAVG &
  \begin{tabular}[c]{@{}c@{}}Image\\ Tabular\\ Time-Series\end{tabular} &
  \begin{tabular}[c]{@{}c@{}}KDD-CUP99\\ NSL-KDD\\ UNSW-NB15\end{tabular} &
  \xmark &
  $\sim$10 &
  Backdoor &
  \cmark &
  \xmark \\ \hline
\textbf{~\cite{jin2022backdoor}} &
  \begin{tabular}[c]{@{}c@{}}WGAN-DP\\ Vanilla GAN\end{tabular} &
  FedAVG &
  Image &
  ISIC &
  \cmark &
  $\sim$10 &
  Backdoor &
  \cmark &
  \cmark \\ \hline
\textbf{~\cite{jin2023backdoor}} &
  GAN &
  FedAVG&
  Image &
  \begin{tabular}[c]{@{}c@{}}ISIC\\ Chest X-ray\end{tabular} &
  \cmark &
  \xmark &
  Backdoor &
  \cmark &
  \xmark \\ \hline
\textbf{~\cite{Wang2023Poisoning}} &
  WGAN &
  \xmark &
  Image &
  \begin{tabular}[c]{@{}c@{}}MNIST\\ CIFAR10\end{tabular} &
  \xmark &
  $\sim$5 &
  Model Poisoning &
  \xmark &
  \xmark \\ \hline
\textbf{
    \begin{tabular}[c]{@{}c@{}}
        FREPD\\
        ~\cite{gu2021frepd}
    \end{tabular}
    } &
  VAE &
  GeoMed &
  Image &
  \begin{tabular}[c]{@{}c@{}}Vehicle\\ MNIST\\ FMNIST\end{tabular} &
  \xmark &
  10-50 &
  \begin{tabular}[c]{@{}c@{}}Byzantine\\ Backdoor\end{tabular} &
  \cmark &
  \xmark\\ \hline
\textbf{~\cite{gu2021detecting}} &
  CVAE &
  \begin{tabular}[c]{@{}c@{}}FedAVG\\ GeoMed\end{tabular}&
  Image &
  \begin{tabular}[c]{@{}c@{}}Vehicle\\ MNIST\\ FMNIST\end{tabular} &
  \xmark &
  10, 50, 1000 &
  \begin{tabular}[c]{@{}c@{}}Byzantine\\ Poisoning \end{tabular} &
  \cmark &
  \xmark\\ \hline 
\textbf{
    \begin{tabular}[c]{@{}c@{}}
        PDGAN\\
        ~\cite{zhao2020pdgan}
    \end{tabular}
    } &
  GAN &
  FedAVG &
  Image &
  \begin{tabular}[c]{@{}c@{}}MNIST\\ FMNIST\end{tabular} &
  \xmark&
  10 &
  Poisoning &
  \cmark &
  \xmark\\ \hline
\textbf{~\cite{zhao2022detecting}} &
  GAN &
  FedAVG &
  Image &
  \begin{tabular}[c]{@{}c@{}}MNIST\\ FMNIST\end{tabular} &
  \xmark &
  $\sim$100 &
  Poisoning &
  \xmark&
  \xmark\\ \hline
\textbf{
    \begin{tabular}[c]{@{}c@{}}
        FedGuard\\
        ~\cite{chelli2023fedguard}
    \end{tabular} 
} &
  CVAE &
  \begin{tabular}[c]{@{}c@{}}
        FedAVG
    \end{tabular}  &
  Image &
  MNIST &
  \xmark &
  100 &
  Poisoning &
  \xmark &
  \cmark \\ \hline
\textbf{~\cite{zhang2019poisoning}} &
  GAN &
  Weight Average &
  Image &
  \begin{tabular}[c]{@{}c@{}}MNIST\\ AT\&T\end{tabular} &
  \xmark &
  10-100 &
  Poisoning &
  \xmark &
  \xmark\\ \hline
\textbf{~\cite{psychogyios2023gan}} &
  GAN &
  \xmark &
  Image &
  Plant Leaf Diseases  &
  \xmark &
  \xmark &
  \begin{tabular}[c]{@{}c@{}}Poisoning\end{tabular} &
  \xmark &
  \xmark \\ \hline
\textbf{~\cite{Veeraragavan2023SecuringFedGan}} &
  GAN &
  FedAVG &
  \xmark &
  \xmark &
  \xmark &
  \xmark &
  Model Poisoning &
  \xmark &
  \xmark\\ \hline

\end{tabular}%
}
\end{table*}

\section{Related Work}
In this section, we review a wide range of surveys on Federated Learning and Generative Models and surveys that review the intersection of these two areas. This overview will cover foundational and advanced topics, showing each field's importance and ongoing development. 

\subsection{Federated Learning Surveys}
Federated Learning was introduced by Google in 2016, marking a change in the focus of Machine Learning models between distributed datasets. Since then, researchers have focused on enhancing performance and minimizing privacy risks. There are many surveys for Federated Learning based on different aspects.\\
First, many surveys cover the State-of-the-art and research challenges across different subsets of FL, such as Non-IID, vertical Federated Learning, heterogeneous data, and cross-silo Federated Learning. ~\cite{ye2023heterogeneoussurvey, zhu2021federated, ye2023heterogeneoussurvey, wei2022vertical} are some good surveys in this specific areas and also flower~\cite{beutel2022flower} is the best framework for different Federated Learning setups. Second, Federated Learning is applied across various fields, including finance, health records, transportation, and NLP. Earlier studies like ~\cite{li2020review, yang2019federated, aledhari2020federated, banabilah2022federated, kairouz2021advances, zhang2021survey, wen2023survey} provide general insights for beginning research and cover application aspects. On the other hand, ~\cite{liu2024vertical, rauniyar2023federated, pandya2023federated, che2023multimodal, li2023review, rauniyar2023federated} focus on specific applications, offering detailed reviews. Finally, considering privacy and security concerns in Federated Learning is a common and ongoing research area. Recent surveys like ~\cite{mothukuri2021survey, rauniyar2023federated, li2023review, shao2023survey, zhang2023systematic, rasha2023federated, rodriguez2023survey, gosselin2022privacy, lyu2020threats} extensively cover related issues of threats, security, toxicity, and privacy in Federated Learning.

\subsection{Generative Models Surveys}
Deep Generative Models have many applications, such as data augmentation, text, video, audio synthesis, anomaly detection, etc. Many surveys focus on different applications and gather different state-of-the-art generative models. ~\cite{salakhutdinov2015learning, oussidi2018deep, ruthotto2021introduction} are some older surveys that cover the whole idea of Deep Generative Models and can be used to introduce generative models. ~\cite{creswell2018generative, wang2017generative, gui2021review} are basic surveys that cover applications of Generative Adversarial Networks. We have well-covered new works for specific applications, like ~\cite{dash2023review, de2023review, brophy2023generative}. ~\cite{chakraborty2024ten, megahed2024comprehensive} are the newest surveys that cover the state-of-the-art of GANs over 10 years. On the other hand, many surveys cover variational autoencoders. ~\cite{zhai2018autoencoder, ghojogh2021factor, wan2017variational} provide us with good intuition and familiarity with basic aspects and variants of autoencoders, and ~\cite{bank2023autoencoders, papadopoulos2023variational, li2023comprehensive} are the newest works in this area. Diffusion Models are a new trend in the field of generative models, and ~\cite{cao2024survey, san2021noise, yang2023diffusion, croitoru2023diffusion} are well-organized surveys on generative Diffusion Models. Generative Models can also be used for security aspects and anomaly detection. ~\cite{vyas2023generative, kos2018adversarial, sun2021adversarial, cai2021generative} are some general reviews in this area.

\subsection{Federated Learning and Generative Models Surveys}
Since we cover the specific combination of Generative Models and Federated Learning, not many reviews cover both in one place. ~\cite{little2023federated} is the only survey that covers the FL and Synthetic data. Their scoping review only focuses on 69 research papers in this area, including Federated synthesis, Improving FL using augmentation, Developing FL, and Using Generative Models for Synthetic Data. Additionally, they describe some state-of-the-art methods such as Fed-TGAN, GTV, and SGDE. Moreover, they only cover synthetic data in a federated setup and consider Generative Models as a method for synthetic data. 

The main difference between our works and ~\cite{little2023federated} is we mainly focus on reviewing all related papers, but their primary goal is having a numerical review and only elaborating on a few important papers. Their primary focus is on using Synthetic Data, whether they use Generative Models or not. Their review also did not consider the privacy and integrity aspects of different papers, which was one of our main goals.

In the context of an attack on Federated Learning using Generative Models, many surveys mentioned it as the application of Generative Models or Threats of Federated Learning~\cite{lyu2020threats}. ~\cite{ma2022state} is a good survey that covers how to handle solving Non-IID data in Federated Learning using GANs.

\section{Discussion and Conclusion}
\label{sec:discussion}

\subsection{State-of-the-Arts Summary}
\paragraph{\textbf{Most Used Federated GANs:}} Most papers compare the performance of their new models with GAN-based FL: MD-GAN, AC-GAN, FL-GAN, DP-FedAvg-GAN, and IFL-GAN.
\paragraph{\textbf{Available Code:}} Among those papers with available code, these are the most common for comparison: DPD-fVAE, HT-Fed-GAN, DP-fedAvg-GAN, Fed-TGAN.
\paragraph{\textbf{Tabular Data:}} For tabular data, the most used papers to compare are Fed-TGAN, HT-Fed-GAN, and GDTS.
\paragraph{\textbf{VAE and Diffusion Model-based FL:}} For VAE and Diffusion Model-based papers, DPD-fVAE, FedDPMS, ADPFedVAE, Phoenix, and FedTabDiff are the most used papers for comparison.
\paragraph{\textbf{Common Attacks:}} As all common attacks in FL, Membership Inference and Poisoning Attacks are the most common in FL-based Generative Models (based on the number of research focusing on these areas). Additionally, researchers commonly use Generative Models for Attack/Defence purposes, but it was not the focus of our surveys.
\paragraph{\textbf{Satisfy DP:}} Fed-TGAN, HT-Fed-GAN, Private FL-GAN, DP-FedAVG-GAN, DPD-fVAE, and SGDE are the most common papers that evaluate different values of Privacy Parameters to satisfy DP definition.
\paragraph{\textbf{Aggregated Methods:}} FedAvg is the most common aggregation method used among different methods. However, model aggregation, MMD Score, FedProx, and FedSGD are also used in different models.
\paragraph{\textbf{Datasets}} Majority of the evaluation datasets in Image-based Models are MNIST, FMNIST, SVHN, CelebA, CIFAR10/CIFAR100, and Breast Cancer Wisconsin,. On the other hand, for Tabular Data, the most used datasets are from the UCI Machine Learning Repository (e.g., ADULT, Covertype, Housing, Credit, Body), Intrusion, and MovieLens. Additionally, Microsoft's Geolife is used for Trajectory data, and OpenImage, NICO++, and Domain-Net are used for One-Shot FL papers.
% \paragraph{\textbf{IID Data}} MD-GAN, FL-GAN, AC-GAN, DP-FedAVG, DPD-fVAE, and FedVAE are FedDM's most used papers for IID dataset comparison.
% \paragraph{\textbf{Non-IID Data}} PerFED-GAN, Fed-GAN, FedDPGAN, GDTS, GTV, FeGAN, 
    % \item Those papers that cover One-Shot FL can be impressive and a communication efficient method.

% bullet point based:
\subsection{What have we learned?}
As we mentioned in Section \ref{sec:model-review}, the main focus of this survey is on Federated Generative Models among all papers that combine FL and GANs. After learning about the SOTA in Federated Generative Models, we elaborate on anything we learned from papers in this section. 

\paragraph{\textbf{Main Reason of Federate Generative Models:}} Among all of the papers that we reviewed, we can summarize the main points of using Federated Generative Models into the following groups: 
    \begin{enumerate}
       
        \item \textbf{Lack of Sufficient Data in Single Clients:} Using FL-based Generative Models for edge devices with a small amount of data can be efficient. In other words, it benefits clients with limited data resources, such as medical records, finance fraud detection, and recommendation systems for distributed devices.

        \item \textbf{Prevent Sharing Clients' Data and Privacy-Preserving:} As all FL models, using Federated Generative Models will leverage the advantage of keeping data local in the generating process.

        \item \textbf{Communication Efficient:} Learning to generate data at the central server, and not at the clients, is communication efficient and does not require more computational power from the clients. Many research studies have proposed an efficient communication method for Federated Generative Models. 
    \end{enumerate}

\paragraph{\textbf{Different Types of Federated GANs:}} For Federated GANs, we can categorize them into three groups:
    \begin{enumerate}
        \item \textbf{Similar to MD-GAN}: One generator at the server, many discriminators at clients.
        \item \textbf{Similar to FL-GAN}: One GAN at the server, many GANs at clients (Federated Learning Adapted to GAN).
        \item \textbf{Splitted GAN:} Having GANs at clients and only having a Generator/Discriminator at the server.
    \end{enumerate}

    \paragraph{\textbf{Satisfy DP:}} Federated Generative Models efficiently satisfy the Differential Privacy (DP) definition, and many research papers test different values of privacy parameters. 

    \paragraph{\textbf{Various Types of Data:}} Much research focuses on using Federated Generative Models for data types such as images, tabular, time-series, and trajectories. 
    It increases research for many applications, such as clinical applications, financial datasets, and IoT devices.
    A major research focus is also using VAE and FL for recommender systems in federated settings and clustering data.

    \paragraph{\textbf{Generative Models and FL:}}
    On the other hand, Many papers focus on using Generative Models only for clients or only at servers to solve data heterogeneity and Non-IID challenges, Anomaly Detection, ensure privacy, and prevent poisoning attacks. This is also one use of FL and Generative Models, \textbf{which is not the focus of this survey}. 

    % \textbf{Address Heterogeneity and Non-IID:} Using Generative Models can be one of the ways to address data heterogeneity and Non-IID challenges.
    % \item Using generative models alone can successfully solve the Non-IID and data heterogeneity challenges. 
    % \item To ensure privacy, some papers propose encryption methods.
    % \item Also, many papers suggest using generative data to synthesize and share synthetic data among clients and servers.
    % \textbf{Clinical Application:} Since much research focuses on using FL-based Generative Models for Image classification, Many clinical applications of FL-based generative models have been proposed.
    % \textbf{Recommender Systems:} Using VAE and FL for recommender systems in federated settings and clustering data is also a major research focus.

    % \item Generative models can be helpful for attack and defense purposes in federated settings.

\subsection{Future Works and Open Problems}
As we mentioned earlier, the main reason for bringing generative models into FL was to protect individual data by preventing sensitive data sharing and communication overhead by keeping it in their devices. On the other hand, some works focus on using generative models in client or server to utilize the applications of generative models such as anomaly detection, addressing Non-IID and data heterogeneity challenges, inference attacks, and protection against reconstruction attacks.

% In summary, we can group the research that combines generative models and Federated Learning into (i) Client-Generator, where synthetic data is generated locally and then communicated to the server for the server model updating; (ii) Server-Generator, where a centralized generator is used to assist in updating the server and client model, (iii) Federated-Generator, both clients and server working together to generate new dataset. However, our primary focus will be on the \textit{Federated-Generator} category.

\paragraph{\textbf{Privacy and Security:}}
Reviewing papers reveals many open research problems regarding privacy and security. 
None of the papers focusing on Integrity Attacks consider Tabular datasets. 
None of the papers focus on Privacy and Integrity attacks in Diffusion-based FL models.
Ranking-based training of Federated Learning to Resist Poisoning Attacks is an interesting topic recently proposed~\cite{mozaffari2023every}, and combining its idea with Federated Generative Models can guarantee more privacy-preserving.

\paragraph{\textbf{Encryption:}}
Although some research exists on parameter-sharing encryption in image-based Federated Models, the impact of using encryption for sharing gradients and parameters in tabular data models or anomaly detection models remains unresolved. Furthermore, the application of parameter-sharing encryption in Diffusion Models and Variational Autoencoders (VAEs) is still an open problem.

% \begin{itemize}
%     \item \textbf{Privacy and Security:}
%     \begin{itemize}
%         \item None of the papers focusing on integrity attacks consider tabular datasets.chr
%         \item None of the papers focus on privacy and integrity attacks in diffusion-based FL models.
%         \item Ranking-Based Training of Federated Learning to Resist Poisoning Attacks is an interesting topic recently proposed, and combining its idea with Generative Models can guarantee more privacy-preserving.
%         \item The effect of using encryption for sharing tabular data or anomaly detection models is still an open problem.[not sure]
%     \end{itemize}

\paragraph{\textbf{Diffusion Models}} In recent months, due to the convergence challenges faced by GANs in both centralized and federated setups, many research efforts have shifted towards using Diffusion Models, particularly in diffusion-based Federated Learning (FL). The competitive focus on Federated Diffusion Models has increased significant research potential in this area.

\paragraph{\textbf{Scalability:}} Most papers focus on the small size of clients; increasing the number of clients might affect the performance and robustness of different Federated Models. Scalability analysis and consideration are still open problems among Generative-based FL models, specifically for Federated VAEs and Diffusion Models. 

    % \item Investigating the combination of FL and generative models for cross-modal and multi-task learning scenarios. This involves generating synthetic data that spans multiple modes (e.g., text, image) or learning tasks to improve model performance and data utility. [can be not related]
\paragraph{\textbf{IoT Devices:}} Developing lightweight generative models suitable for FL in resource-constrained environments like IoT devices. This includes optimizing models for energy efficiency and memory usage. Currently, only GANs are used for IoT devices, which is not a lightweight generative model. Using other generative models and making them more efficient is still an open problem.

\paragraph{\textbf{One-shot FL / Pre-Trained Models:}}
    \begin{itemize}
        \item Methods in one-shot FL have demonstrated strong performance under substantial communication constraints. One-shot FL is an open and interesting topic in FL-related research, and only three papers try to use Generative Models for one-shot FL.
        \item Only ~\cite{heinbaugh2023datafree} in one-shot FL did an experiment with high statistical heterogeneity. It is still an open problem that can focus on it.
        \item Only a few studies have focused on the potential of pre-trained Diffusion Models in FL.
    \end{itemize}

\paragraph{\textbf{Ensemble Models:}} Ensemble Models combining all \textit{weak} learner or \textit{well-chosen strong} models can be an open problem to gain more robustness. For example, different generative models can be used for different clients, or different generative models can be used on servers to recognize poisoning attacks based on their method.

\paragraph{\textbf{Prompt-Based + Generative AI}}
The foundation generative models are still under-explored in Federated Learning, though a few related works study foundation models in Federated Learning. Moreover, combining Prompt-based and Generative AI with FL and Diffusion Models can still be a good research idea that can focus on it. 

\paragraph{\textbf{Multi-objective Optimization:}} ~\cite{ran2024multi} propose a Multi-objective evolutionary GAN for tabular data synthesis and applying it to federated-based GANs can be an open research problem.

\bibliographystyle{ACM-Reference-Format}
\bibliography{references.bib}


\begin{thebibliography}{180}

%%% the Web address.

\ifx \showCODEN    \undefined \def \showCODEN     #1{\unskip}     \fi
\ifx \showDOI      \undefined \def \showDOI       #1{#1}\fi
\ifx \showISBNx    \undefined \def \showISBNx     #1{\unskip}     \fi
\ifx \showISBNxiii \undefined \def \showISBNxiii  #1{\unskip}     \fi
\ifx \showISSN     \undefined \def \showISSN      #1{\unskip}     \fi
\ifx \showLCCN     \undefined \def \showLCCN      #1{\unskip}     \fi
\ifx \shownote     \undefined \def \shownote      #1{#1}          \fi
\ifx \showarticletitle \undefined \def \showarticletitle #1{#1}   \fi
\ifx \showURL      \undefined \def \showURL       {\relax}        \fi
% invisible to TeX
\providecommand\bibfield[2]{#2}
\providecommand\bibinfo[2]{#2}
\providecommand\natexlab[1]{#1}
\providecommand\showeprint[2][]{arXiv:#2}

\bibitem[Acs et~al\mbox{.}(2018)]%
        {acs2018differentially}
\bibfield{author}{\bibinfo{person}{Gergely Acs}, \bibinfo{person}{Luca Melis}, \bibinfo{person}{Claude Castelluccia}, {and} \bibinfo{person}{Emiliano~De Cristofaro}.} \bibinfo{year}{2018}\natexlab{}.
\newblock \bibinfo{title}{Differentially Private Mixture of Generative Neural Networks}.
\newblock
\newblock
\showeprint[arxiv]{1709.04514}~[cs.LG]


\bibitem[Ahn et~al\mbox{.}(2023)]%
        {ahn2023communicationefficient}
\bibfield{author}{\bibinfo{person}{Seyoung Ahn}, \bibinfo{person}{Soohyeong Kim}, \bibinfo{person}{Yongseok Kwon}, \bibinfo{person}{Joohan Park}, \bibinfo{person}{Jiseung Youn}, {and} \bibinfo{person}{Sunghyun Cho}.} \bibinfo{year}{2023}\natexlab{}.
\newblock \bibinfo{title}{Communication-Efficient Diffusion Strategy for Performance Improvement of Federated Learning with Non-IID Data}.
\newblock
\newblock
\showeprint[arxiv]{2207.07493}~[cs.DC]


\bibitem[Aledhari et~al\mbox{.}(2020)]%
        {aledhari2020federated}
\bibfield{author}{\bibinfo{person}{Mohammed Aledhari}, \bibinfo{person}{Rehma Razzak}, \bibinfo{person}{Reza~M Parizi}, {and} \bibinfo{person}{Fahad Saeed}.} \bibinfo{year}{2020}\natexlab{}.
\newblock \showarticletitle{Federated learning: A survey on enabling technologies, protocols, and applications}.
\newblock \bibinfo{journal}{\emph{IEEE Access}}  \bibinfo{volume}{8} (\bibinfo{year}{2020}), \bibinfo{pages}{140699--140725}.
\newblock


\bibitem[Annamalai et~al\mbox{.}(2023)]%
        {annamalai2023fp}
\bibfield{author}{\bibinfo{person}{Meenatchi Sundaram Muthu~Selva Annamalai}, \bibinfo{person}{Igor Bilogrevic}, {and} \bibinfo{person}{Emiliano De~Cristofaro}.} \bibinfo{year}{2023}\natexlab{}.
\newblock \showarticletitle{FP-Fed: Privacy-Preserving Federated Detection of Browser Fingerprinting}.
\newblock \bibinfo{journal}{\emph{arXiv:2311.16940}} (\bibinfo{year}{2023}).
\newblock


\bibitem[Arafeh et~al\mbox{.}(2022)]%
        {arafeh2022independent}
\bibfield{author}{\bibinfo{person}{Mohamad Arafeh}, \bibinfo{person}{Ahmad Hammoud}, \bibinfo{person}{Hadi Otrok}, \bibinfo{person}{Azzam Mourad}, \bibinfo{person}{Chamseddine Talhi}, {and} \bibinfo{person}{Zbigniew Dziong}.} \bibinfo{year}{2022}\natexlab{}.
\newblock \showarticletitle{Independent and identically distributed (iid) data assessment in federated learning}. In \bibinfo{booktitle}{\emph{GLOBECOM 2022-2022 IEEE Global Communications Conference}}. IEEE, \bibinfo{pages}{293--298}.
\newblock


\bibitem[Augenstein et~al\mbox{.}(2019)]%
        {augenstein2019generative}
\bibfield{author}{\bibinfo{person}{Sean Augenstein}, \bibinfo{person}{H~Brendan McMahan}, \bibinfo{person}{Daniel Ramage}, \bibinfo{person}{Swaroop Ramaswamy}, \bibinfo{person}{Peter Kairouz}, \bibinfo{person}{Mingqing Chen}, \bibinfo{person}{Rajiv Mathews}, {et~al\mbox{.}}} \bibinfo{year}{2019}\natexlab{}.
\newblock \showarticletitle{Generative models for effective ML on private, decentralized datasets}.
\newblock \bibinfo{journal}{\emph{arXiv:1911.06679}} (\bibinfo{year}{2019}).
\newblock


\bibitem[Balelli et~al\mbox{.}(2023)]%
        {balelli2023fed}
\bibfield{author}{\bibinfo{person}{Irene Balelli}, \bibinfo{person}{Aude Sportisse}, \bibinfo{person}{Francesco Cremonesi}, \bibinfo{person}{Pierre-Alexandre Mattei}, {and} \bibinfo{person}{Marco Lorenzi}.} \bibinfo{year}{2023}\natexlab{}.
\newblock \showarticletitle{Fed-MIWAE: Federated Imputation of Incomplete Data via Deep Generative Models}.
\newblock \bibinfo{journal}{\emph{arXiv:2304.08054}} (\bibinfo{year}{2023}).
\newblock


\bibitem[Banabilah et~al\mbox{.}(2022)]%
        {banabilah2022federated}
\bibfield{author}{\bibinfo{person}{Syreen Banabilah}, \bibinfo{person}{Moayad Aloqaily}, \bibinfo{person}{Eitaa Alsayed}, \bibinfo{person}{Nida Malik}, {and} \bibinfo{person}{Yaser Jararweh}.} \bibinfo{year}{2022}\natexlab{}.
\newblock \showarticletitle{Federated learning review: Fundamentals, enabling technologies, and future applications}.
\newblock \bibinfo{journal}{\emph{Information processing \& management}} \bibinfo{volume}{59}, \bibinfo{number}{6} (\bibinfo{year}{2022}), \bibinfo{pages}{103061}.
\newblock


\bibitem[Bank et~al\mbox{.}(2023)]%
        {bank2023autoencoders}
\bibfield{author}{\bibinfo{person}{Dor Bank}, \bibinfo{person}{Noam Koenigstein}, {and} \bibinfo{person}{Raja Giryes}.} \bibinfo{year}{2023}\natexlab{}.
\newblock \showarticletitle{Autoencoders}.
\newblock \bibinfo{journal}{\emph{Machine learning for data science handbook: data mining and knowledge discovery handbook}} (\bibinfo{year}{2023}), \bibinfo{pages}{353--374}.
\newblock


\bibitem[Behera et~al\mbox{.}(2022)]%
        {behera2022fedsyn}
\bibfield{author}{\bibinfo{person}{Monik~Raj Behera}, \bibinfo{person}{Sudhir Upadhyay}, \bibinfo{person}{Suresh Shetty}, \bibinfo{person}{Sudha Priyadarshini}, \bibinfo{person}{Palka Patel}, {and} \bibinfo{person}{Ker~Farn Lee}.} \bibinfo{year}{2022}\natexlab{}.
\newblock \showarticletitle{Fedsyn: Synthetic data generation using federated learning}.
\newblock \bibinfo{journal}{\emph{arXiv:2203.05931}} (\bibinfo{year}{2022}).
\newblock


\bibitem[Beutel et~al\mbox{.}(2022)]%
        {beutel2022flower}
\bibfield{author}{\bibinfo{person}{Daniel~J. Beutel}, \bibinfo{person}{Taner Topal}, \bibinfo{person}{Akhil Mathur}, \bibinfo{person}{Xinchi Qiu}, \bibinfo{person}{Javier Fernandez-Marques}, \bibinfo{person}{Yan Gao}, \bibinfo{person}{Lorenzo Sani}, \bibinfo{person}{Kwing~Hei Li}, \bibinfo{person}{Titouan Parcollet}, \bibinfo{person}{Pedro Porto~Buarque de Gusmão}, {and} \bibinfo{person}{Nicholas~D. Lane}.} \bibinfo{year}{2022}\natexlab{}.
\newblock \bibinfo{title}{Flower: A Friendly Federated Learning Research Framework}.
\newblock
\newblock
\showeprint[arxiv]{2007.14390}~[cs.LG]


\bibitem[Bhagoji et~al\mbox{.}(2019)]%
        {bhagoji2019analyzing}
\bibfield{author}{\bibinfo{person}{Arjun~Nitin Bhagoji}, \bibinfo{person}{Supriyo Chakraborty}, \bibinfo{person}{Prateek Mittal}, {and} \bibinfo{person}{Seraphin Calo}.} \bibinfo{year}{2019}\natexlab{}.
\newblock \showarticletitle{Analyzing federated learning through an adversarial lens}. In \bibinfo{booktitle}{\emph{International Conference on Machine Learning}}. PMLR, \bibinfo{pages}{634--643}.
\newblock


\bibitem[Bouacida and Mohapatra(2021)]%
        {Bouacidea2021VulnerabilitiesFL}
\bibfield{author}{\bibinfo{person}{Nader Bouacida} {and} \bibinfo{person}{Prasant Mohapatra}.} \bibinfo{year}{2021}\natexlab{}.
\newblock \showarticletitle{Vulnerabilities in Federated Learning}.
\newblock \bibinfo{journal}{\emph{IEEE Access}}  \bibinfo{volume}{9} (\bibinfo{year}{2021}), \bibinfo{pages}{63229--63249}.
\newblock
\urldef\tempurl%
\url{https://doi.org/10.1109/ACCESS.2021.3075203}
\showDOI{\tempurl}


\bibitem[Brophy et~al\mbox{.}(2023)]%
        {brophy2023generative}
\bibfield{author}{\bibinfo{person}{Eoin Brophy}, \bibinfo{person}{Zhengwei Wang}, \bibinfo{person}{Qi She}, {and} \bibinfo{person}{Tom{\'a}s Ward}.} \bibinfo{year}{2023}\natexlab{}.
\newblock \showarticletitle{Generative adversarial networks in time series: A systematic literature review}.
\newblock \bibinfo{journal}{\emph{Comput. Surveys}} \bibinfo{volume}{55}, \bibinfo{number}{10} (\bibinfo{year}{2023}), \bibinfo{pages}{1--31}.
\newblock


\bibitem[Cai et~al\mbox{.}(2021)]%
        {cai2021generative}
\bibfield{author}{\bibinfo{person}{Zhipeng Cai}, \bibinfo{person}{Zuobin Xiong}, \bibinfo{person}{Honghui Xu}, \bibinfo{person}{Peng Wang}, \bibinfo{person}{Wei Li}, {and} \bibinfo{person}{Yi Pan}.} \bibinfo{year}{2021}\natexlab{}.
\newblock \showarticletitle{Generative adversarial networks: A survey toward private and secure applications}.
\newblock \bibinfo{journal}{\emph{ACM Computing Surveys (CSUR)}} \bibinfo{volume}{54}, \bibinfo{number}{6} (\bibinfo{year}{2021}), \bibinfo{pages}{1--38}.
\newblock


\bibitem[Cao et~al\mbox{.}(2024)]%
        {cao2024survey}
\bibfield{author}{\bibinfo{person}{Hanqun Cao}, \bibinfo{person}{Cheng Tan}, \bibinfo{person}{Zhangyang Gao}, \bibinfo{person}{Yilun Xu}, \bibinfo{person}{Guangyong Chen}, \bibinfo{person}{Pheng-Ann Heng}, {and} \bibinfo{person}{Stan~Z Li}.} \bibinfo{year}{2024}\natexlab{}.
\newblock \showarticletitle{A survey on generative diffusion models}.
\newblock \bibinfo{journal}{\emph{IEEE Transactions on Knowledge and Data Engineering}} (\bibinfo{year}{2024}).
\newblock


\bibitem[Cao et~al\mbox{.}(2022)]%
        {perfedGan2022cao}
\bibfield{author}{\bibinfo{person}{Xingjian Cao}, \bibinfo{person}{Gang Sun}, \bibinfo{person}{Hongfang Yu}, {and} \bibinfo{person}{Mohsen Guizani}.} \bibinfo{year}{2022}\natexlab{}.
\newblock \showarticletitle{PerFED-GAN: Personalized federated learning via generative adversarial networks}.
\newblock \bibinfo{journal}{\emph{IEEE Internet of Things Journal}} \bibinfo{volume}{10}, \bibinfo{number}{5} (\bibinfo{year}{2022}), \bibinfo{pages}{3749--3762}.
\newblock


\bibitem[Chakraborty et~al\mbox{.}(2024)]%
        {chakraborty2024ten}
\bibfield{author}{\bibinfo{person}{Tanujit Chakraborty}, \bibinfo{person}{Ujjwal~Reddy KS}, \bibinfo{person}{Shraddha~M Naik}, \bibinfo{person}{Madhurima Panja}, {and} \bibinfo{person}{Bayapureddy Manvitha}.} \bibinfo{year}{2024}\natexlab{}.
\newblock \showarticletitle{Ten years of generative adversarial nets (GANs): a survey of the state-of-the-art}.
\newblock \bibinfo{journal}{\emph{Machine Learning: Science and Technology}} \bibinfo{volume}{5}, \bibinfo{number}{1} (\bibinfo{year}{2024}), \bibinfo{pages}{011001}.
\newblock


\bibitem[Che et~al\mbox{.}(2023)]%
        {che2023multimodal}
\bibfield{author}{\bibinfo{person}{Liwei Che}, \bibinfo{person}{Jiaqi Wang}, \bibinfo{person}{Yao Zhou}, {and} \bibinfo{person}{Fenglong Ma}.} \bibinfo{year}{2023}\natexlab{}.
\newblock \showarticletitle{Multimodal federated learning: A survey}.
\newblock \bibinfo{journal}{\emph{Sensors}} \bibinfo{volume}{23}, \bibinfo{number}{15} (\bibinfo{year}{2023}), \bibinfo{pages}{6986}.
\newblock


\bibitem[Chelli et~al\mbox{.}(2023)]%
        {chelli2023fedguard}
\bibfield{author}{\bibinfo{person}{Melvin Chelli}, \bibinfo{person}{C{\'e}dric Prigent}, \bibinfo{person}{Ren{\'e} Schubotz}, \bibinfo{person}{Alexandru Costan}, \bibinfo{person}{Gabriel Antoniu}, \bibinfo{person}{Lo{\"\i}c Cudennec}, {and} \bibinfo{person}{Philipp Slusallek}.} \bibinfo{year}{2023}\natexlab{}.
\newblock \showarticletitle{FedGuard: Selective Parameter Aggregation for Poisoning Attack Mitigation in Federated Learning}. In \bibinfo{booktitle}{\emph{2023 IEEE International Conference on Cluster Computing (CLUSTER)}}. IEEE, \bibinfo{pages}{72--81}.
\newblock


\bibitem[Chen et~al\mbox{.}(2023a)]%
        {chen2023label}
\bibfield{author}{\bibinfo{person}{Baojian Chen}, \bibinfo{person}{Hongjia Li}, \bibinfo{person}{Lu Guo}, {and} \bibinfo{person}{Liming Wang}.} \bibinfo{year}{2023}\natexlab{a}.
\newblock \showarticletitle{Label-wise Distribution Adaptive Federated Learning on Non-IID Data}. In \bibinfo{booktitle}{\emph{2023 IEEE Wireless Communications and Networking Conference (WCNC)}}. IEEE, \bibinfo{pages}{1--6}.
\newblock


\bibitem[Chen and Vikalo(2023)]%
        {chen2023federated}
\bibfield{author}{\bibinfo{person}{Huancheng Chen} {and} \bibinfo{person}{Haris Vikalo}.} \bibinfo{year}{2023}\natexlab{}.
\newblock \showarticletitle{Federated learning in non-iid settings aided by differentially private synthetic data}. In \bibinfo{booktitle}{\emph{Proceedings of the IEEE/CVF Conference on Computer Vision and Pattern Recognition}}. \bibinfo{pages}{5026--5035}.
\newblock


\bibitem[Chen et~al\mbox{.}(2023b)]%
        {chen2023fedlgan}
\bibfield{author}{\bibinfo{person}{Zheliang Chen}, \bibinfo{person}{Xianhan Ni}, \bibinfo{person}{Huan Li}, {and} \bibinfo{person}{Xiangjie Kong}.} \bibinfo{year}{2023}\natexlab{b}.
\newblock \showarticletitle{FedLGAN: a method for anomaly detection and repair of hydrological telemetry data based on federated learning}.
\newblock \bibinfo{journal}{\emph{PeerJ Computer Science}}  \bibinfo{volume}{9} (\bibinfo{year}{2023}), \bibinfo{pages}{e1664}.
\newblock


\bibitem[Chiaro et~al\mbox{.}(2023)]%
        {chiaro2023flenhance}
\bibfield{author}{\bibinfo{person}{Diletta Chiaro}, \bibinfo{person}{Edoardo Prezioso}, \bibinfo{person}{Michele Ianni}, {and} \bibinfo{person}{Fabio Giampaolo}.} \bibinfo{year}{2023}\natexlab{}.
\newblock \showarticletitle{FL-Enhance: A federated learning framework for balancing non-IID data with augmented and shared compressed samples}.
\newblock \bibinfo{journal}{\emph{Information Fusion}}  \bibinfo{volume}{98} (\bibinfo{year}{2023}), \bibinfo{pages}{101836}.
\newblock


\bibitem[Creswell et~al\mbox{.}(2018)]%
        {creswell2018generative}
\bibfield{author}{\bibinfo{person}{Antonia Creswell}, \bibinfo{person}{Tom White}, \bibinfo{person}{Vincent Dumoulin}, \bibinfo{person}{Kai Arulkumaran}, \bibinfo{person}{Biswa Sengupta}, {and} \bibinfo{person}{Anil~A Bharath}.} \bibinfo{year}{2018}\natexlab{}.
\newblock \showarticletitle{Generative adversarial networks: An overview}.
\newblock \bibinfo{journal}{\emph{IEEE signal processing magazine}} \bibinfo{volume}{35}, \bibinfo{number}{1} (\bibinfo{year}{2018}), \bibinfo{pages}{53--65}.
\newblock


\bibitem[Croitoru et~al\mbox{.}(2023)]%
        {croitoru2023diffusion}
\bibfield{author}{\bibinfo{person}{Florinel-Alin Croitoru}, \bibinfo{person}{Vlad Hondru}, \bibinfo{person}{Radu~Tudor Ionescu}, {and} \bibinfo{person}{Mubarak Shah}.} \bibinfo{year}{2023}\natexlab{}.
\newblock \showarticletitle{Diffusion models in vision: A survey}.
\newblock \bibinfo{journal}{\emph{IEEE Transactions on Pattern Analysis and Machine Intelligence}} (\bibinfo{year}{2023}).
\newblock


\bibitem[Dash et~al\mbox{.}(2023)]%
        {dash2023review}
\bibfield{author}{\bibinfo{person}{Ankan Dash}, \bibinfo{person}{Junyi Ye}, {and} \bibinfo{person}{Guiling Wang}.} \bibinfo{year}{2023}\natexlab{}.
\newblock \showarticletitle{A review of Generative Adversarial Networks (GANs) and its applications in a wide variety of disciplines: From Medical to Remote Sensing}.
\newblock \bibinfo{journal}{\emph{IEEE Access}} (\bibinfo{year}{2023}).
\newblock


\bibitem[De~Cristofaro(2024)]%
        {de2024synthetic}
\bibfield{author}{\bibinfo{person}{Emiliano De~Cristofaro}.} \bibinfo{year}{2024}\natexlab{}.
\newblock \showarticletitle{Synthetic Data: Methods, Use Cases, and Risks}.
\newblock \bibinfo{journal}{\emph{IEEE Security \& Privacy}} (\bibinfo{year}{2024}).
\newblock


\bibitem[de~Goede(2023)]%
        {de2023training}
\bibfield{author}{\bibinfo{person}{Matthijs de Goede}.} \bibinfo{year}{2023}\natexlab{}.
\newblock \bibinfo{title}{Training diffusion models with federated learning: A communication-efficient model for cross-silo federated image generation}.
\newblock \bibinfo{howpublished}{\url{https://repository.tudelft.nl/islandora/object/uuid:49e11cf3-5a0a-40bc-9a62-1d7fe05fbe4d}}.
\newblock


\bibitem[de~Souza et~al\mbox{.}(2023)]%
        {de2023review}
\bibfield{author}{\bibinfo{person}{Vinicius Luis~Trevisan de Souza}, \bibinfo{person}{Bruno Augusto~Dorta Marques}, \bibinfo{person}{Harlen~Costa Batagelo}, {and} \bibinfo{person}{Jo{\~a}o~Paulo Gois}.} \bibinfo{year}{2023}\natexlab{}.
\newblock \showarticletitle{A review on generative adversarial networks for image generation}.
\newblock \bibinfo{journal}{\emph{Computers \& Graphics}} (\bibinfo{year}{2023}).
\newblock


\bibitem[Ding et~al\mbox{.}(2023)]%
        {ding2023combining}
\bibfield{author}{\bibinfo{person}{Xuanang Ding}, \bibinfo{person}{Guohui Li}, \bibinfo{person}{Ling Yuan}, \bibinfo{person}{Lu Zhang}, {and} \bibinfo{person}{Qian Rong}.} \bibinfo{year}{2023}\natexlab{}.
\newblock \showarticletitle{Combining Autoencoder with Adaptive Differential Privacy for Federated Collaborative Filtering}. In \bibinfo{booktitle}{\emph{International Conference on Database Systems for Advanced Applications}}. Springer, \bibinfo{pages}{661--676}.
\newblock


\bibitem[Duan et~al\mbox{.}(2021)]%
        {duan2021feddna}
\bibfield{author}{\bibinfo{person}{Jian-Hui Duan}, \bibinfo{person}{Wenzhong Li}, {and} \bibinfo{person}{Sanglu Lu}.} \bibinfo{year}{2021}\natexlab{}.
\newblock \showarticletitle{FedDNA: Federated learning with decoupled normalization-layer aggregation for non-iid data}. In \bibinfo{booktitle}{\emph{Machine Learning and Knowledge Discovery in Databases. Research Track: European Conference, ECML PKDD 2021, Bilbao, Spain, September 13--17, 2021, Proceedings, Part I 21}}. Springer, \bibinfo{pages}{722--737}.
\newblock


\bibitem[Duan et~al\mbox{.}(2023)]%
        {duan2023federated}
\bibfield{author}{\bibinfo{person}{Jian-hui Duan}, \bibinfo{person}{Wenzhong Li}, \bibinfo{person}{Derun Zou}, \bibinfo{person}{Ruichen Li}, {and} \bibinfo{person}{Sanglu Lu}.} \bibinfo{year}{2023}\natexlab{}.
\newblock \showarticletitle{Federated Learning With Data-Agnostic Distribution Fusion}. In \bibinfo{booktitle}{\emph{Proceedings of the IEEE/CVF Conference on Computer Vision and Pattern Recognition}}. \bibinfo{pages}{8074--8083}.
\newblock


\bibitem[Duan et~al\mbox{.}(2022a)]%
        {duan2022fedTDA}
\bibfield{author}{\bibinfo{person}{Shaoming Duan}, \bibinfo{person}{Chuanyi Liu}, \bibinfo{person}{Peiyi Han}, \bibinfo{person}{Tianyu He}, \bibinfo{person}{Yifeng Xu}, {and} \bibinfo{person}{Qiyuan Deng}.} \bibinfo{year}{2022}\natexlab{a}.
\newblock \showarticletitle{Fed-TDA: Federated Tabular Data Augmentation on Non-IID Data}.
\newblock \bibinfo{journal}{\emph{arXiv:2211.13116}} (\bibinfo{year}{2022}).
\newblock


\bibitem[Duan et~al\mbox{.}(2022b)]%
        {duan2022htfedgan}
\bibfield{author}{\bibinfo{person}{Shaoming Duan}, \bibinfo{person}{Chuanyi Liu}, \bibinfo{person}{Peiyi Han}, \bibinfo{person}{Xiaopeng Jin}, \bibinfo{person}{Xinyi Zhang}, \bibinfo{person}{Tianyu He}, \bibinfo{person}{Hezhong Pan}, {and} \bibinfo{person}{Xiayu Xiang}.} \bibinfo{year}{2022}\natexlab{b}.
\newblock \showarticletitle{HT-Fed-GAN: Federated Generative Model for Decentralized Tabular Data Synthesis}.
\newblock \bibinfo{journal}{\emph{Entropy}} \bibinfo{volume}{25}, \bibinfo{number}{1} (\bibinfo{year}{2022}), \bibinfo{pages}{88}.
\newblock


\bibitem[Dugdale(2023)]%
        {dugdale2023federated}
\bibfield{author}{\bibinfo{person}{Hugo Dugdale}.} \bibinfo{year}{2023}\natexlab{}.
\newblock \bibinfo{title}{Federated Learning with Variational Autoencoders}.
\newblock \bibinfo{howpublished}{\url{https://openreview.net/forum?id=mvo72yTjhTl}}.
\newblock


\bibitem[Dwork et~al\mbox{.}(2006)]%
        {dwork2006calibrating}
\bibfield{author}{\bibinfo{person}{Cynthia Dwork}, \bibinfo{person}{Frank McSherry}, \bibinfo{person}{Kobbi Nissim}, {and} \bibinfo{person}{Adam Smith}.} \bibinfo{year}{2006}\natexlab{}.
\newblock \showarticletitle{Calibrating noise to sensitivity in private data analysis}. In \bibinfo{booktitle}{\emph{Theory of Cryptography: Third Theory of Cryptography Conference, TCC 2006, New York, NY, USA, March 4-7, 2006. Proceedings 3}}. Springer, \bibinfo{pages}{265--284}.
\newblock


\bibitem[Dwork et~al\mbox{.}(2014)]%
        {differntialprivacy2014dwork}
\bibfield{author}{\bibinfo{person}{Cynthia Dwork}, \bibinfo{person}{Aaron Roth}, {et~al\mbox{.}}} \bibinfo{year}{2014}\natexlab{}.
\newblock \showarticletitle{The algorithmic foundations of differential privacy}.
\newblock \bibinfo{journal}{\emph{Foundations and Trends{\textregistered} in Theoretical Computer Science}} \bibinfo{volume}{9}, \bibinfo{number}{3--4} (\bibinfo{year}{2014}), \bibinfo{pages}{211--407}.
\newblock


\bibitem[Ekblom et~al\mbox{.}(2022)]%
        {ekblom2022effgan}
\bibfield{author}{\bibinfo{person}{Ebba Ekblom}, \bibinfo{person}{Edvin~Listo Zec}, {and} \bibinfo{person}{Olof Mogren}.} \bibinfo{year}{2022}\natexlab{}.
\newblock \showarticletitle{EFFGAN: Ensembles of fine-tuned federated GANs}. In \bibinfo{booktitle}{\emph{2022 IEEE International Conference on Big Data (Big Data)}}. IEEE, \bibinfo{pages}{884--892}.
\newblock


\bibitem[Fang et~al\mbox{.}(2020)]%
        {fang2020local}
\bibfield{author}{\bibinfo{person}{Minghong Fang}, \bibinfo{person}{Xiaoyu Cao}, \bibinfo{person}{Jinyuan Jia}, {and} \bibinfo{person}{Neil Gong}.} \bibinfo{year}{2020}\natexlab{}.
\newblock \showarticletitle{Local model poisoning attacks to $\{$Byzantine-Robust$\}$ federated learning}. In \bibinfo{booktitle}{\emph{29th USENIX security symposium (USENIX Security 20)}}. \bibinfo{pages}{1605--1622}.
\newblock


\bibitem[Geyer et~al\mbox{.}(2017)]%
        {geyer2017differentially}
\bibfield{author}{\bibinfo{person}{Robin~C Geyer}, \bibinfo{person}{Tassilo Klein}, {and} \bibinfo{person}{Moin Nabi}.} \bibinfo{year}{2017}\natexlab{}.
\newblock \showarticletitle{Differentially private federated learning: A client level perspective}.
\newblock \bibinfo{journal}{\emph{arXiv:1712.07557}} (\bibinfo{year}{2017}).
\newblock


\bibitem[Ghavamipour et~al\mbox{.}(2023)]%
        {ghavamipour2023federated}
\bibfield{author}{\bibinfo{person}{Ali~Reza Ghavamipour}, \bibinfo{person}{Fatih Turkmen}, \bibinfo{person}{Rui Wang}, {and} \bibinfo{person}{Kaitai Liang}.} \bibinfo{year}{2023}\natexlab{}.
\newblock \showarticletitle{Federated Synthetic Data Generation with Stronger Security Guarantees}. In \bibinfo{booktitle}{\emph{Proceedings of the 28th ACM Symposium on Access Control Models and Technologies}}. \bibinfo{pages}{31--42}.
\newblock


\bibitem[Ghojogh et~al\mbox{.}(2021)]%
        {ghojogh2021factor}
\bibfield{author}{\bibinfo{person}{Benyamin Ghojogh}, \bibinfo{person}{Ali Ghodsi}, \bibinfo{person}{Fakhri Karray}, {and} \bibinfo{person}{Mark Crowley}.} \bibinfo{year}{2021}\natexlab{}.
\newblock \showarticletitle{Factor analysis, probabilistic principal component analysis, variational inference, and variational autoencoder: Tutorial and survey}.
\newblock \bibinfo{journal}{\emph{arXiv:2101.00734}} (\bibinfo{year}{2021}).
\newblock


\bibitem[Gong and Liu(2018)]%
        {gong2018attribute}
\bibfield{author}{\bibinfo{person}{Neil~Zhenqiang Gong} {and} \bibinfo{person}{Bin Liu}.} \bibinfo{year}{2018}\natexlab{}.
\newblock \showarticletitle{Attribute inference attacks in online social networks}.
\newblock \bibinfo{journal}{\emph{ACM Transactions on Privacy and Security (TOPS)}} \bibinfo{volume}{21}, \bibinfo{number}{1} (\bibinfo{year}{2018}), \bibinfo{pages}{1--30}.
\newblock


\bibitem[Goodfellow et~al\mbox{.}(2020)]%
        {goodfellow2020generative}
\bibfield{author}{\bibinfo{person}{Ian Goodfellow}, \bibinfo{person}{Jean Pouget-Abadie}, \bibinfo{person}{Mehdi Mirza}, \bibinfo{person}{Bing Xu}, \bibinfo{person}{David Warde-Farley}, \bibinfo{person}{Sherjil Ozair}, \bibinfo{person}{Aaron Courville}, {and} \bibinfo{person}{Yoshua Bengio}.} \bibinfo{year}{2020}\natexlab{}.
\newblock \showarticletitle{Generative adversarial networks}.
\newblock \bibinfo{journal}{\emph{Commun. ACM}} \bibinfo{volume}{63}, \bibinfo{number}{11} (\bibinfo{year}{2020}), \bibinfo{pages}{139--144}.
\newblock


\bibitem[Gosselin et~al\mbox{.}(2022)]%
        {gosselin2022privacy}
\bibfield{author}{\bibinfo{person}{R{\'e}mi Gosselin}, \bibinfo{person}{Lo{\"\i}c Vieu}, \bibinfo{person}{Faiza Loukil}, {and} \bibinfo{person}{Alexandre Benoit}.} \bibinfo{year}{2022}\natexlab{}.
\newblock \showarticletitle{Privacy and security in federated learning: A survey}.
\newblock \bibinfo{journal}{\emph{Applied Sciences}} \bibinfo{volume}{12}, \bibinfo{number}{19} (\bibinfo{year}{2022}), \bibinfo{pages}{9901}.
\newblock


\bibitem[Gu et~al\mbox{.}(2022)]%
        {gu2022cs}
\bibfield{author}{\bibinfo{person}{Yuhao Gu}, \bibinfo{person}{Yuebin Bai}, {and} \bibinfo{person}{Shubin Xu}.} \bibinfo{year}{2022}\natexlab{}.
\newblock \showarticletitle{CS-MIA: Membership inference attack based on prediction confidence series in federated learning}.
\newblock \bibinfo{journal}{\emph{Journal of Information Security and Applications}}  \bibinfo{volume}{67} (\bibinfo{year}{2022}), \bibinfo{pages}{103201}.
\newblock


\bibitem[Gu et~al\mbox{.}(2021)]%
        {gu2021frepd}
\bibfield{author}{\bibinfo{person}{Zhipin Gu}, \bibinfo{person}{Liangzhong He}, \bibinfo{person}{Peiyan Li}, \bibinfo{person}{Peng Sun}, \bibinfo{person}{Jiangyong Shi}, {and} \bibinfo{person}{Yuexiang Yang}.} \bibinfo{year}{2021}\natexlab{}.
\newblock \showarticletitle{FREPD: A Robust Federated Learning Framework on Variational Autoencoder.}
\newblock \bibinfo{journal}{\emph{Comput. Syst. Sci. Eng.}} \bibinfo{volume}{39}, \bibinfo{number}{3} (\bibinfo{year}{2021}), \bibinfo{pages}{307--320}.
\newblock


\bibitem[Gu and Yang(2021)]%
        {gu2021detecting}
\bibfield{author}{\bibinfo{person}{Zhipin Gu} {and} \bibinfo{person}{Yuexiang Yang}.} \bibinfo{year}{2021}\natexlab{}.
\newblock \showarticletitle{Detecting malicious model updates from federated learning on conditional variational autoencoder}. In \bibinfo{booktitle}{\emph{2021 IEEE international parallel and distributed processing symposium (IPDPS)}}. IEEE, \bibinfo{pages}{671--680}.
\newblock


\bibitem[Guendouzi et~al\mbox{.}(2023)]%
        {guendouzi2023systematic}
\bibfield{author}{\bibinfo{person}{Badra~Souhila Guendouzi}, \bibinfo{person}{Samir Ouchani}, \bibinfo{person}{Hiba~EL Assaad}, {and} \bibinfo{person}{Madeleine~EL Zaher}.} \bibinfo{year}{2023}\natexlab{}.
\newblock \showarticletitle{A systematic review of federated learning: Challenges, aggregation methods, and development tools}.
\newblock \bibinfo{journal}{\emph{Journal of Network and Computer Applications}} (\bibinfo{year}{2023}), \bibinfo{pages}{103714}.
\newblock


\bibitem[Guerraoui et~al\mbox{.}(2020)]%
        {guerraoui2020fegan}
\bibfield{author}{\bibinfo{person}{Rachid Guerraoui}, \bibinfo{person}{Arsany Guirguis}, \bibinfo{person}{Anne-Marie Kermarrec}, {and} \bibinfo{person}{Erwan~Le Merrer}.} \bibinfo{year}{2020}\natexlab{}.
\newblock \showarticletitle{Fegan: Scaling distributed gans}. In \bibinfo{booktitle}{\emph{Proceedings of the 21st International Middleware Conference}}. \bibinfo{pages}{193--206}.
\newblock


\bibitem[Gui et~al\mbox{.}(2021)]%
        {gui2021review}
\bibfield{author}{\bibinfo{person}{Jie Gui}, \bibinfo{person}{Zhenan Sun}, \bibinfo{person}{Yonggang Wen}, \bibinfo{person}{Dacheng Tao}, {and} \bibinfo{person}{Jieping Ye}.} \bibinfo{year}{2021}\natexlab{}.
\newblock \showarticletitle{A review on generative adversarial networks: Algorithms, theory, and applications}.
\newblock \bibinfo{journal}{\emph{IEEE transactions on knowledge and data engineering}} \bibinfo{volume}{35}, \bibinfo{number}{4} (\bibinfo{year}{2021}), \bibinfo{pages}{3313--3332}.
\newblock


\bibitem[Guo(2023)]%
        {guo2023enhancefl}
\bibfield{author}{\bibinfo{person}{Zihao Guo}.} \bibinfo{year}{2023}\natexlab{}.
\newblock \showarticletitle{Enhancing Federated Learning Efficiency with Generative Model-Based Data Augmentation for Non-IID Data}. In \bibinfo{booktitle}{\emph{2023 European Conference on Communication Systems (ECCS)}}. \bibinfo{pages}{24--28}.
\newblock
\urldef\tempurl%
\url{https://doi.org/10.1109/ECCS58882.2023.00013}
\showDOI{\tempurl}


\bibitem[Ha and Dang(2022)]%
        {ha2022inference}
\bibfield{author}{\bibinfo{person}{Trung Ha} {and} \bibinfo{person}{Tran~Khanh Dang}.} \bibinfo{year}{2022}\natexlab{}.
\newblock \showarticletitle{Inference attacks based on GAN in federated learning}.
\newblock \bibinfo{journal}{\emph{International Journal of Web Information Systems}} \bibinfo{volume}{18}, \bibinfo{number}{2/3} (\bibinfo{year}{2022}), \bibinfo{pages}{117--136}.
\newblock


\bibitem[Han and Guan(2023)]%
        {han2023gan}
\bibfield{author}{\bibinfo{person}{Yujin Han} {and} \bibinfo{person}{Leying Guan}.} \bibinfo{year}{2023}\natexlab{}.
\newblock \showarticletitle{GAN-based federated learning for label protection in binary classification}.
\newblock \bibinfo{journal}{\emph{arXiv:2302.02245}} (\bibinfo{year}{2023}).
\newblock


\bibitem[Hardy et~al\mbox{.}(2019)]%
        {mdgan2019hardy}
\bibfield{author}{\bibinfo{person}{Corentin Hardy}, \bibinfo{person}{Erwan Le~Merrer}, {and} \bibinfo{person}{Bruno Sericola}.} \bibinfo{year}{2019}\natexlab{}.
\newblock \showarticletitle{Md-gan: Multi-discriminator generative adversarial networks for distributed datasets}. In \bibinfo{booktitle}{\emph{2019 IEEE international parallel and distributed processing symposium (IPDPS)}}. IEEE, \bibinfo{pages}{866--877}.
\newblock


\bibitem[Harshvardhan et~al\mbox{.}(2020)]%
        {harshvardhan2020comprehensive}
\bibfield{author}{\bibinfo{person}{GM Harshvardhan}, \bibinfo{person}{Mahendra~Kumar Gourisaria}, \bibinfo{person}{Manjusha Pandey}, {and} \bibinfo{person}{Siddharth~Swarup Rautaray}.} \bibinfo{year}{2020}\natexlab{}.
\newblock \showarticletitle{A comprehensive survey and analysis of generative models in machine learning}.
\newblock \bibinfo{journal}{\emph{Computer Science Review}}  \bibinfo{volume}{38} (\bibinfo{year}{2020}), \bibinfo{pages}{100285}.
\newblock


\bibitem[Hatamizadeh et~al\mbox{.}(2023)]%
        {hatamizadeh2023gradient}
\bibfield{author}{\bibinfo{person}{Ali Hatamizadeh}, \bibinfo{person}{Hongxu Yin}, \bibinfo{person}{Pavlo Molchanov}, \bibinfo{person}{Andriy Myronenko}, \bibinfo{person}{Wenqi Li}, \bibinfo{person}{Prerna Dogra}, \bibinfo{person}{Andrew Feng}, \bibinfo{person}{Mona~G Flores}, \bibinfo{person}{Jan Kautz}, \bibinfo{person}{Daguang Xu}, {et~al\mbox{.}}} \bibinfo{year}{2023}\natexlab{}.
\newblock \showarticletitle{Do gradient inversion attacks make federated learning unsafe?}
\newblock \bibinfo{journal}{\emph{IEEE Transactions on Medical Imaging}} (\bibinfo{year}{2023}).
\newblock


\bibitem[Heinbaugh et~al\mbox{.}(2023)]%
        {heinbaugh2023datafree}
\bibfield{author}{\bibinfo{person}{Clare~Elizabeth Heinbaugh}, \bibinfo{person}{Emilio Luz-Ricca}, {and} \bibinfo{person}{Huajie Shao}.} \bibinfo{year}{2023}\natexlab{}.
\newblock \showarticletitle{Data-Free One-Shot Federated Learning Under Very High Statistical Heterogeneity}. In \bibinfo{booktitle}{\emph{The Eleventh International Conference on Learning Representations}}.
\newblock
\urldef\tempurl%
\url{https://openreview.net/forum?id=_hb4vM3jspB}
\showURL{%
\tempurl}


\bibitem[Ho et~al\mbox{.}(2020)]%
        {ho2020denoising}
\bibfield{author}{\bibinfo{person}{Jonathan Ho}, \bibinfo{person}{Ajay Jain}, {and} \bibinfo{person}{Pieter Abbeel}.} \bibinfo{year}{2020}\natexlab{}.
\newblock \showarticletitle{Denoising diffusion probabilistic models}.
\newblock \bibinfo{journal}{\emph{Advances in neural information processing systems}}  \bibinfo{volume}{33} (\bibinfo{year}{2020}), \bibinfo{pages}{6840--6851}.
\newblock


\bibitem[Huang et~al\mbox{.}(2022)]%
        {huang2022cross}
\bibfield{author}{\bibinfo{person}{Chao Huang}, \bibinfo{person}{Jianwei Huang}, {and} \bibinfo{person}{Xin Liu}.} \bibinfo{year}{2022}\natexlab{}.
\newblock \showarticletitle{Cross-silo federated learning: Challenges and opportunities}.
\newblock \bibinfo{journal}{\emph{arXiv:2206.12949}} (\bibinfo{year}{2022}).
\newblock


\bibitem[Huang et~al\mbox{.}(2021a)]%
        {huang2021personalized}
\bibfield{author}{\bibinfo{person}{Yutao Huang}, \bibinfo{person}{Lingyang Chu}, \bibinfo{person}{Zirui Zhou}, \bibinfo{person}{Lanjun Wang}, \bibinfo{person}{Jiangchuan Liu}, \bibinfo{person}{Jian Pei}, {and} \bibinfo{person}{Yong Zhang}.} \bibinfo{year}{2021}\natexlab{a}.
\newblock \showarticletitle{Personalized cross-silo federated learning on non-iid data}. In \bibinfo{booktitle}{\emph{Proceedings of the AAAI conference on artificial intelligence}}. \bibinfo{pages}{7865--7873}.
\newblock


\bibitem[Huang et~al\mbox{.}(2021b)]%
        {huang2021evaluating}
\bibfield{author}{\bibinfo{person}{Yangsibo Huang}, \bibinfo{person}{Samyak Gupta}, \bibinfo{person}{Zhao Song}, \bibinfo{person}{Kai Li}, {and} \bibinfo{person}{Sanjeev Arora}.} \bibinfo{year}{2021}\natexlab{b}.
\newblock \showarticletitle{Evaluating gradient inversion attacks and defenses in federated learning}.
\newblock \bibinfo{journal}{\emph{Advances in Neural Information Processing Systems}}  \bibinfo{volume}{34} (\bibinfo{year}{2021}), \bibinfo{pages}{7232--7241}.
\newblock


\bibitem[Jiang et~al\mbox{.}(2023)]%
        {jiang2023fedvae}
\bibfield{author}{\bibinfo{person}{Yuchen Jiang}, \bibinfo{person}{Ying Wu}, \bibinfo{person}{Shiyao Zhang}, {and} \bibinfo{person}{JQ James}.} \bibinfo{year}{2023}\natexlab{}.
\newblock \showarticletitle{FedVAE: Trajectory privacy preserving based on Federated Variational AutoEncoder}. In \bibinfo{booktitle}{\emph{2023 IEEE 98th Vehicular Technology Conference (VTC2023-Fall)}}. IEEE, \bibinfo{pages}{1--7}.
\newblock


\bibitem[Jin and Li(2022)]%
        {jin2022backdoor}
\bibfield{author}{\bibinfo{person}{Ruinan Jin} {and} \bibinfo{person}{Xiaoxiao Li}.} \bibinfo{year}{2022}\natexlab{}.
\newblock \showarticletitle{Backdoor attack is a devil in federated GAN-based medical image synthesis}. In \bibinfo{booktitle}{\emph{International Workshop on Simulation and Synthesis in Medical Imaging}}. Springer, \bibinfo{pages}{154--165}.
\newblock


\bibitem[Jin and Li(2023)]%
        {jin2023backdoor}
\bibfield{author}{\bibinfo{person}{Ruinan Jin} {and} \bibinfo{person}{Xiaoxiao Li}.} \bibinfo{year}{2023}\natexlab{}.
\newblock \showarticletitle{Backdoor attack and defense in federated generative adversarial network-based medical image synthesis}.
\newblock \bibinfo{journal}{\emph{Medical Image Analysis}}  \bibinfo{volume}{90} (\bibinfo{year}{2023}), \bibinfo{pages}{102965}.
\newblock


\bibitem[Jothiraj(2023)]%
        {jothiraj2023phoenix}
\bibfield{author}{\bibinfo{person}{Fiona Victoria~Stanley Jothiraj}.} \bibinfo{year}{2023}\natexlab{}.
\newblock \emph{\bibinfo{title}{Phoenix: Federated Learning for Generative Diffusion Model}}.
\newblock \bibinfo{thesistype}{Ph.\,D. Dissertation}. \bibinfo{school}{University of Washington}.
\newblock


\bibitem[Kairouz et~al\mbox{.}(2021)]%
        {kairouz2021advances}
\bibfield{author}{\bibinfo{person}{Peter Kairouz}, \bibinfo{person}{H~Brendan McMahan}, \bibinfo{person}{Brendan Avent}, \bibinfo{person}{Aur{\'e}lien Bellet}, \bibinfo{person}{Mehdi Bennis}, \bibinfo{person}{Arjun~Nitin Bhagoji}, \bibinfo{person}{Kallista Bonawitz}, \bibinfo{person}{Zachary Charles}, \bibinfo{person}{Graham Cormode}, \bibinfo{person}{Rachel Cummings}, {et~al\mbox{.}}} \bibinfo{year}{2021}\natexlab{}.
\newblock \showarticletitle{Advances and open problems in federated learning}.
\newblock \bibinfo{journal}{\emph{Foundations and Trends{\textregistered} in Machine Learning}} \bibinfo{volume}{14}, \bibinfo{number}{1--2} (\bibinfo{year}{2021}), \bibinfo{pages}{1--210}.
\newblock


\bibitem[Karimireddy et~al\mbox{.}(2021)]%
        {karimireddy2021breaking}
\bibfield{author}{\bibinfo{person}{Sai~Praneeth Karimireddy}, \bibinfo{person}{Martin Jaggi}, \bibinfo{person}{Satyen Kale}, \bibinfo{person}{Mehryar Mohri}, \bibinfo{person}{Sashank Reddi}, \bibinfo{person}{Sebastian~U Stich}, {and} \bibinfo{person}{Ananda~Theertha Suresh}.} \bibinfo{year}{2021}\natexlab{}.
\newblock \showarticletitle{Breaking the centralized barrier for cross-device federated learning}.
\newblock \bibinfo{journal}{\emph{Advances in Neural Information Processing Systems}}  \bibinfo{volume}{34} (\bibinfo{year}{2021}), \bibinfo{pages}{28663--28676}.
\newblock


\bibitem[Kaspour and Yassine(2023)]%
        {kaspour2023variational}
\bibfield{author}{\bibinfo{person}{Shamisa Kaspour} {and} \bibinfo{person}{Abdulsalam Yassine}.} \bibinfo{year}{2023}\natexlab{}.
\newblock \showarticletitle{Variational Auto-Encoder Model and Federated Approach for Non-Intrusive Load Monitoring in Smart Homes}. In \bibinfo{booktitle}{\emph{2023 IEEE Symposium on Computers and Communications (ISCC)}}. IEEE, \bibinfo{pages}{1110--1115}.
\newblock


\bibitem[Kingma et~al\mbox{.}(2021)]%
        {kingma2021variational}
\bibfield{author}{\bibinfo{person}{Diederik Kingma}, \bibinfo{person}{Tim Salimans}, \bibinfo{person}{Ben Poole}, {and} \bibinfo{person}{Jonathan Ho}.} \bibinfo{year}{2021}\natexlab{}.
\newblock \showarticletitle{Variational diffusion models}.
\newblock \bibinfo{journal}{\emph{Advances in neural information processing systems}}  \bibinfo{volume}{34} (\bibinfo{year}{2021}), \bibinfo{pages}{21696--21707}.
\newblock


\bibitem[Kos et~al\mbox{.}(2018)]%
        {kos2018adversarial}
\bibfield{author}{\bibinfo{person}{Jernej Kos}, \bibinfo{person}{Ian Fischer}, {and} \bibinfo{person}{Dawn Song}.} \bibinfo{year}{2018}\natexlab{}.
\newblock \showarticletitle{Adversarial examples for generative models}. In \bibinfo{booktitle}{\emph{2018 ieee security and privacy workshops (spw)}}. IEEE, \bibinfo{pages}{36--42}.
\newblock


\bibitem[Li et~al\mbox{.}(2023b)]%
        {li2023synthetic}
\bibfield{author}{\bibinfo{person}{Bo Li}, \bibinfo{person}{Yasin Esfandiari}, \bibinfo{person}{Mikkel~N Schmidt}, \bibinfo{person}{Tommy~S Alstr{\o}m}, {and} \bibinfo{person}{Sebastian~U Stich}.} \bibinfo{year}{2023}\natexlab{b}.
\newblock \showarticletitle{Synthetic data shuffling accelerates the convergence of federated learning under data heterogeneity}.
\newblock \bibinfo{journal}{\emph{arXiv:2306.13263}} (\bibinfo{year}{2023}).
\newblock


\bibitem[Li et~al\mbox{.}(2023c)]%
        {li2023review}
\bibfield{author}{\bibinfo{person}{Hao Li}, \bibinfo{person}{Chengcheng Li}, \bibinfo{person}{Jian Wang}, \bibinfo{person}{Aimin Yang}, \bibinfo{person}{Zezhong Ma}, \bibinfo{person}{Zunqian Zhang}, {and} \bibinfo{person}{Dianbo Hua}.} \bibinfo{year}{2023}\natexlab{c}.
\newblock \showarticletitle{Review on security of federated learning and its application in healthcare}.
\newblock \bibinfo{journal}{\emph{Future Generation Computer Systems}}  \bibinfo{volume}{144} (\bibinfo{year}{2023}), \bibinfo{pages}{271--290}.
\newblock


\bibitem[Li et~al\mbox{.}(2020a)]%
        {li2020review}
\bibfield{author}{\bibinfo{person}{Li Li}, \bibinfo{person}{Yuxi Fan}, \bibinfo{person}{Mike Tse}, {and} \bibinfo{person}{Kuo-Yi Lin}.} \bibinfo{year}{2020}\natexlab{a}.
\newblock \showarticletitle{A review of applications in federated learning}.
\newblock \bibinfo{journal}{\emph{Computers \& Industrial Engineering}}  \bibinfo{volume}{149} (\bibinfo{year}{2020}), \bibinfo{pages}{106854}.
\newblock


\bibitem[Li et~al\mbox{.}(2023e)]%
        {li2023distvae}
\bibfield{author}{\bibinfo{person}{Li Li}, \bibinfo{person}{Jianbing Xiahou}, \bibinfo{person}{Fan Lin}, {and} \bibinfo{person}{Songzhi Su}.} \bibinfo{year}{2023}\natexlab{e}.
\newblock \showarticletitle{DistVAE: Distributed Variational Autoencoder for sequential recommendation}.
\newblock \bibinfo{journal}{\emph{Knowledge-Based Systems}}  \bibinfo{volume}{264} (\bibinfo{year}{2023}), \bibinfo{pages}{110313}.
\newblock


\bibitem[Li et~al\mbox{.}(2023d)]%
        {li2023comprehensive}
\bibfield{author}{\bibinfo{person}{Pengzhi Li}, \bibinfo{person}{Yan Pei}, {and} \bibinfo{person}{Jianqiang Li}.} \bibinfo{year}{2023}\natexlab{d}.
\newblock \showarticletitle{A comprehensive survey on design and application of autoencoder in deep learning}.
\newblock \bibinfo{journal}{\emph{Applied Soft Computing}} (\bibinfo{year}{2023}), \bibinfo{pages}{110176}.
\newblock


\bibitem[Li et~al\mbox{.}(2020b)]%
        {li2020federated}
\bibfield{author}{\bibinfo{person}{Tian Li}, \bibinfo{person}{Anit~Kumar Sahu}, \bibinfo{person}{Manzil Zaheer}, \bibinfo{person}{Maziar Sanjabi}, \bibinfo{person}{Ameet Talwalkar}, {and} \bibinfo{person}{Virginia Smith}.} \bibinfo{year}{2020}\natexlab{b}.
\newblock \showarticletitle{Federated optimization in heterogeneous networks}.
\newblock \bibinfo{journal}{\emph{Proceedings of Machine learning and systems}}  \bibinfo{volume}{2} (\bibinfo{year}{2020}), \bibinfo{pages}{429--450}.
\newblock


\bibitem[Li et~al\mbox{.}(2023a)]%
        {IFLGAN2023Li}
\bibfield{author}{\bibinfo{person}{Wei Li}, \bibinfo{person}{Jinlin Chen}, \bibinfo{person}{Zhenyu Wang}, \bibinfo{person}{Zhidong Shen}, \bibinfo{person}{Chao Ma}, {and} \bibinfo{person}{Xiaohui Cui}.} \bibinfo{year}{2023}\natexlab{a}.
\newblock \showarticletitle{IFL-GAN: Improved Federated Learning Generative Adversarial Network With Maximum Mean Discrepancy Model Aggregation}.
\newblock \bibinfo{journal}{\emph{IEEE Transactions on Neural Networks and Learning Systems}} \bibinfo{volume}{34}, \bibinfo{number}{12} (\bibinfo{year}{2023}), \bibinfo{pages}{10502--10515}.
\newblock
\urldef\tempurl%
\url{https://doi.org/10.1109/TNNLS.2022.3167482}
\showDOI{\tempurl}


\bibitem[Li et~al\mbox{.}(2022)]%
        {li2022federated}
\bibfield{author}{\bibinfo{person}{Zijian Li}, \bibinfo{person}{Jiawei Shao}, \bibinfo{person}{Yuyi Mao}, \bibinfo{person}{Jessie~Hui Wang}, {and} \bibinfo{person}{Jun Zhang}.} \bibinfo{year}{2022}\natexlab{}.
\newblock \showarticletitle{Federated learning with gan-based data synthesis for non-iid clients}. In \bibinfo{booktitle}{\emph{International Workshop on Trustworthy Federated Learning}}. Springer, \bibinfo{pages}{17--32}.
\newblock


\bibitem[Little et~al\mbox{.}(2023)]%
        {little2023federated}
\bibfield{author}{\bibinfo{person}{Claire Little}, \bibinfo{person}{Mark Elliot}, {and} \bibinfo{person}{Richard Allmendinger}.} \bibinfo{year}{2023}\natexlab{}.
\newblock \showarticletitle{Federated learning for generating synthetic data: a scoping review}.
\newblock \bibinfo{journal}{\emph{International Journal of Population Data Science}} \bibinfo{volume}{8}, \bibinfo{number}{1} (\bibinfo{year}{2023}).
\newblock


\bibitem[Liu et~al\mbox{.}(2022)]%
        {liu2022vertical}
\bibfield{author}{\bibinfo{person}{Yang Liu}, \bibinfo{person}{Yan Kang}, \bibinfo{person}{Tianyuan Zou}, \bibinfo{person}{Yanhong Pu}, \bibinfo{person}{Yuanqin He}, \bibinfo{person}{Xiaozhou Ye}, \bibinfo{person}{Ye Ouyang}, \bibinfo{person}{Ya-Qin Zhang}, {and} \bibinfo{person}{Qiang Yang}.} \bibinfo{year}{2022}\natexlab{}.
\newblock \showarticletitle{Vertical federated learning}.
\newblock \bibinfo{journal}{\emph{arXiv:2211.12814}} (\bibinfo{year}{2022}).
\newblock


\bibitem[Liu et~al\mbox{.}(2024)]%
        {liu2024vertical}
\bibfield{author}{\bibinfo{person}{Yang Liu}, \bibinfo{person}{Yan Kang}, \bibinfo{person}{Tianyuan Zou}, \bibinfo{person}{Yanhong Pu}, \bibinfo{person}{Yuanqin He}, \bibinfo{person}{Xiaozhou Ye}, \bibinfo{person}{Ye Ouyang}, \bibinfo{person}{Ya-Qin Zhang}, {and} \bibinfo{person}{Qiang Yang}.} \bibinfo{year}{2024}\natexlab{}.
\newblock \showarticletitle{Vertical Federated Learning: Concepts, Advances, and Challenges}.
\newblock \bibinfo{journal}{\emph{IEEE Transactions on Knowledge and Data Engineering}} (\bibinfo{year}{2024}).
\newblock


\bibitem[Lomurno et~al\mbox{.}(2022)]%
        {lomurno2022sgde}
\bibfield{author}{\bibinfo{person}{Eugenio Lomurno}, \bibinfo{person}{Alberto Archetti}, \bibinfo{person}{Lorenzo Cazzella}, \bibinfo{person}{Stefano Samele}, \bibinfo{person}{Leonardo Di~Perna}, {and} \bibinfo{person}{Matteo Matteucci}.} \bibinfo{year}{2022}\natexlab{}.
\newblock \showarticletitle{SGDE: Secure generative data exchange for cross-silo federated learning}. In \bibinfo{booktitle}{\emph{Proceedings of the 2022 5th International Conference on Artificial Intelligence and Pattern Recognition}}. \bibinfo{pages}{205--214}.
\newblock


\bibitem[Lyu and Chen(2021)]%
        {lyu2021novel}
\bibfield{author}{\bibinfo{person}{Lingjuan Lyu} {and} \bibinfo{person}{Chen Chen}.} \bibinfo{year}{2021}\natexlab{}.
\newblock \showarticletitle{A novel attribute reconstruction attack in federated learning}.
\newblock \bibinfo{journal}{\emph{arXiv:2108.06910}} (\bibinfo{year}{2021}).
\newblock


\bibitem[Lyu et~al\mbox{.}(2020)]%
        {lyu2020threats}
\bibfield{author}{\bibinfo{person}{Lingjuan Lyu}, \bibinfo{person}{Han Yu}, \bibinfo{person}{Jun Zhao}, {and} \bibinfo{person}{Qiang Yang}.} \bibinfo{year}{2020}\natexlab{}.
\newblock \showarticletitle{Threats to federated learning}.
\newblock \bibinfo{journal}{\emph{Federated Learning: Privacy and Incentive}} (\bibinfo{year}{2020}), \bibinfo{pages}{3--16}.
\newblock


\bibitem[Ma et~al\mbox{.}(2022)]%
        {ma2022state}
\bibfield{author}{\bibinfo{person}{Xiaodong Ma}, \bibinfo{person}{Jia Zhu}, \bibinfo{person}{Zhihao Lin}, \bibinfo{person}{Shanxuan Chen}, {and} \bibinfo{person}{Yangjie Qin}.} \bibinfo{year}{2022}\natexlab{}.
\newblock \showarticletitle{A state-of-the-art survey on solving non-IID data in Federated Learning}.
\newblock \bibinfo{journal}{\emph{Future Generation Computer Systems}}  \bibinfo{volume}{135} (\bibinfo{year}{2022}), \bibinfo{pages}{244--258}.
\newblock


\bibitem[Ma et~al\mbox{.}(2023)]%
        {ma2023FLGAN}
\bibfield{author}{\bibinfo{person}{Zhuoran Ma}, \bibinfo{person}{Yang Liu}, \bibinfo{person}{Yinbin Miao}, \bibinfo{person}{Guowen Xu}, \bibinfo{person}{Ximeng Liu}, \bibinfo{person}{Jianfeng Ma}, {and} \bibinfo{person}{Robert~H. Deng}.} \bibinfo{year}{2023}\natexlab{}.
\newblock \showarticletitle{FLGAN: GAN-Based Unbiased FederatedLearning under Non-IID Settings}.
\newblock \bibinfo{journal}{\emph{IEEE Transactions on Knowledge and Data Engineering}} (\bibinfo{year}{2023}), \bibinfo{pages}{1--16}.
\newblock
\urldef\tempurl%
\url{https://doi.org/10.1109/TKDE.2023.3309858}
\showDOI{\tempurl}


\bibitem[Maliakel et~al\mbox{.}(2024)]%
        {maliakel2024fligan}
\bibfield{author}{\bibinfo{person}{Paul~Joe Maliakel}, \bibinfo{person}{Shashikant Ilager}, {and} \bibinfo{person}{Ivona Brandic}.} \bibinfo{year}{2024}\natexlab{}.
\newblock \bibinfo{title}{FLIGAN: Enhancing Federated Learning with Incomplete Data using GAN}.
\newblock
\newblock
\showeprint[arxiv]{2403.16930}~[cs.LG]


\bibitem[Mamun et~al\mbox{.}(2023)]%
        {mamun2023deepmem}
\bibfield{author}{\bibinfo{person}{Md~Abdullah~Al Mamun}, \bibinfo{person}{Quazi~Mishkatul Alam}, \bibinfo{person}{Erfan Shaigani}, \bibinfo{person}{Pedram Zaree}, \bibinfo{person}{Ihsen Alouani}, {and} \bibinfo{person}{Nael Abu-Ghazaleh}.} \bibinfo{year}{2023}\natexlab{}.
\newblock \showarticletitle{DeepMem: ML Models as storage channels and their (mis-) applications}.
\newblock \bibinfo{journal}{\emph{arXiv preprint arXiv:2307.08811}} (\bibinfo{year}{2023}).
\newblock


\bibitem[McMahan et~al\mbox{.}(2017a)]%
        {mcmahan2017communication}
\bibfield{author}{\bibinfo{person}{Brendan McMahan}, \bibinfo{person}{Eider Moore}, \bibinfo{person}{Daniel Ramage}, \bibinfo{person}{Seth Hampson}, {and} \bibinfo{person}{Blaise~Aguera y Arcas}.} \bibinfo{year}{2017}\natexlab{a}.
\newblock \showarticletitle{Communication-efficient learning of deep networks from decentralized data}. In \bibinfo{booktitle}{\emph{Artificial intelligence and statistics}}. PMLR, \bibinfo{pages}{1273--1282}.
\newblock


\bibitem[McMahan et~al\mbox{.}(2017b)]%
        {mcmahan2017learning}
\bibfield{author}{\bibinfo{person}{H~Brendan McMahan}, \bibinfo{person}{Daniel Ramage}, \bibinfo{person}{Kunal Talwar}, {and} \bibinfo{person}{Li Zhang}.} \bibinfo{year}{2017}\natexlab{b}.
\newblock \showarticletitle{Learning differentially private recurrent language models}.
\newblock \bibinfo{journal}{\emph{arXiv:1710.06963}} (\bibinfo{year}{2017}).
\newblock


\bibitem[Megahed and Mohammed(2024)]%
        {megahed2024comprehensive}
\bibfield{author}{\bibinfo{person}{Mohammed Megahed} {and} \bibinfo{person}{Ammar Mohammed}.} \bibinfo{year}{2024}\natexlab{}.
\newblock \showarticletitle{A comprehensive review of generative adversarial networks: Fundamentals, applications, and challenges}.
\newblock \bibinfo{journal}{\emph{Wiley Interdisciplinary Reviews: Computational Statistics}} \bibinfo{volume}{16}, \bibinfo{number}{1} (\bibinfo{year}{2024}), \bibinfo{pages}{e1629}.
\newblock


\bibitem[Mothukuri et~al\mbox{.}(2021)]%
        {mothukuri2021survey}
\bibfield{author}{\bibinfo{person}{Viraaji Mothukuri}, \bibinfo{person}{Reza~M Parizi}, \bibinfo{person}{Seyedamin Pouriyeh}, \bibinfo{person}{Yan Huang}, \bibinfo{person}{Ali Dehghantanha}, {and} \bibinfo{person}{Gautam Srivastava}.} \bibinfo{year}{2021}\natexlab{}.
\newblock \showarticletitle{A survey on security and privacy of federated learning}.
\newblock \bibinfo{journal}{\emph{Future Generation Computer Systems}}  \bibinfo{volume}{115} (\bibinfo{year}{2021}), \bibinfo{pages}{619--640}.
\newblock


\bibitem[Mou et~al\mbox{.}(2023)]%
        {mou2023pfedv}
\bibfield{author}{\bibinfo{person}{Yongli Mou}, \bibinfo{person}{Jiahui Geng}, \bibinfo{person}{Feng Zhou}, \bibinfo{person}{Oya Beyan}, \bibinfo{person}{Chunming Rong}, {and} \bibinfo{person}{Stefan Decker}.} \bibinfo{year}{2023}\natexlab{}.
\newblock \showarticletitle{pFedV: Mitigating Feature Distribution Skewness via Personalized Federated Learning with Variational Distribution Constraints}. In \bibinfo{booktitle}{\emph{Pacific-Asia Conference on Knowledge Discovery and Data Mining}}. Springer, \bibinfo{pages}{283--294}.
\newblock


\bibitem[Mozaffari et~al\mbox{.}(2023)]%
        {mozaffari2023every}
\bibfield{author}{\bibinfo{person}{Hamid Mozaffari}, \bibinfo{person}{Virat Shejwalkar}, {and} \bibinfo{person}{Amir Houmansadr}.} \bibinfo{year}{2023}\natexlab{}.
\newblock \showarticletitle{Every Vote Counts:$\{$Ranking-Based$\}$ Training of Federated Learning to Resist Poisoning Attacks}. In \bibinfo{booktitle}{\emph{32nd USENIX Security Symposium (USENIX Security 23)}}. \bibinfo{pages}{1721--1738}.
\newblock


\bibitem[Naseri et~al\mbox{.}(2022)]%
        {naseri2022cerberus}
\bibfield{author}{\bibinfo{person}{Mohammad Naseri}, \bibinfo{person}{Yufei Han}, \bibinfo{person}{Enrico Mariconti}, \bibinfo{person}{Yun Shen}, \bibinfo{person}{Gianluca Stringhini}, {and} \bibinfo{person}{Emiliano De~Cristofaro}.} \bibinfo{year}{2022}\natexlab{}.
\newblock \showarticletitle{Cerberus: exploring federated prediction of security events}. In \bibinfo{booktitle}{\emph{Proceedings of the 2022 ACM SIGSAC Conference on Computer and Communications Security}}. \bibinfo{pages}{2337--2351}.
\newblock


\bibitem[Naseri et~al\mbox{.}(2020)]%
        {naseri2020local}
\bibfield{author}{\bibinfo{person}{Mohammad Naseri}, \bibinfo{person}{Jamie Hayes}, {and} \bibinfo{person}{Emiliano De~Cristofaro}.} \bibinfo{year}{2020}\natexlab{}.
\newblock \showarticletitle{Local and central differential privacy for robustness and privacy in federated learning}.
\newblock \bibinfo{journal}{\emph{arXiv:2009.03561}} (\bibinfo{year}{2020}).
\newblock


\bibitem[Nasr et~al\mbox{.}(2019)]%
        {nasr2019comprehensive}
\bibfield{author}{\bibinfo{person}{Milad Nasr}, \bibinfo{person}{Reza Shokri}, {and} \bibinfo{person}{Amir Houmansadr}.} \bibinfo{year}{2019}\natexlab{}.
\newblock \showarticletitle{Comprehensive privacy analysis of deep learning: Passive and active white-box inference attacks against centralized and federated learning}. In \bibinfo{booktitle}{\emph{2019 IEEE symposium on security and privacy (SP)}}. IEEE, \bibinfo{pages}{739--753}.
\newblock


\bibitem[Nguyen et~al\mbox{.}(2021)]%
        {nguyen2021federated}
\bibfield{author}{\bibinfo{person}{Dinh~C Nguyen}, \bibinfo{person}{Ming Ding}, \bibinfo{person}{Pubudu~N Pathirana}, \bibinfo{person}{Aruna Seneviratne}, {and} \bibinfo{person}{Albert~Y Zomaya}.} \bibinfo{year}{2021}\natexlab{}.
\newblock \showarticletitle{Federated learning for COVID-19 detection with generative adversarial networks in edge cloud computing}.
\newblock \bibinfo{journal}{\emph{IEEE Internet of Things Journal}} \bibinfo{volume}{9}, \bibinfo{number}{12} (\bibinfo{year}{2021}), \bibinfo{pages}{10257--10271}.
\newblock


\bibitem[Oussidi and Elhassouny(2018)]%
        {oussidi2018deep}
\bibfield{author}{\bibinfo{person}{Achraf Oussidi} {and} \bibinfo{person}{Azeddine Elhassouny}.} \bibinfo{year}{2018}\natexlab{}.
\newblock \showarticletitle{Deep generative models: Survey}. In \bibinfo{booktitle}{\emph{2018 International conference on intelligent systems and computer vision (ISCV)}}. IEEE, \bibinfo{pages}{1--8}.
\newblock


\bibitem[Pandya et~al\mbox{.}(2023)]%
        {pandya2023federated}
\bibfield{author}{\bibinfo{person}{Sharnil Pandya}, \bibinfo{person}{Gautam Srivastava}, \bibinfo{person}{Rutvij Jhaveri}, \bibinfo{person}{M~Rajasekhara Babu}, \bibinfo{person}{Sweta Bhattacharya}, \bibinfo{person}{Praveen Kumar~Reddy Maddikunta}, \bibinfo{person}{Spyridon Mastorakis}, \bibinfo{person}{Md~Jalil Piran}, {and} \bibinfo{person}{Thippa~Reddy Gadekallu}.} \bibinfo{year}{2023}\natexlab{}.
\newblock \showarticletitle{Federated learning for smart cities: A comprehensive survey}.
\newblock \bibinfo{journal}{\emph{Sustainable Energy Technologies and Assessments}}  \bibinfo{volume}{55} (\bibinfo{year}{2023}), \bibinfo{pages}{102987}.
\newblock


\bibitem[Papadopoulos and Karalis(2023)]%
        {papadopoulos2023variational}
\bibfield{author}{\bibinfo{person}{Dimitris Papadopoulos} {and} \bibinfo{person}{Vangelis~D Karalis}.} \bibinfo{year}{2023}\natexlab{}.
\newblock \showarticletitle{Variational Autoencoders for Data Augmentation in Clinical Studies}.
\newblock \bibinfo{journal}{\emph{Applied Sciences}} \bibinfo{volume}{13}, \bibinfo{number}{15} (\bibinfo{year}{2023}), \bibinfo{pages}{8793}.
\newblock


\bibitem[Pejic et~al\mbox{.}(2022)]%
        {pejic2022effect}
\bibfield{author}{\bibinfo{person}{Ignjat Pejic}, \bibinfo{person}{Rui Wang}, {and} \bibinfo{person}{Kaitai Liang}.} \bibinfo{year}{2022}\natexlab{}.
\newblock \showarticletitle{Effect of Homomorphic Encryption on the Performance of Training Federated Learning Generative Adversarial Networks}.
\newblock \bibinfo{journal}{\emph{arXiv:2207.00263}} (\bibinfo{year}{2022}).
\newblock


\bibitem[Pfitzner and Arnrich(2022)]%
        {pfitzner2022dpdfvae}
\bibfield{author}{\bibinfo{person}{Bjarne Pfitzner} {and} \bibinfo{person}{Bert Arnrich}.} \bibinfo{year}{2022}\natexlab{}.
\newblock \showarticletitle{Dpd-fvae: Synthetic data generation using federated variational autoencoders with differentially-private decoder}.
\newblock \bibinfo{journal}{\emph{arXiv:2211.11591}} (\bibinfo{year}{2022}).
\newblock


\bibitem[Pillutla et~al\mbox{.}(2022)]%
        {pillutla2022robust}
\bibfield{author}{\bibinfo{person}{Krishna Pillutla}, \bibinfo{person}{Sham~M Kakade}, {and} \bibinfo{person}{Zaid Harchaoui}.} \bibinfo{year}{2022}\natexlab{}.
\newblock \showarticletitle{Robust aggregation for federated learning}.
\newblock \bibinfo{journal}{\emph{IEEE Transactions on Signal Processing}}  \bibinfo{volume}{70} (\bibinfo{year}{2022}), \bibinfo{pages}{1142--1154}.
\newblock


\bibitem[Prediger et~al\mbox{.}(2023)]%
        {prediger2023collaborative}
\bibfield{author}{\bibinfo{person}{Lukas Prediger}, \bibinfo{person}{Joonas J{\"a}lk{\"o}}, \bibinfo{person}{Antti Honkela}, {and} \bibinfo{person}{Samuel Kaski}.} \bibinfo{year}{2023}\natexlab{}.
\newblock \showarticletitle{Collaborative Learning From Distributed Data With Differentially Private Synthetic Twin Data}.
\newblock \bibinfo{journal}{\emph{arXiv:2308.04755}} (\bibinfo{year}{2023}).
\newblock


\bibitem[Psychogyios et~al\mbox{.}(2023)]%
        {psychogyios2023gan}
\bibfield{author}{\bibinfo{person}{Konstantinos Psychogyios}, \bibinfo{person}{Terpsichori-Helen Velivassaki}, \bibinfo{person}{Stavroula Bourou}, \bibinfo{person}{Artemis Voulkidis}, \bibinfo{person}{Dimitrios Skias}, {and} \bibinfo{person}{Theodore Zahariadis}.} \bibinfo{year}{2023}\natexlab{}.
\newblock \showarticletitle{GAN-Driven Data Poisoning Attacks and Their Mitigation in Federated Learning Systems}.
\newblock \bibinfo{journal}{\emph{Electronics}} \bibinfo{volume}{12}, \bibinfo{number}{8} (\bibinfo{year}{2023}), \bibinfo{pages}{1805}.
\newblock


\bibitem[Qi et~al\mbox{.}(2023)]%
        {qi2023model}
\bibfield{author}{\bibinfo{person}{Pian Qi}, \bibinfo{person}{Diletta Chiaro}, \bibinfo{person}{Antonella Guzzo}, \bibinfo{person}{Michele Ianni}, \bibinfo{person}{Giancarlo Fortino}, {and} \bibinfo{person}{Francesco Piccialli}.} \bibinfo{year}{2023}\natexlab{}.
\newblock \showarticletitle{Model aggregation techniques in federated learning: A comprehensive survey}.
\newblock \bibinfo{journal}{\emph{Future Generation Computer Systems}} (\bibinfo{year}{2023}).
\newblock


\bibitem[Ran et~al\mbox{.}(2024)]%
        {ran2024multi}
\bibfield{author}{\bibinfo{person}{Nian Ran}, \bibinfo{person}{Bahrul~Ilmi Nasution}, \bibinfo{person}{Claire Little}, \bibinfo{person}{Richard Allmendinger}, {and} \bibinfo{person}{Mark Elliot}.} \bibinfo{year}{2024}\natexlab{}.
\newblock \showarticletitle{Multi-objective evolutionary GAN for tabular data synthesis}.
\newblock \bibinfo{journal}{\emph{arXiv:2404.10176}} (\bibinfo{year}{2024}).
\newblock


\bibitem[Rasha et~al\mbox{.}(2023)]%
        {rasha2023federated}
\bibfield{author}{\bibinfo{person}{Al-Huthaifi Rasha}, \bibinfo{person}{Tianrui Li}, \bibinfo{person}{Wei Huang}, \bibinfo{person}{Jin Gu}, {and} \bibinfo{person}{Chongshou Li}.} \bibinfo{year}{2023}\natexlab{}.
\newblock \showarticletitle{Federated learning in smart cities: Privacy and security survey}.
\newblock \bibinfo{journal}{\emph{Information Sciences}} (\bibinfo{year}{2023}).
\newblock


\bibitem[Rasouli et~al\mbox{.}(2020)]%
        {fedgan2020Rasouli}
\bibfield{author}{\bibinfo{person}{Mohammad Rasouli}, \bibinfo{person}{Tao Sun}, {and} \bibinfo{person}{Ram Rajagopal}.} \bibinfo{year}{2020}\natexlab{}.
\newblock \showarticletitle{Fedgan: Federated generative adversarial networks for distributed data}.
\newblock \bibinfo{journal}{\emph{arXiv:2006.07228}} (\bibinfo{year}{2020}).
\newblock


\bibitem[Rauniyar et~al\mbox{.}(2023)]%
        {rauniyar2023federated}
\bibfield{author}{\bibinfo{person}{Ashish Rauniyar}, \bibinfo{person}{Desta~Haileselassie Hagos}, \bibinfo{person}{Debesh Jha}, \bibinfo{person}{Jan~Erik H{\aa}keg{\aa}rd}, \bibinfo{person}{Ulas Bagci}, \bibinfo{person}{Danda~B Rawat}, {and} \bibinfo{person}{Vladimir Vlassov}.} \bibinfo{year}{2023}\natexlab{}.
\newblock \showarticletitle{Federated learning for medical applications: A taxonomy, current trends, challenges, and future research directions}.
\newblock \bibinfo{journal}{\emph{IEEE Internet of Things Journal}} (\bibinfo{year}{2023}).
\newblock


\bibitem[Rehman et~al\mbox{.}(2024)]%
        {rehman2024fedcscd}
\bibfield{author}{\bibinfo{person}{Amir Rehman}, \bibinfo{person}{Huanlai Xing}, \bibinfo{person}{Li Feng}, \bibinfo{person}{Mehboob Hussain}, \bibinfo{person}{Nighat Gulzar}, \bibinfo{person}{Muhammad~Adnan Khan}, \bibinfo{person}{Abid Hussain}, {and} \bibinfo{person}{Dhekra Saeed}.} \bibinfo{year}{2024}\natexlab{}.
\newblock \showarticletitle{FedCSCD-GAN: A secure and collaborative framework for clinical cancer diagnosis via optimized federated learning and GAN}.
\newblock \bibinfo{journal}{\emph{Biomedical Signal Processing and Control}}  \bibinfo{volume}{89} (\bibinfo{year}{2024}), \bibinfo{pages}{105893}.
\newblock


\bibitem[Rodr{\'\i}guez-Barroso et~al\mbox{.}(2023)]%
        {rodriguez2023survey}
\bibfield{author}{\bibinfo{person}{Nuria Rodr{\'\i}guez-Barroso}, \bibinfo{person}{Daniel Jim{\'e}nez-L{\'o}pez}, \bibinfo{person}{M~Victoria Luz{\'o}n}, \bibinfo{person}{Francisco Herrera}, {and} \bibinfo{person}{Eugenio Mart{\'\i}nez-C{\'a}mara}.} \bibinfo{year}{2023}\natexlab{}.
\newblock \showarticletitle{Survey on federated learning threats: Concepts, taxonomy on attacks and defences, experimental study and challenges}.
\newblock \bibinfo{journal}{\emph{Information Fusion}}  \bibinfo{volume}{90} (\bibinfo{year}{2023}), \bibinfo{pages}{148--173}.
\newblock


\bibitem[Ruthotto and Haber(2021)]%
        {ruthotto2021introduction}
\bibfield{author}{\bibinfo{person}{Lars Ruthotto} {and} \bibinfo{person}{Eldad Haber}.} \bibinfo{year}{2021}\natexlab{}.
\newblock \showarticletitle{An introduction to deep generative modeling}.
\newblock \bibinfo{journal}{\emph{GAMM-Mitteilungen}} \bibinfo{volume}{44}, \bibinfo{number}{2} (\bibinfo{year}{2021}), \bibinfo{pages}{e202100008}.
\newblock


\bibitem[Salakhutdinov(2015)]%
        {salakhutdinov2015learning}
\bibfield{author}{\bibinfo{person}{Ruslan Salakhutdinov}.} \bibinfo{year}{2015}\natexlab{}.
\newblock \showarticletitle{Learning deep generative models}.
\newblock \bibinfo{journal}{\emph{Annual Review of Statistics and Its Application}}  \bibinfo{volume}{2} (\bibinfo{year}{2015}), \bibinfo{pages}{361--385}.
\newblock


\bibitem[San-Roman et~al\mbox{.}(2021)]%
        {san2021noise}
\bibfield{author}{\bibinfo{person}{Robin San-Roman}, \bibinfo{person}{Eliya Nachmani}, {and} \bibinfo{person}{Lior Wolf}.} \bibinfo{year}{2021}\natexlab{}.
\newblock \showarticletitle{Noise estimation for generative diffusion models}.
\newblock \bibinfo{journal}{\emph{arXiv:2104.02600}} (\bibinfo{year}{2021}).
\newblock


\bibitem[Sattarov et~al\mbox{.}(2024)]%
        {sattarov2024fedtabdiff}
\bibfield{author}{\bibinfo{person}{Timur Sattarov}, \bibinfo{person}{Marco Schreyer}, {and} \bibinfo{person}{Damian Borth}.} \bibinfo{year}{2024}\natexlab{}.
\newblock \bibinfo{title}{FedTabDiff: Federated Learning of Diffusion Probabilistic Models for Synthetic Mixed-Type Tabular Data Generation}.
\newblock
\newblock
\showeprint[arxiv]{2401.06263}~[cs.LG]


\bibitem[Shao et~al\mbox{.}(2023)]%
        {shao2023survey}
\bibfield{author}{\bibinfo{person}{Jiawei Shao}, \bibinfo{person}{Zijian Li}, \bibinfo{person}{Wenqiang Sun}, \bibinfo{person}{Tailin Zhou}, \bibinfo{person}{Yuchang Sun}, \bibinfo{person}{Lumin Liu}, \bibinfo{person}{Zehong Lin}, {and} \bibinfo{person}{Jun Zhang}.} \bibinfo{year}{2023}\natexlab{}.
\newblock \showarticletitle{A survey of what to share in federated learning: perspectives on model utility, privacy leakage, and communication efficiency}.
\newblock \bibinfo{journal}{\emph{arXiv:2307.10655}} (\bibinfo{year}{2023}).
\newblock


\bibitem[Sharma and Kumar(2023)]%
        {sharma2023federated}
\bibfield{author}{\bibinfo{person}{Sunena Sharma} {and} \bibinfo{person}{Sunil Kumar}.} \bibinfo{year}{2023}\natexlab{}.
\newblock \showarticletitle{Federated Learning Approaches to Diverse Machine Learning Model: A Review}. In \bibinfo{booktitle}{\emph{International Conference on Information and Communication Technology for Intelligent Systems}}. Springer, \bibinfo{pages}{259--269}.
\newblock


\bibitem[Shayegani et~al\mbox{.}(2023)]%
        {shayegani2023survey}
\bibfield{author}{\bibinfo{person}{Erfan Shayegani}, \bibinfo{person}{Md~Abdullah~Al Mamun}, \bibinfo{person}{Yu Fu}, \bibinfo{person}{Pedram Zaree}, \bibinfo{person}{Yue Dong}, {and} \bibinfo{person}{Nael Abu-Ghazaleh}.} \bibinfo{year}{2023}\natexlab{}.
\newblock \showarticletitle{Survey of vulnerabilities in large language models revealed by adversarial attacks}.
\newblock \bibinfo{journal}{\emph{arXiv:2310.10844}} (\bibinfo{year}{2023}).
\newblock


\bibitem[Sikandar et~al\mbox{.}(2023)]%
        {sikandar2023detailed}
\bibfield{author}{\bibinfo{person}{Hira~Shahzadi Sikandar}, \bibinfo{person}{Huda Waheed}, \bibinfo{person}{Sibgha Tahir}, \bibinfo{person}{Saif~UR Malik}, {and} \bibinfo{person}{Waqas Rafique}.} \bibinfo{year}{2023}\natexlab{}.
\newblock \showarticletitle{A Detailed Survey on Federated Learning Attacks and Defenses}.
\newblock \bibinfo{journal}{\emph{Electronics}} \bibinfo{volume}{12}, \bibinfo{number}{2} (\bibinfo{year}{2023}), \bibinfo{pages}{260}.
\newblock


\bibitem[Sun et~al\mbox{.}(2021b)]%
        {sun2021adversarial}
\bibfield{author}{\bibinfo{person}{Hui Sun}, \bibinfo{person}{Tianqing Zhu}, \bibinfo{person}{Zhiqiu Zhang}, \bibinfo{person}{Dawei Jin}, \bibinfo{person}{Ping Xiong}, {and} \bibinfo{person}{Wanlei Zhou}.} \bibinfo{year}{2021}\natexlab{b}.
\newblock \showarticletitle{Adversarial attacks against deep generative models on data: a survey}.
\newblock \bibinfo{journal}{\emph{IEEE Transactions on Knowledge and Data Engineering}} (\bibinfo{year}{2021}).
\newblock


\bibitem[Sun et~al\mbox{.}(2021a)]%
        {sun2021information}
\bibfield{author}{\bibinfo{person}{Yuwei Sun}, \bibinfo{person}{Ng~ST Chong}, {and} \bibinfo{person}{Hideya Ochiai}.} \bibinfo{year}{2021}\natexlab{a}.
\newblock \showarticletitle{Information stealing in federated learning systems based on generative adversarial networks}. In \bibinfo{booktitle}{\emph{2021 IEEE International Conference on Systems, Man, and Cybernetics (SMC)}}. IEEE, \bibinfo{pages}{2749--2754}.
\newblock


\bibitem[Tabassum et~al\mbox{.}(2022)]%
        {fedganids2022tabassum}
\bibfield{author}{\bibinfo{person}{Aliya Tabassum}, \bibinfo{person}{Aiman Erbad}, \bibinfo{person}{Wadha Lebda}, \bibinfo{person}{Amr Mohamed}, {and} \bibinfo{person}{Mohsen Guizani}.} \bibinfo{year}{2022}\natexlab{}.
\newblock \showarticletitle{Fedgan-ids: Privacy-preserving ids using gan and federated learning}.
\newblock \bibinfo{journal}{\emph{Computer Communications}}  \bibinfo{volume}{192} (\bibinfo{year}{2022}), \bibinfo{pages}{299--310}.
\newblock


\bibitem[Tolpegin et~al\mbox{.}(2020)]%
        {tolpegin2020data}
\bibfield{author}{\bibinfo{person}{Vale Tolpegin}, \bibinfo{person}{Stacey Truex}, \bibinfo{person}{Mehmet~Emre Gursoy}, {and} \bibinfo{person}{Ling Liu}.} \bibinfo{year}{2020}\natexlab{}.
\newblock \showarticletitle{Data poisoning attacks against federated learning systems}. In \bibinfo{booktitle}{\emph{Computer Security--ESORICS 2020: 25th European Symposium on Research in Computer Security, ESORICS 2020, Guildford, UK, September 14--18, 2020, Proceedings, Part I 25}}. Springer, \bibinfo{pages}{480--501}.
\newblock


\bibitem[Tran et~al\mbox{.}(2024)]%
        {tran2024personalized}
\bibfield{author}{\bibinfo{person}{Van-Tuan Tran}, \bibinfo{person}{Huy-Hieu Pham}, {and} \bibinfo{person}{Kok-Seng Wong}.} \bibinfo{year}{2024}\natexlab{}.
\newblock \showarticletitle{Personalized privacy-preserving framework for cross-silo federated learning}.
\newblock \bibinfo{journal}{\emph{IEEE Transactions on Emerging Topics in Computing}} (\bibinfo{year}{2024}).
\newblock


\bibitem[Truex et~al\mbox{.}(2020)]%
        {truex2020ldp}
\bibfield{author}{\bibinfo{person}{Stacey Truex}, \bibinfo{person}{Ling Liu}, \bibinfo{person}{Ka-Ho Chow}, \bibinfo{person}{Mehmet~Emre Gursoy}, {and} \bibinfo{person}{Wenqi Wei}.} \bibinfo{year}{2020}\natexlab{}.
\newblock \showarticletitle{LDP-Fed: Federated learning with local differential privacy}. In \bibinfo{booktitle}{\emph{Proceedings of the third ACM international workshop on edge systems, analytics and networking}}. \bibinfo{pages}{61--66}.
\newblock


\bibitem[Tun et~al\mbox{.}(2023)]%
        {tun2023fedDM}
\bibfield{author}{\bibinfo{person}{Ye~Lin Tun}, \bibinfo{person}{Chu~Myaet Thwal}, \bibinfo{person}{Ji~Su Yoon}, \bibinfo{person}{Sun~Moo Kang}, \bibinfo{person}{Chaoning Zhang}, {and} \bibinfo{person}{Choong~Seon Hong}.} \bibinfo{year}{2023}\natexlab{}.
\newblock \showarticletitle{Federated learning with diffusion models for privacy-sensitive vision tasks}. In \bibinfo{booktitle}{\emph{2023 International Conference on Advanced Technologies for Communications (ATC)}}. IEEE, \bibinfo{pages}{305--310}.
\newblock


\bibitem[Veeraragavan and Nyg\r{a}rd(2023)]%
        {Veeraragavan2023SecuringFedGan}
\bibfield{author}{\bibinfo{person}{Narasimha~Raghavan Veeraragavan} {and} \bibinfo{person}{Jan~Franz Nyg\r{a}rd}.} \bibinfo{year}{2023}\natexlab{}.
\newblock \showarticletitle{Securing Federated GANs: Enabling Synthetic Data Generation for Health Registry Consortiums}. In \bibinfo{booktitle}{\emph{Proceedings of the 18th International Conference on Availability, Reliability and Security}} (<conf-loc>, <city>Benevento</city>, <country>Italy</country>, </conf-loc>) \emph{(\bibinfo{series}{ARES '23})}. \bibinfo{publisher}{Association for Computing Machinery}, \bibinfo{address}{New York, NY, USA}, Article \bibinfo{articleno}{89}, \bibinfo{numpages}{9}~pages.
\newblock
\showISBNx{9798400707728}
\urldef\tempurl%
\url{https://doi.org/10.1145/3600160.3605041}
\showDOI{\tempurl}


\bibitem[Vyas and Rajendran(2023)]%
        {vyas2023generative}
\bibfield{author}{\bibinfo{person}{Bhuman Vyas} {and} \bibinfo{person}{Rajashree~Manjulalayam Rajendran}.} \bibinfo{year}{2023}\natexlab{}.
\newblock \showarticletitle{Generative Adversarial Networks for Anomaly Detection in Medical Images}.
\newblock \bibinfo{journal}{\emph{International Journal of Multidisciplinary Innovation and Research Methodology, ISSN: 2960-2068}} \bibinfo{volume}{2}, \bibinfo{number}{4} (\bibinfo{year}{2023}), \bibinfo{pages}{52--58}.
\newblock


\bibitem[Wan et~al\mbox{.}(2017)]%
        {wan2017variational}
\bibfield{author}{\bibinfo{person}{Zhiqiang Wan}, \bibinfo{person}{Yazhou Zhang}, {and} \bibinfo{person}{Haibo He}.} \bibinfo{year}{2017}\natexlab{}.
\newblock \showarticletitle{Variational autoencoder based synthetic data generation for imbalanced learning}. In \bibinfo{booktitle}{\emph{2017 IEEE symposium series on computational intelligence (SSCI)}}. IEEE, \bibinfo{pages}{1--7}.
\newblock


\bibitem[Wang et~al\mbox{.}(2023a)]%
        {Wang2023Poisoning}
\bibfield{author}{\bibinfo{person}{Chao Wang}, \bibinfo{person}{Xiuyuan Liu}, \bibinfo{person}{Yunhua He}, \bibinfo{person}{Ke Xiao}, {and} \bibinfo{person}{Wei Li}.} \bibinfo{year}{2023}\natexlab{a}.
\newblock \showarticletitle{Poisoning the Competition: Fake Gradient Attacks on Distributed Generative Adversarial Networks}. In \bibinfo{booktitle}{\emph{2023 IEEE 20th International Conference on Mobile Ad Hoc and Smart Systems (MASS)}}. \bibinfo{pages}{487--495}.
\newblock
\urldef\tempurl%
\url{https://doi.org/10.1109/MASS58611.2023.00067}
\showDOI{\tempurl}


\bibitem[Wang et~al\mbox{.}(2023c)]%
        {WANG2023FedMedGAN}
\bibfield{author}{\bibinfo{person}{Jinbao Wang}, \bibinfo{person}{Guoyang Xie}, \bibinfo{person}{Yawen Huang}, \bibinfo{person}{Jiayi Lyu}, \bibinfo{person}{Feng Zheng}, \bibinfo{person}{Yefeng Zheng}, {and} \bibinfo{person}{Yaochu Jin}.} \bibinfo{year}{2023}\natexlab{c}.
\newblock \showarticletitle{FedMed-GAN: Federated domain translation on unsupervised cross-modality brain image synthesis}.
\newblock \bibinfo{journal}{\emph{Neurocomputing}}  \bibinfo{volume}{546} (\bibinfo{year}{2023}), \bibinfo{pages}{126282}.
\newblock
\showISSN{0925-2312}
\urldef\tempurl%
\url{https://doi.org/10.1016/j.neucom.2023.126282}
\showDOI{\tempurl}


\bibitem[Wang et~al\mbox{.}(2017)]%
        {wang2017generative}
\bibfield{author}{\bibinfo{person}{Kunfeng Wang}, \bibinfo{person}{Chao Gou}, \bibinfo{person}{Yanjie Duan}, \bibinfo{person}{Yilun Lin}, \bibinfo{person}{Xinhu Zheng}, {and} \bibinfo{person}{Fei-Yue Wang}.} \bibinfo{year}{2017}\natexlab{}.
\newblock \showarticletitle{Generative adversarial networks: introduction and outlook}.
\newblock \bibinfo{journal}{\emph{IEEE/CAA Journal of Automatica Sinica}} \bibinfo{volume}{4}, \bibinfo{number}{4} (\bibinfo{year}{2017}), \bibinfo{pages}{588--598}.
\newblock


\bibitem[Wang et~al\mbox{.}(2023b)]%
        {wang2023fda}
\bibfield{author}{\bibinfo{person}{Yue Wang}, \bibinfo{person}{Hongjuan Wang}, \bibinfo{person}{Fang Zhang}, {and} \bibinfo{person}{Xuxin Li}.} \bibinfo{year}{2023}\natexlab{b}.
\newblock \showarticletitle{FDA-CDM: Data Augmentation Framework for Personalized Federated Learning in Non-IID Scenarios}. In \bibinfo{booktitle}{\emph{Proceedings of the 2023 International Conference on Electronics, Computers and Communication Technology}}. \bibinfo{pages}{71--76}.
\newblock


\bibitem[Wei et~al\mbox{.}(2020)]%
        {wei2020federated}
\bibfield{author}{\bibinfo{person}{Kang Wei}, \bibinfo{person}{Jun Li}, \bibinfo{person}{Ming Ding}, \bibinfo{person}{Chuan Ma}, \bibinfo{person}{Howard~H Yang}, \bibinfo{person}{Farhad Farokhi}, \bibinfo{person}{Shi Jin}, \bibinfo{person}{Tony~QS Quek}, {and} \bibinfo{person}{H~Vincent Poor}.} \bibinfo{year}{2020}\natexlab{}.
\newblock \showarticletitle{Federated learning with differential privacy: Algorithms and performance analysis}.
\newblock \bibinfo{journal}{\emph{IEEE transactions on information forensics and security}}  \bibinfo{volume}{15} (\bibinfo{year}{2020}), \bibinfo{pages}{3454--3469}.
\newblock


\bibitem[Wei et~al\mbox{.}(2022)]%
        {wei2022vertical}
\bibfield{author}{\bibinfo{person}{Kang Wei}, \bibinfo{person}{Jun Li}, \bibinfo{person}{Chuan Ma}, \bibinfo{person}{Ming Ding}, \bibinfo{person}{Sha Wei}, \bibinfo{person}{Fan Wu}, \bibinfo{person}{Guihai Chen}, {and} \bibinfo{person}{Thilina Ranbaduge}.} \bibinfo{year}{2022}\natexlab{}.
\newblock \showarticletitle{Vertical federated learning: Challenges, methodologies and experiments}.
\newblock \bibinfo{journal}{\emph{arXiv:2202.04309}} (\bibinfo{year}{2022}).
\newblock


\bibitem[Wen et~al\mbox{.}(2023)]%
        {wen2023survey}
\bibfield{author}{\bibinfo{person}{Jie Wen}, \bibinfo{person}{Zhixia Zhang}, \bibinfo{person}{Yang Lan}, \bibinfo{person}{Zhihua Cui}, \bibinfo{person}{Jianghui Cai}, {and} \bibinfo{person}{Wensheng Zhang}.} \bibinfo{year}{2023}\natexlab{}.
\newblock \showarticletitle{A survey on federated learning: challenges and applications}.
\newblock \bibinfo{journal}{\emph{International Journal of Machine Learning and Cybernetics}} \bibinfo{volume}{14}, \bibinfo{number}{2} (\bibinfo{year}{2023}), \bibinfo{pages}{513--535}.
\newblock


\bibitem[Wijesinghe et~al\mbox{.}(2023a)]%
        {wijesinghe2023pfl}
\bibfield{author}{\bibinfo{person}{Achintha Wijesinghe}, \bibinfo{person}{Songyang Zhang}, {and} \bibinfo{person}{Zhi Ding}.} \bibinfo{year}{2023}\natexlab{a}.
\newblock \showarticletitle{PFL-GAN: When Client Heterogeneity Meets Generative Models in Personalized Federated Learning}.
\newblock \bibinfo{journal}{\emph{arXiv:2308.12454}} (\bibinfo{year}{2023}).
\newblock


\bibitem[Wijesinghe et~al\mbox{.}(2023b)]%
        {wijesinghe2023psfedgan}
\bibfield{author}{\bibinfo{person}{Achintha Wijesinghe}, \bibinfo{person}{Songyang Zhang}, {and} \bibinfo{person}{Zhi Ding}.} \bibinfo{year}{2023}\natexlab{b}.
\newblock \showarticletitle{PS-FedGAN: An Efficient Federated Learning Framework Based on Partially Shared Generative Adversarial Networks For Data Privacy}.
\newblock \bibinfo{journal}{\emph{arXiv:2305.11437}} (\bibinfo{year}{2023}).
\newblock


\bibitem[Wijesinghe et~al\mbox{.}(2023c)]%
        {wijesinghe2023ufed}
\bibfield{author}{\bibinfo{person}{Achintha Wijesinghe}, \bibinfo{person}{Songyang Zhang}, \bibinfo{person}{Siyu Qi}, {and} \bibinfo{person}{Zhi Ding}.} \bibinfo{year}{2023}\natexlab{c}.
\newblock \showarticletitle{UFed-GAN: A Secure Federated Learning Framework with Constrained Computation and Unlabeled Data}.
\newblock \bibinfo{journal}{\emph{arXiv:2308.05870}} (\bibinfo{year}{2023}).
\newblock


\bibitem[Wu et~al\mbox{.}(2022)]%
        {Wu2022FedCG}
\bibfield{author}{\bibinfo{person}{Yuezhou Wu}, \bibinfo{person}{Yan Kang}, \bibinfo{person}{Jiahuan Luo}, \bibinfo{person}{Yuanqin He}, \bibinfo{person}{Lixin Fan}, \bibinfo{person}{Rong Pan}, {and} \bibinfo{person}{Qiang Yang}.} \bibinfo{year}{2022}\natexlab{}.
\newblock \showarticletitle{FedCG: Leverage Conditional GAN for Protecting Privacy and Maintaining Competitive Performance in Federated Learning}. In \bibinfo{booktitle}{\emph{Proceedings of the Thirty-First International Joint Conference on Artificial Intelligence}} \emph{(\bibinfo{series}{IJCAI-2022})}. \bibinfo{publisher}{International Joint Conferences on Artificial Intelligence Organization}.
\newblock
\urldef\tempurl%
\url{https://doi.org/10.24963/ijcai.2022/324}
\showDOI{\tempurl}


\bibitem[Xin et~al\mbox{.}(2022)]%
        {federatedDP2022xin}
\bibfield{author}{\bibinfo{person}{Bangzhou Xin}, \bibinfo{person}{Yangyang Geng}, \bibinfo{person}{Teng Hu}, \bibinfo{person}{Sheng Chen}, \bibinfo{person}{Wei Yang}, \bibinfo{person}{Shaowei Wang}, {and} \bibinfo{person}{Liusheng Huang}.} \bibinfo{year}{2022}\natexlab{}.
\newblock \showarticletitle{Federated synthetic data generation with differential privacy}.
\newblock \bibinfo{journal}{\emph{Neurocomputing}}  \bibinfo{volume}{468} (\bibinfo{year}{2022}), \bibinfo{pages}{1--10}.
\newblock
\showISSN{0925-2312}
\urldef\tempurl%
\url{https://doi.org/10.1016/j.neucom.2021.10.027}
\showDOI{\tempurl}


\bibitem[Xin et~al\mbox{.}(2020)]%
        {privateFLGAN2020Xin}
\bibfield{author}{\bibinfo{person}{Bangzhou Xin}, \bibinfo{person}{Wei Yang}, \bibinfo{person}{Yangyang Geng}, \bibinfo{person}{Sheng Chen}, \bibinfo{person}{Shaowei Wang}, {and} \bibinfo{person}{Liusheng Huang}.} \bibinfo{year}{2020}\natexlab{}.
\newblock \showarticletitle{Private FL-GAN: Differential Privacy Synthetic Data Generation Based on Federated Learning}. In \bibinfo{booktitle}{\emph{ICASSP 2020 - 2020 IEEE International Conference on Acoustics, Speech and Signal Processing (ICASSP)}}. \bibinfo{pages}{2927--2931}.
\newblock
\urldef\tempurl%
\url{https://doi.org/10.1109/ICASSP40776.2020.9054559}
\showDOI{\tempurl}


\bibitem[Xiong et~al\mbox{.}(2023)]%
        {xiong2023feddm}
\bibfield{author}{\bibinfo{person}{Yuanhao Xiong}, \bibinfo{person}{Ruochen Wang}, \bibinfo{person}{Minhao Cheng}, \bibinfo{person}{Felix Yu}, {and} \bibinfo{person}{Cho-Jui Hsieh}.} \bibinfo{year}{2023}\natexlab{}.
\newblock \showarticletitle{Feddm: Iterative distribution matching for communication-efficient federated learning}. In \bibinfo{booktitle}{\emph{Proceedings of the IEEE/CVF Conference on Computer Vision and Pattern Recognition}}. \bibinfo{pages}{16323--16332}.
\newblock


\bibitem[Yang et~al\mbox{.}(2022)]%
        {yang2022using}
\bibfield{author}{\bibinfo{person}{Haomiao Yang}, \bibinfo{person}{Mengyu Ge}, \bibinfo{person}{Kunlan Xiang}, {and} \bibinfo{person}{Jingwei Li}.} \bibinfo{year}{2022}\natexlab{}.
\newblock \showarticletitle{Using highly compressed gradients in federated learning for data reconstruction attacks}.
\newblock \bibinfo{journal}{\emph{IEEE Transactions on Information Forensics and Security}}  \bibinfo{volume}{18} (\bibinfo{year}{2022}), \bibinfo{pages}{818--830}.
\newblock


\bibitem[Yang et~al\mbox{.}(2023c)]%
        {yang2023diffusion}
\bibfield{author}{\bibinfo{person}{Ling Yang}, \bibinfo{person}{Zhilong Zhang}, \bibinfo{person}{Yang Song}, \bibinfo{person}{Shenda Hong}, \bibinfo{person}{Runsheng Xu}, \bibinfo{person}{Yue Zhao}, \bibinfo{person}{Wentao Zhang}, \bibinfo{person}{Bin Cui}, {and} \bibinfo{person}{Ming-Hsuan Yang}.} \bibinfo{year}{2023}\natexlab{c}.
\newblock \showarticletitle{Diffusion models: A comprehensive survey of methods and applications}.
\newblock \bibinfo{journal}{\emph{Comput. Surveys}} \bibinfo{volume}{56}, \bibinfo{number}{4} (\bibinfo{year}{2023}), \bibinfo{pages}{1--39}.
\newblock


\bibitem[Yang et~al\mbox{.}(2023a)]%
        {yang2023model}
\bibfield{author}{\bibinfo{person}{Ming Yang}, \bibinfo{person}{Hang Cheng}, \bibinfo{person}{Fei Chen}, \bibinfo{person}{Ximeng Liu}, \bibinfo{person}{Meiqing Wang}, {and} \bibinfo{person}{Xibin Li}.} \bibinfo{year}{2023}\natexlab{a}.
\newblock \showarticletitle{Model poisoning attack in differential privacy-based federated learning}.
\newblock \bibinfo{journal}{\emph{Information Sciences}}  \bibinfo{volume}{630} (\bibinfo{year}{2023}), \bibinfo{pages}{158--172}.
\newblock


\bibitem[Yang et~al\mbox{.}(2023b)]%
        {yang2023one}
\bibfield{author}{\bibinfo{person}{Mingzhao Yang}, \bibinfo{person}{Shangchao Su}, \bibinfo{person}{Bin Li}, {and} \bibinfo{person}{Xiangyang Xue}.} \bibinfo{year}{2023}\natexlab{b}.
\newblock \showarticletitle{One-Shot Federated Learning with Classifier-Guided Diffusion Models}.
\newblock \bibinfo{journal}{\emph{arXiv:2311.08870}} (\bibinfo{year}{2023}).
\newblock


\bibitem[Yang et~al\mbox{.}(2024a)]%
        {yang2024exploring}
\bibfield{author}{\bibinfo{person}{Mingzhao Yang}, \bibinfo{person}{Shangchao Su}, \bibinfo{person}{Bin Li}, {and} \bibinfo{person}{Xiangyang Xue}.} \bibinfo{year}{2024}\natexlab{a}.
\newblock \showarticletitle{Exploring One-Shot Semi-supervised Federated Learning with Pre-trained Diffusion Models}. In \bibinfo{booktitle}{\emph{Proceedings of the AAAI Conference on Artificial Intelligence}}. \bibinfo{pages}{16325--16333}.
\newblock


\bibitem[Yang et~al\mbox{.}(2019)]%
        {yang2019federated}
\bibfield{author}{\bibinfo{person}{Qiang Yang}, \bibinfo{person}{Yang Liu}, \bibinfo{person}{Tianjian Chen}, {and} \bibinfo{person}{Yongxin Tong}.} \bibinfo{year}{2019}\natexlab{}.
\newblock \showarticletitle{Federated machine learning: Concept and applications}.
\newblock \bibinfo{journal}{\emph{ACM Transactions on Intelligent Systems and Technology (TIST)}} \bibinfo{volume}{10}, \bibinfo{number}{2} (\bibinfo{year}{2019}), \bibinfo{pages}{1--19}.
\newblock


\bibitem[Yang et~al\mbox{.}(2024b)]%
        {yang2024fedfed}
\bibfield{author}{\bibinfo{person}{Zhiqin Yang}, \bibinfo{person}{Yonggang Zhang}, \bibinfo{person}{Yu Zheng}, \bibinfo{person}{Xinmei Tian}, \bibinfo{person}{Hao Peng}, \bibinfo{person}{Tongliang Liu}, {and} \bibinfo{person}{Bo Han}.} \bibinfo{year}{2024}\natexlab{b}.
\newblock \showarticletitle{FedFed: Feature distillation against data heterogeneity in federated learning}.
\newblock \bibinfo{journal}{\emph{Advances in Neural Information Processing Systems}}  \bibinfo{volume}{36} (\bibinfo{year}{2024}).
\newblock


\bibitem[Yao and Zhao(2022)]%
        {FedTMI2022Yao}
\bibfield{author}{\bibinfo{person}{Zoujing Yao} {and} \bibinfo{person}{Chunhui Zhao}.} \bibinfo{year}{2022}\natexlab{}.
\newblock \showarticletitle{FedTMI: Knowledge aided federated transfer learning for industrial missing data imputation}.
\newblock \bibinfo{journal}{\emph{Journal of Process Control}}  \bibinfo{volume}{117} (\bibinfo{year}{2022}), \bibinfo{pages}{206--215}.
\newblock
\showISSN{0959-1524}
\urldef\tempurl%
\url{https://doi.org/10.1016/j.jprocont.2022.08.004}
\showDOI{\tempurl}


\bibitem[Ye et~al\mbox{.}(2023)]%
        {ye2023heterogeneoussurvey}
\bibfield{author}{\bibinfo{person}{Mang Ye}, \bibinfo{person}{Xiuwen Fang}, \bibinfo{person}{Bo Du}, \bibinfo{person}{Pong~C Yuen}, {and} \bibinfo{person}{Dacheng Tao}.} \bibinfo{year}{2023}\natexlab{}.
\newblock \showarticletitle{Heterogeneous federated learning: State-of-the-art and research challenges}.
\newblock \bibinfo{journal}{\emph{Comput. Surveys}} \bibinfo{volume}{56}, \bibinfo{number}{3} (\bibinfo{year}{2023}), \bibinfo{pages}{1--44}.
\newblock


\bibitem[Yoon et~al\mbox{.}(2021)]%
        {yoon2021fedmix}
\bibfield{author}{\bibinfo{person}{Tehrim Yoon}, \bibinfo{person}{Sumin Shin}, \bibinfo{person}{Sung~Ju Hwang}, {and} \bibinfo{person}{Eunho Yang}.} \bibinfo{year}{2021}\natexlab{}.
\newblock \showarticletitle{FedMix: Approximation of Mixup under Mean Augmented Federated Learning}. In \bibinfo{booktitle}{\emph{International Conference on Learning Representations}}.
\newblock
\urldef\tempurl%
\url{https://openreview.net/forum?id=Ogga20D2HO-}
\showURL{%
\tempurl}


\bibitem[Zhai et~al\mbox{.}(2018)]%
        {zhai2018autoencoder}
\bibfield{author}{\bibinfo{person}{Junhai Zhai}, \bibinfo{person}{Sufang Zhang}, \bibinfo{person}{Junfen Chen}, {and} \bibinfo{person}{Qiang He}.} \bibinfo{year}{2018}\natexlab{}.
\newblock \showarticletitle{Autoencoder and its various variants}. In \bibinfo{booktitle}{\emph{2018 IEEE international conference on systems, man, and cybernetics (SMC)}}. IEEE, \bibinfo{pages}{415--419}.
\newblock


\bibitem[Zhang et~al\mbox{.}(2021c)]%
        {zhang2021survey}
\bibfield{author}{\bibinfo{person}{Chen Zhang}, \bibinfo{person}{Yu Xie}, \bibinfo{person}{Hang Bai}, \bibinfo{person}{Bin Yu}, \bibinfo{person}{Weihong Li}, {and} \bibinfo{person}{Yuan Gao}.} \bibinfo{year}{2021}\natexlab{c}.
\newblock \showarticletitle{A survey on federated learning}.
\newblock \bibinfo{journal}{\emph{Knowledge-Based Systems}}  \bibinfo{volume}{216} (\bibinfo{year}{2021}), \bibinfo{pages}{106775}.
\newblock


\bibitem[Zhang et~al\mbox{.}(2022)]%
        {zhang2022dense}
\bibfield{author}{\bibinfo{person}{Jie Zhang}, \bibinfo{person}{Chen Chen}, \bibinfo{person}{Bo Li}, \bibinfo{person}{Lingjuan Lyu}, \bibinfo{person}{Shuang Wu}, \bibinfo{person}{Shouhong Ding}, \bibinfo{person}{Chunhua Shen}, {and} \bibinfo{person}{Chao Wu}.} \bibinfo{year}{2022}\natexlab{}.
\newblock \showarticletitle{{DENSE}: Data-Free One-Shot Federated Learning}. In \bibinfo{booktitle}{\emph{Advances in Neural Information Processing Systems}}, \bibfield{editor}{\bibinfo{person}{Alice~H. Oh}, \bibinfo{person}{Alekh Agarwal}, \bibinfo{person}{Danielle Belgrave}, {and} \bibinfo{person}{Kyunghyun Cho}} (Eds.).
\newblock
\urldef\tempurl%
\url{https://openreview.net/forum?id=QFQoxCFYEkA}
\showURL{%
\tempurl}


\bibitem[Zhang et~al\mbox{.}(2019)]%
        {zhang2019poisoning}
\bibfield{author}{\bibinfo{person}{Jiale Zhang}, \bibinfo{person}{Junjun Chen}, \bibinfo{person}{Di Wu}, \bibinfo{person}{Bing Chen}, {and} \bibinfo{person}{Shui Yu}.} \bibinfo{year}{2019}\natexlab{}.
\newblock \showarticletitle{Poisoning attack in federated learning using generative adversarial nets}. In \bibinfo{booktitle}{\emph{2019 18th IEEE international conference on trust, security and privacy in computing and communications/13th IEEE international conference on big data science and engineering (TrustCom/BigDataSE)}}. IEEE, \bibinfo{pages}{374--380}.
\newblock


\bibitem[Zhang et~al\mbox{.}(2020)]%
        {Zhang2020GANmembership}
\bibfield{author}{\bibinfo{person}{Jingwen Zhang}, \bibinfo{person}{Jiale Zhang}, \bibinfo{person}{Junjun Chen}, {and} \bibinfo{person}{Shui Yu}.} \bibinfo{year}{2020}\natexlab{}.
\newblock \showarticletitle{GAN Enhanced Membership Inference: A Passive Local Attack in Federated Learning}. In \bibinfo{booktitle}{\emph{ICC 2020 - 2020 IEEE International Conference on Communications (ICC)}}. \bibinfo{pages}{1--6}.
\newblock
\urldef\tempurl%
\url{https://doi.org/10.1109/ICC40277.2020.9148790}
\showDOI{\tempurl}


\bibitem[Zhang et~al\mbox{.}(2023b)]%
        {zhang2022flschemeGAN}
\bibfield{author}{\bibinfo{person}{Jiaxin Zhang}, \bibinfo{person}{Liang Zhao}, \bibinfo{person}{Keping Yu}, \bibinfo{person}{Geyong Min}, \bibinfo{person}{Ahmed~Y. Al-Dubai}, {and} \bibinfo{person}{Albert~Y. Zomaya}.} \bibinfo{year}{2023}\natexlab{b}.
\newblock \showarticletitle{A Novel Federated Learning Scheme for Generative Adversarial Networks}.
\newblock \bibinfo{journal}{\emph{IEEE Transactions on Mobile Computing}} (\bibinfo{year}{2023}), \bibinfo{pages}{1--17}.
\newblock
\urldef\tempurl%
\url{https://doi.org/10.1109/TMC.2023.3278668}
\showDOI{\tempurl}


\bibitem[Zhang et~al\mbox{.}(2021b)]%
        {zhang2021feddpgan}
\bibfield{author}{\bibinfo{person}{Longling Zhang}, \bibinfo{person}{Bochen Shen}, \bibinfo{person}{Ahmed Barnawi}, \bibinfo{person}{Shan Xi}, \bibinfo{person}{Neeraj Kumar}, {and} \bibinfo{person}{Yi Wu}.} \bibinfo{year}{2021}\natexlab{b}.
\newblock \showarticletitle{FedDPGAN: federated differentially private generative adversarial networks framework for the detection of COVID-19 pneumonia}.
\newblock \bibinfo{journal}{\emph{Information Systems Frontiers}} \bibinfo{volume}{23}, \bibinfo{number}{6} (\bibinfo{year}{2021}), \bibinfo{pages}{1403--1415}.
\newblock


\bibitem[Zhang et~al\mbox{.}(2021d)]%
        {zhang2021dance}
\bibfield{author}{\bibinfo{person}{Xiongtao Zhang}, \bibinfo{person}{Xiaomin Zhu}, \bibinfo{person}{Ji Wang}, \bibinfo{person}{Weidong Bao}, {and} \bibinfo{person}{Laurence~T Yang}.} \bibinfo{year}{2021}\natexlab{d}.
\newblock \showarticletitle{Dance: Distributed generative adversarial networks with communication compression}.
\newblock \bibinfo{journal}{\emph{ACM Transactions on Internet Technology (TOIT)}} \bibinfo{volume}{22}, \bibinfo{number}{2} (\bibinfo{year}{2021}), \bibinfo{pages}{1--32}.
\newblock


\bibitem[Zhang et~al\mbox{.}(2023a)]%
        {zhang2023systematic}
\bibfield{author}{\bibinfo{person}{Yi Zhang}, \bibinfo{person}{Yunfan Lu}, {and} \bibinfo{person}{Fengxia Liu}.} \bibinfo{year}{2023}\natexlab{a}.
\newblock \showarticletitle{A systematic survey for differential privacy techniques in federated learning}.
\newblock \bibinfo{journal}{\emph{Journal of Information Security}} \bibinfo{volume}{14}, \bibinfo{number}{2} (\bibinfo{year}{2023}), \bibinfo{pages}{111--135}.
\newblock


\bibitem[Zhang et~al\mbox{.}(2021a)]%
        {zhang2021training}
\bibfield{author}{\bibinfo{person}{Yikai Zhang}, \bibinfo{person}{Hui Qu}, \bibinfo{person}{Qi Chang}, \bibinfo{person}{Huidong Liu}, \bibinfo{person}{Dimitris Metaxas}, {and} \bibinfo{person}{Chao Chen}.} \bibinfo{year}{2021}\natexlab{a}.
\newblock \showarticletitle{Training federated GANs with theoretical guarantees: A universal aggregation approach}.
\newblock \bibinfo{journal}{\emph{arXiv:2102.04655}} (\bibinfo{year}{2021}).
\newblock


\bibitem[Zhao et~al\mbox{.}(2021b)]%
        {zhao2021efficient}
\bibfield{author}{\bibinfo{person}{Jie Zhao}, \bibinfo{person}{Xinghua Zhu}, \bibinfo{person}{Jianzong Wang}, {and} \bibinfo{person}{Jing Xiao}.} \bibinfo{year}{2021}\natexlab{b}.
\newblock \showarticletitle{Efficient client contribution evaluation for horizontal federated learning}. In \bibinfo{booktitle}{\emph{ICASSP 2021-2021 IEEE International Conference on Acoustics, Speech and Signal Processing (ICASSP)}}. IEEE, \bibinfo{pages}{3060--3064}.
\newblock


\bibitem[Zhao et~al\mbox{.}(2017)]%
        {zhao2017learning}
\bibfield{author}{\bibinfo{person}{Shengjia Zhao}, \bibinfo{person}{Jiaming Song}, {and} \bibinfo{person}{Stefano Ermon}.} \bibinfo{year}{2017}\natexlab{}.
\newblock \showarticletitle{Learning hierarchical features from deep generative models}. In \bibinfo{booktitle}{\emph{International Conference on Machine Learning}}. PMLR, \bibinfo{pages}{4091--4099}.
\newblock


\bibitem[Zhao et~al\mbox{.}(2022)]%
        {zhao2022detecting}
\bibfield{author}{\bibinfo{person}{Ying Zhao}, \bibinfo{person}{Junjun Chen}, \bibinfo{person}{Jiale Zhang}, \bibinfo{person}{Di Wu}, \bibinfo{person}{Michael Blumenstein}, {and} \bibinfo{person}{Shui Yu}.} \bibinfo{year}{2022}\natexlab{}.
\newblock \showarticletitle{Detecting and mitigating poisoning attacks in federated learning using generative adversarial networks}.
\newblock \bibinfo{journal}{\emph{Concurrency and Computation: Practice and Experience}} \bibinfo{volume}{34}, \bibinfo{number}{7} (\bibinfo{year}{2022}), \bibinfo{pages}{e5906}.
\newblock


\bibitem[Zhao et~al\mbox{.}(2020)]%
        {zhao2020pdgan}
\bibfield{author}{\bibinfo{person}{Ying Zhao}, \bibinfo{person}{Junjun Chen}, \bibinfo{person}{Jiale Zhang}, \bibinfo{person}{Di Wu}, \bibinfo{person}{Jian Teng}, {and} \bibinfo{person}{Shui Yu}.} \bibinfo{year}{2020}\natexlab{}.
\newblock \showarticletitle{PDGAN: A novel poisoning defense method in federated learning using generative adversarial network}. In \bibinfo{booktitle}{\emph{Algorithms and Architectures for Parallel Processing: 19th International Conference, ICA3PP 2019, Melbourne, VIC, Australia, December 9--11, 2019, Proceedings, Part I 19}}. Springer, \bibinfo{pages}{595--609}.
\newblock


\bibitem[Zhao et~al\mbox{.}(2018)]%
        {zhao2018federated}
\bibfield{author}{\bibinfo{person}{Yue Zhao}, \bibinfo{person}{Meng Li}, \bibinfo{person}{Liangzhen Lai}, \bibinfo{person}{Naveen Suda}, \bibinfo{person}{Damon Civin}, {and} \bibinfo{person}{Vikas Chandra}.} \bibinfo{year}{2018}\natexlab{}.
\newblock \showarticletitle{Federated learning with non-iid data}.
\newblock \bibinfo{journal}{\emph{arXiv:1806.00582}} (\bibinfo{year}{2018}).
\newblock


\bibitem[Zhao et~al\mbox{.}(2023a)]%
        {zhao2023gdts}
\bibfield{author}{\bibinfo{person}{Zilong Zhao}, \bibinfo{person}{Robert Birke}, {and} \bibinfo{person}{Lydia~Y. Chen}.} \bibinfo{year}{2023}\natexlab{a}.
\newblock \showarticletitle{GDTS: GAN-Based Distributed Tabular Synthesizer}. In \bibinfo{booktitle}{\emph{2023 IEEE 16th International Conference on Cloud Computing (CLOUD)}}. \bibinfo{pages}{570--576}.
\newblock
\urldef\tempurl%
\url{https://doi.org/10.1109/CLOUD60044.2023.00078}
\showDOI{\tempurl}


\bibitem[Zhao et~al\mbox{.}(2021a)]%
        {zhao2021fed}
\bibfield{author}{\bibinfo{person}{Zilong Zhao}, \bibinfo{person}{Robert Birke}, \bibinfo{person}{Aditya Kunar}, {and} \bibinfo{person}{Lydia~Y Chen}.} \bibinfo{year}{2021}\natexlab{a}.
\newblock \showarticletitle{Fed-tgan: Federated learning framework for synthesizing tabular data}.
\newblock \bibinfo{journal}{\emph{arXiv:2108.07927}} (\bibinfo{year}{2021}).
\newblock


\bibitem[Zhao et~al\mbox{.}(2023b)]%
        {zhao2023gtv}
\bibfield{author}{\bibinfo{person}{Zilong Zhao}, \bibinfo{person}{Han Wu}, \bibinfo{person}{Aad Van~Moorsel}, {and} \bibinfo{person}{Lydia~Y Chen}.} \bibinfo{year}{2023}\natexlab{b}.
\newblock \showarticletitle{Gtv: Generating tabular data via vertical federated learning}.
\newblock \bibinfo{journal}{\emph{arXiv:2302.01706}} (\bibinfo{year}{2023}).
\newblock


\bibitem[Zhao et~al\mbox{.}(2023c)]%
        {zhao2023federated}
\bibfield{author}{\bibinfo{person}{Zhuang Zhao}, \bibinfo{person}{Feng Yang}, {and} \bibinfo{person}{Guirong Liang}.} \bibinfo{year}{2023}\natexlab{c}.
\newblock \showarticletitle{Federated Learning Based on Diffusion Model to Cope with Non-IID Data}. In \bibinfo{booktitle}{\emph{Chinese Conference on Pattern Recognition and Computer Vision (PRCV)}}. Springer, \bibinfo{pages}{220--231}.
\newblock


\bibitem[Zhou et~al\mbox{.}(2021)]%
        {Zhou2021Fed}
\bibfield{author}{\bibinfo{person}{Xu Zhou}, \bibinfo{person}{Xiaofeng Liu}, \bibinfo{person}{Gongjin Lan}, {and} \bibinfo{person}{Jian Wu}.} \bibinfo{year}{2021}\natexlab{}.
\newblock \showarticletitle{Federated conditional generative adversarial nets imputation method for air quality missing data}.
\newblock \bibinfo{journal}{\emph{Knowledge-Based Systems}}  \bibinfo{volume}{228} (\bibinfo{year}{2021}), \bibinfo{pages}{107261}.
\newblock
\showISSN{0950-7051}
\urldef\tempurl%
\url{https://doi.org/10.1016/j.knosys.2021.107261}
\showDOI{\tempurl}


\bibitem[Zhou et~al\mbox{.}(2023)]%
        {zhou2023communicationefficient}
\bibfield{author}{\bibinfo{person}{Yuhao Zhou}, \bibinfo{person}{Mingjia Shi}, \bibinfo{person}{Yuanxi Li}, \bibinfo{person}{Qing Ye}, \bibinfo{person}{Yanan Sun}, {and} \bibinfo{person}{Jiancheng Lv}.} \bibinfo{year}{2023}\natexlab{}.
\newblock \bibinfo{title}{Communication-efficient Federated Learning with Single-Step Synthetic Features Compressor for Faster Convergence}.
\newblock
\newblock
\showeprint[arxiv]{2302.13562}~[cs.LG]


\bibitem[Zhu et~al\mbox{.}(2021b)]%
        {zhu2021federated}
\bibfield{author}{\bibinfo{person}{Hangyu Zhu}, \bibinfo{person}{Jinjin Xu}, \bibinfo{person}{Shiqing Liu}, {and} \bibinfo{person}{Yaochu Jin}.} \bibinfo{year}{2021}\natexlab{b}.
\newblock \showarticletitle{Federated learning on non-IID data: A survey}.
\newblock \bibinfo{journal}{\emph{Neurocomputing}}  \bibinfo{volume}{465} (\bibinfo{year}{2021}), \bibinfo{pages}{371--390}.
\newblock


\bibitem[Zhu et~al\mbox{.}(2021a)]%
        {zhu2021data}
\bibfield{author}{\bibinfo{person}{Zhuangdi Zhu}, \bibinfo{person}{Junyuan Hong}, {and} \bibinfo{person}{Jiayu Zhou}.} \bibinfo{year}{2021}\natexlab{a}.
\newblock \showarticletitle{Data-free knowledge distillation for heterogeneous federated learning}. In \bibinfo{booktitle}{\emph{International conference on machine learning}}. PMLR, \bibinfo{pages}{12878--12889}.
\newblock


\end{thebibliography}
% \addbibresource{references.bib}
% \printbibliography

\end{document}